%% file: main_v1.tex
\documentclass[10pt,twocolumn,letterpaper]{article}

\usepackage[pagenumbers]{cvpr} 

\usepackage{graphicx}
\usepackage{amsmath}
\usepackage{amssymb}
\usepackage{booktabs}
\usepackage{multirow}
\usepackage{tabularx}
\usepackage{caption}
\usepackage{subcaption}
\usepackage{makecell}
\usepackage{soul}
\usepackage[accsupp]{axessibility}  

\usepackage[pagebackref,breaklinks,colorlinks,urlcolor=blue]{hyperref}
\usepackage{mathtools}
\newcommand{\defeq}{\vcentcolon=}
\usepackage{algorithm}
\usepackage{algorithmicx}  
\usepackage{algpseudocode}
\usepackage{pifont}
\usepackage[capitalize]{cleveref}
\crefname{section}{Sec.}{Secs.}
\Crefname{section}{Section}{Sections}
\Crefname{table}{Table}{Tables}
\crefname{table}{Tab.}{Tabs.}

\def\cvprPaperID{2101} 
\def\confName{CVPR}
\def\confYear{2022}

\begin{document}

\title{OpenTAL: Towards Open Set Temporal Action Localization}

\author{Wentao Bao, Qi Yu, Yu Kong\\
Rochester Institute of Technology, Rochester, NY 14623, USA\\
{\tt\small \{wb6219, qi.yu, yu.kong\}@rit.edu}
}
\maketitle

\begin{abstract}
   Temporal Action Localization (TAL) has experienced remarkable success under the supervised learning paradigm. However, existing TAL methods are rooted in the closed set assumption, which cannot handle the inevitable unknown actions in open-world scenarios. In this paper, we, for the first time, step toward the Open Set TAL (OSTAL) problem and propose a general framework \textbf{OpenTAL} based on Evidential Deep Learning (EDL). Specifically, the OpenTAL consists of uncertainty-aware action classification, actionness prediction, and temporal location regression. With the proposed importance-balanced EDL method, classification uncertainty is learned by collecting categorical evidence majorly from important samples. To distinguish the unknown actions from background video frames, the actionness is learned by the positive-unlabeled learning. The classification uncertainty is further calibrated by leveraging the guidance from the temporal localization quality. The OpenTAL is general to enable existing TAL models for open set scenarios, and experimental results on THUMOS14 and ActivityNet1.3 benchmarks show the effectiveness of our method. The code and pre-trained models are released at {\small{\url{https://www.rit.edu/actionlab/opental}}}.
\end{abstract}

\vspace{-2mm}\section{Introduction}
\label{sec:intro}

Temporal Action Localization (TAL) aims to temporally localize and recognize human actions in an untrimmed video. With the success of deep learning in video understanding~\cite{I3DCVPR2017,SlowFastICCV2019,KongArXiv2018,BaoICCV2021,Chen2020ECCV} and object detection~\cite{fasterrcnn_tpami,detr_eccv20,BaoIROS2020}, TAL has experienced remarkable advance in recent years~\cite{TALNet_CVPR2018,PGCN_ICCV2019,GTAD_CVPR2020,AFSD_CVPR2021}. However, these works are rooted in the closed set assumption that testing videos are assumed to contain only the pre-defined action categories, which is impractical in an open world where unknown human actions are inevitable to appear. In this paper, we for the first time step forward the Open Set Temporal Action Localization (OSTAL) problem. 

\begin{figure}
    \centering
    \includegraphics[width=\linewidth]{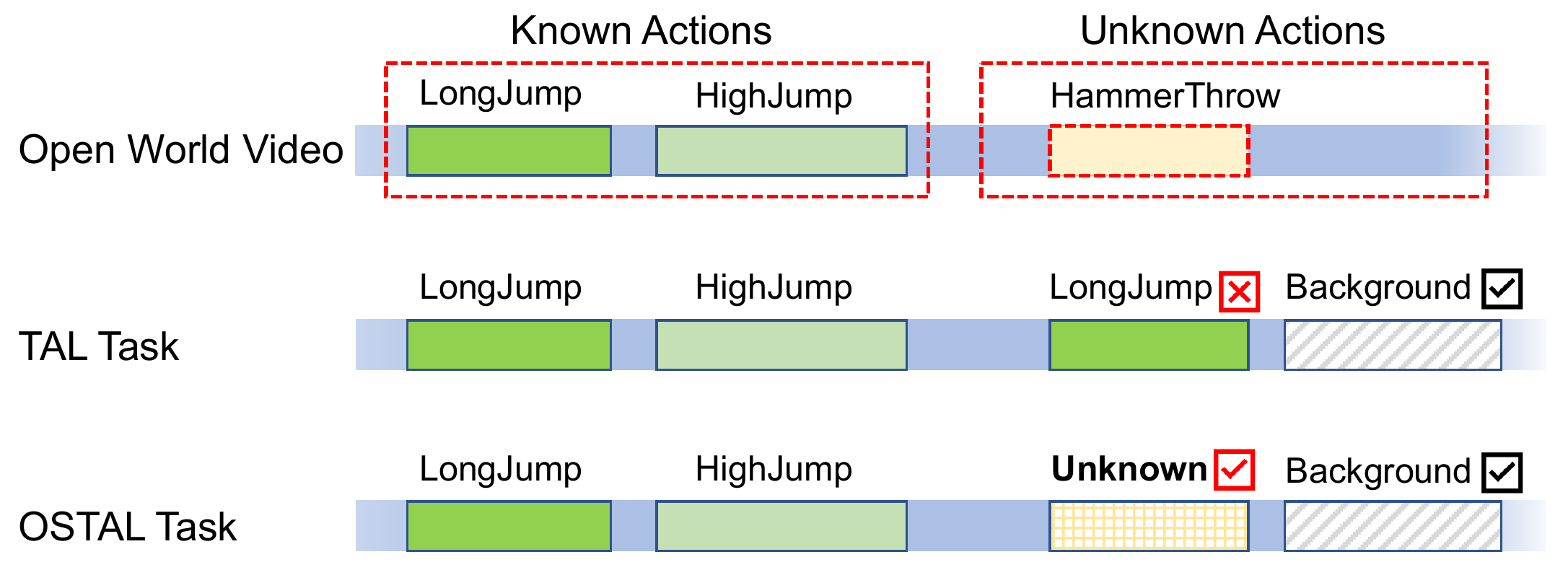}
    \caption{\textbf{OSTAL and TAL Tasks}. The OSTAL task is different from the TAL in that, there exist unknown actions in untrimmed open-world videos and the OSTAL models need to reject the positively localized action (e.g., \textit{HammerThrow}) as the \textit{Unknown}, rather than falsely assign it a known label such as the \textit{LongJump}.}
    \label{fig:ostal}
\end{figure}

OSTAL aims to not only temporally localize and recognize the known actions but also reject the localized unknown actions. As shown in Fig.~\ref{fig:ostal}, given an untrimmed video (the top row) from open world, traditional TAL (the middle row) could falsely accept the unknown action clip \emph{HammerThrow} as one of the known actions such as the \emph{LongJump}, while the proposed OSTAL (the bottom row) could correctly reject the clip as the \emph{Unknown}. Besides, both tasks need to differentiate between foreground actions and the \emph{Backgrounds} which are purely background frames.

The proposed OSTAL task is fundamentally more challenging than both the TAL and the closely relevant open set recognition (OSR)~\cite{ScheirerTPAMI2012} problems. On one hand, the recognition and localization of known actions become harder due to the mixture of background frames and the unknown foreground actions. Existing TAL methods typically assign the mixture with an non-informative \textit{Background} label or a wrong action label, which are unable to differentiate between them. On the other hand, different to the OSR problem, rejecting an unknown action is conditioned on positively localizing a foreground action so that the localization quality is critical to the OSTAL.

To tackle these challenges, we propose a general framework OpenTAL by decoupling the overall OSTAL objective into three interconnected components: uncertainty-aware action classification, actionness prediction, and temporal location regression. In essence, the foreground actions are distinguished from the background by the actionness prediction and localized by the temporal localization, while the known and unknown foreground actions are discriminated by the learned evidential uncertainty from the classification module. To achieve these goals, we propose three novel technical approaches as follows.

First, action classification is developed to recognize known actions and quantify the classification uncertainty by recent evidential deep learning (EDL)~\cite{SensoyNIPS2018,ZhaoAAAI2019,AminiNIPS2020,BaoICCV2021}. To enable this module to learn from important samples, we propose an importance-balanced EDL method by leveraging the magnitude of EDL gradient and evidential features.
Second, actionness prediction is to differentiate between foreground actions (positives) and background frames (negatives). In the open set setting, due to the mixture of unknown foreground actions (unlabeled) and background frames, learning from the labeled known actions and the mixture intrinsically reduces to a positive-unlabeled (PU) learning problem~\cite{BekkerML2020}.  To this end, we propose a PU learning method by selecting the top negative samples from the mixture as the true negatives. 
Third, the temporal localization module is trained to not only localize the known actions but also calibrate the classification uncertainty. We propose an IoU-aware uncertainty calibration (IoUC) method by using the temporal Intersection-over-Union (IoU) as the localization quality to calibrate the uncertainty. 

Based on the existing TAL datasets THUMOS14~\cite{THUMOS14} and ActivityNet1.3~\cite{ANetCVPR2015}, we set up a new benchmark to evaluate baselines and the proposed OpenTAL method for the OSTAL task, where the \emph{Open Set Detection Rate} is introduced to comprehensively evaluate the OSTAL performance. Experimental results show significant superiority of our method and indicate large room for improvement in this direction. Our main contribution is threefold:
\begin{itemize}
    \item To the best of our knowledge, this work is the first attempt on open set temporal action localization (OSTAL), which is fundamentally more challenging but highly valuable in open-world settings.
    \item We propose a general OpenTAL framework to address the unique challenges of OSTAL as compared with existing TAL and OSR problems. It is flexible to enable existing TAL models for open set scenarios. 
    \item The proposed importance-balanced EDL, PU learning, and IoUC methods are found effective for OSTAL tasks based on the OpenTAL framework.
\end{itemize}

\section{Related Work}

\paragraph{Temporal Action Localization}
The goal of Temporal Action Localization (TAL) is to recognize and temporally localize all the action instances in an untrimmed video. Existing TAL methods fall into two dominant paradigms: one-stage and two-stage approaches. 
The two-stage approaches~\cite{GTAD_CVPR2020, PGCN_ICCV2019, sridhar2021class, xu2017r, long2019gaussian} generate class-agnostic temporal proposals\cite{bai2020boundary,heilbron2016fast,lin2018bsn,lin2019bmn} at first and then perform the classification and boundary refinement of each proposal. The heuristic anchor design and the closed-set definition of the pre-trained proposal generation limit their applicability to the open-set problem.
One stage methods~\cite{yeung2016end,buch2017end,long2019gaussian,AFSD_CVPR2021} do not rely on the action proposal generation and can be typically trained in an end-to-end manner. These methods obtain the temporal boundaries first based on frame-level features and then perform global reasoning by multi-stage refinement or modeling the temporal transitions. 
Recently, AFSD~\cite{AFSD_CVPR2021} is proposed following the anchor-free design without actionness and proposals, which is a lightweight and flexible framework. While a lot of recent methods focus on improving the proposal generation~\cite{bai2020boundary,heilbron2016fast,lin2018bsn,lin2019bmn, zhao2020bottom} or boundary refinement~\cite{AFSD_CVPR2021}, a few focus on boosting the classification accuracy~\cite{zhao2017temporal,shou2017cdc,zhu2021enriching}. 

The above approaches assume that all of the action instances in untrimmed videos belong to pre-defined categories, which impedes their application to open-world scenarios. Though the open set is considered in~\cite{Zou2021MM}, their method is designed for efficient annotation in few-shot learning tasks. In this paper, an OSTAL problem is formulated to handle the unknown actions in TAL applications.

\begin{figure*}
    \centering
    \includegraphics[width=\textwidth]{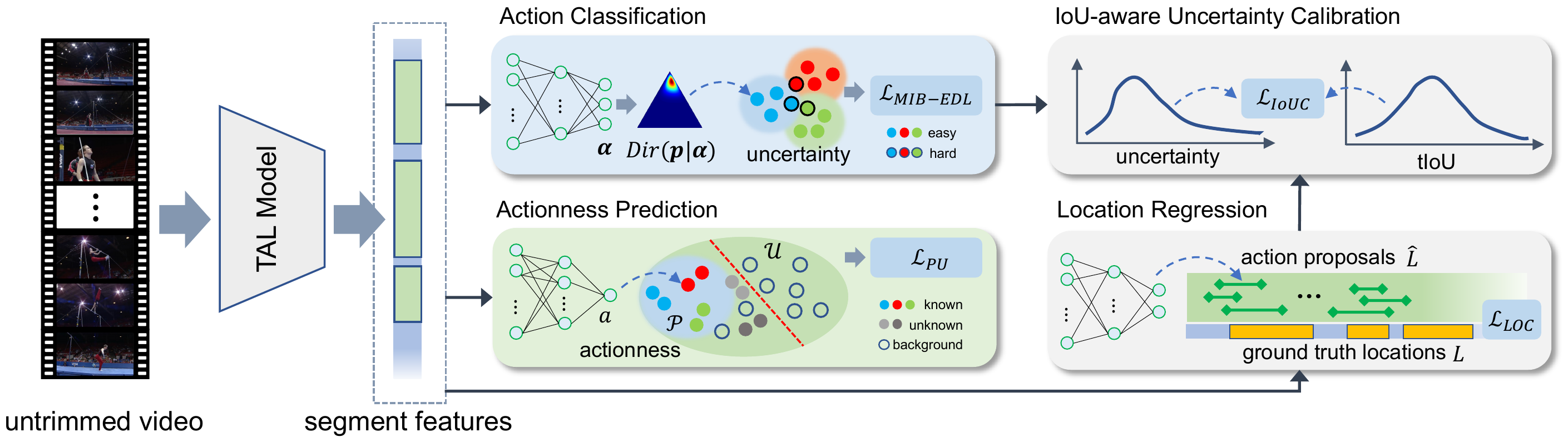}
    \caption{\textbf{The Proposed OpenTAL.} Given untrimmed videos as input, the OpenTAL method is developed on existing TAL models (such as AFSD~\cite{AFSD_CVPR2021}) toward the OSTAL scenario. It consists of action classification, actionness prediction, and location regression, which are learned by the proposed MIB-EDL loss (Eq.~\eqref{eq:ghedl}), PU learning (Eq.~\eqref{eq:act}), and localization loss (Eq.~\eqref{eq:loc}), respectively. Furthermore, the IoU-aware uncertainty calibration is proposed to calibrate the uncertainty estimation by considering localization quality (Eq.~\eqref{eq:iouc}). In inference, with a two-step decision procedure by leveraging the uncertainty and actionness, video actions from the known and unknown classes, as well as background frames can be distinguished in the OSTAL setting (see Algorithm~\ref{alg:test}).
    }
    \label{fig:opental}
\end{figure*}

\paragraph{Open Set Recognition} Open set recognition (OSR) aims to recognize known classes and reject the unknown. The pioneering work by Scheirer~\etal~\cite{ScheirerTPAMI2012} formalized the definition of OSR and introduced an ``one-vs-set" machine based on binary SVM, which inspired a line of SVM-based OSR methods~\cite{ScheirerTPAMI2014,JainECCV2014,junior2016specialized}. Benefited by the deep neural networks (DNNs), Bendale~\etal~\cite{BendaleCVPR2016} proposed the first DNN-based OSR method OpenMax, which leverages Extreme Value Theory (EVT) to expand the $K$-class softmax classifier. Recently, Fang~\etal~\cite{fangICML2021} theoretically proved the learnability of OSR classifier and the generalization bound. Existing generative OSR methods~\cite{GeBMVC2017,DitriaACCV2020,PereraCVPR2020,kong2021opengan,chenPAMI2021,zhou2021learning,Yue_2021_CVPR} utilize GAN~\cite{GoodfellowNIPS2014}, generative causal model, or mixup augmentation to generate the samples of the unknown. From the reconstruction perspective, some literature~\cite{yoshihashiCVPR2019,OzaCVPR2019,SunCVPR2020} leverage VAE~\cite{kingmaICLR2014} or self-supervised learning to reconstruct the representation of known class data to identify the unknown. Prototype learning and metric learning methods~\cite{YangCVPR2018,shu2020p,YangTPAMI2020,chenECCV2020,chenPAMI2021,zhangICCV2021,cenICCV2021} aim to identify the unknown by producing large distance to the prototype of known class data. Recently, uncertainty estimation methods~\cite{mundt2019open,BaoICCV2021,wangICCV2021} by probabilistic and evidential deep learning show promising results on OSR problems.

In this paper, we step further toward the OSTAL problem. We are aware of analogous  extensions from OSR to open set object detection~\cite{miller2018dropout,DhamijaWACV2020,joseph2021towards} and segmentation~\cite{pham2018bayesian,oliveira2021fully,WangICCV2021Uniden,hwang2021exemplar}. However, it is the uniqueness of the localization in open-world that makes the OSTAL problem even more challenging and valuable in practice.

\section{Proposed Method}

\paragraph{Setup} Given an untrimmed video, the OSTAL task requires a model to localize all actions with temporal locations $l_i=(s_i, e_i)$, assign the actions with labels $y_i\in \{0,1,\ldots,K\}$ where $y_i=0$ indicates the action consisting of background frames, and reject the actions from novel classes as the unknown. In the training, the model only has access to the video data and the annotations of known actions, while the annotations of unknown actions are not given. This setting is different from the OSR problem where both annotations and data of unknown classes are not given, because it is impractical in the TAL task to discard video segments of unknown actions.

\vspace{-2mm}
\paragraph{Overview} Fig.~\ref{fig:opental} shows an overview of the proposed OpenTAL. Given an untrimmed video, the features of action proposals are obtained from an existing TAL model such as the AFSD~\cite{AFSD_CVPR2021}. To fulfill OSTAL, we decouple the objective into three sub-tasks by a trident head, including action classification, actionness prediction, and location regression. The three branches are learned by multi-task loss functions, which will be introduced in detail. 

\vspace{-2mm}
\paragraph{Motivations}Existing TAL models typically adopt a $(K+1)$-way action classification by assigning the background video frames with the $(K+1)$-th class \textit{Background}. However, this paradigm is unable to handle the OSTAL case when unknown actions exist in the \textit{Background} class. 

To solve this problem, on one hand, one would attempt to append the $K$ known classes with an additional \textit{Unknown} category in an existing TAL system. However, this solution is practically infeasible under the OSTAL setting, because finding the video segments to train a classifier with the class \textit{Unknown} relies on the temporal boundary annotations of unknown actions, which are not available under our OSTAL setting. Though one could relax the OSTAL setting by providing temporal annotations of the unknowns in training, learning a $(K+1)$-way classifier is nontrivial due to the vague semantics of the \textit{Unknown}, and this relaxation has little practical significance in an open-world where we have nothing about the prior knowledge of unknown actions. 
On the other hand, one may remove the \textit{Unknown} or the \textit{Background} class from training data, which are both infeasible under the OSTAL setting because (i) we have no temporal annotations of the unknown actions to remove them, and (ii) the pure background frames provide indispensable temporal context for action localization. Therefore, in contrast to the OSR problem, an unique technical challenge of OSTAL lies in distinguishing between actions of known and unknown classes, as well as the background frames.  

Moreover, since the unknown actions are mixed with background frames without annotations, learning to distinguish foreground actions essentially reduces to a semi-supervised OSR problem~\cite{yu2020multi,saito2021NIPS}, that the model is trained with the labeled ``known known" actions and the unlabeled ``known unknown" actions while testing with data containing the ``unknown unknown" actions\footnote{Refer to~\cite{GengTPAMI2020,DhamijaWACV2020} for more detailed discussions on these terminologies.}.

To tackle these unique challenges, we propose to decouple the $(K+1)$-way action classification into $K$-way uncertainty-aware classification (Sec.~\ref{sec:cls}) and actionness prediction (Sec.~\ref{sec:act}). Thus, we could address the first challenge above by jointly leveraging the uncertainty and actionness in a two-level decision-making (see Table~\ref{tab:setting}) and the second challenge by the PU learning (Sec.~\ref{sec:act}).

\subsection{Action Classification}
\label{sec:cls}

\paragraph{$K$-way Uncertainty-aware Classification}
Following the existing Evidential Deep Learning (EDL)~\cite{SensoyNIPS2018,BaoICCV2021}, which is efficient to quantify the classification uncertainty, we assume a Dirichlet distribution $\text{Dir}(\mathbf{p}|\boldsymbol{\alpha})$ over the categorical probability $\mathbf{p}\in \mathbb{R}^K$, where $\boldsymbol{\alpha}\in \mathbb{R}^{K}$ is the Dirichlet strength. The EDL aims to directly predict $\boldsymbol{\alpha}$ by deep neural networks (DNNs). The model is trained by minimizing the following negative log-likelihood of data $\{x_i,y_i\}$:
\begin{equation}
    \mathcal{L}_{\text{EDL}}^{(i)}(\boldsymbol{\alpha}_i) = \sum_{j=1}^K t_{ij}(\log(S_{i}) - \log(\alpha_{ij})),
\label{eq:edl}
\end{equation}
where $t_{ij}$ is a binary element of the one-hot form of label $y_i$, and $t_{ij}=1$ only when $y_i=j$, and $S_i=\sum_j \alpha_{ij}$ is the total strength over $K$ classes. 

In testing, given the sample $x_i^*$, the action classification branch (DNN) produces non-negative evidence output $\mathbf{e}_i\in \mathbb{R}_+^{K}$. Then, the expectation of the classification probability is obtained by $\mathbb{E}[\mathbf{p}_i]=\boldsymbol{\alpha}_i / S_i$ where $\boldsymbol{\alpha}_i = \mathbf{e}_i + 1$ according to the evidence theory~\cite{SentzBook2002} and subjective logic~\cite{JosangBook2016}. And the classification uncertainty is estimated by $u_i=K/S_i$.

\begin{table}[t]
    \centering
    \captionsetup{font=small,aboveskip=3pt}
    \caption{\textbf{Our Motivations for the OSTAL.} The notations $\downarrow$ and $\uparrow$ denote small and large values, repsectively.}
    \label{tab:setting}
    \small
    \setlength{\tabcolsep}{0.8mm}
    \setlength{\extrarowheight}{0.5mm}
    \begin{tabular}{c|c|c|c}
    \toprule
         & Known Action & Unknown Action & Background\\
    \midrule
         uncertainty ($u$) & $\downarrow$ & $\uparrow$ &  $\uparrow$ \\
         actionness ($a$) & $\uparrow$ & $\uparrow$ & $\downarrow$\\
    \bottomrule
    \end{tabular}
\end{table}

However, the above EDL method is empirically found ineffective in the OSTAL task since Eq.~\eqref{eq:edl} gives equal consideration to each sample, which is practically not the case in OSTAL. In this paper, we propose to improve the generalization capability of EDL by encouraging the model to focus more on important samples in a principled way.

\vspace{-2mm}
\paragraph{Momentum Importance-Balanced EDL}  Inspired by the recent advances in imbalanced visual classification~\cite{koh2017understanding,Park_2021_ICCV}, the sample importance can be measured by the influence function which is determined by the gradient norm.
Specifically, let $\mathbf{h}_i\in \mathbb{R}^D$ be the feature input of the last DNN layer, an exponential evidence function is applied to predict the evidence, i.e., $\mathbf{e}_i \triangleq \exp (\mathbf{w}^T \mathbf{h}_i)$ where $\mathbf{w}\in \mathbb{R}^{D\times K}$ are the learnable weights of the DNN layer. The gradient $\mathbf{g}_{i}$ of the EDL loss $\mathcal{L}_{\text{EDL}}^{(i)}$ w.r.t. the logits $\mathbf{z}_i\triangleq \mathbf{w}^T \mathbf{h}_i$ is derived:
\begin{equation}
    g_{ij} = \frac{\partial \mathcal{L}_{\text{EDL}}^{(ij)}}{\partial z_{ij}}
    = t_{ij}\left[ \frac{S_i - K\alpha_{ij}}{S_i\alpha_{ij}}\right]
    = t_{ij}\left[\frac{1}{\alpha_{ij}} - u_i \right],
\label{eq:grad}
\end{equation}
where the chain rule and the equality $u_i=K/S_i$ are used. Since $t_{ij}=0$ when $j\neq y_i$, it is interesting to see a simple but meaningful gradient form, i.e., $g_{ik}=1/\alpha_{ik} - u_i$ where $k=y_i$, and in our supplement we proved that $|g_{ik}| \in [0,1)$. 

Furthermore, inspired by~\cite{Park_2021_ICCV}, we consider the influence function given by the gradient norm of EDL loss w.r.t. the network parameters $\mathbf{w}$. According to the chain rule of $\mathbf{z}_i = \mathbf{w}^T \mathbf{h}_i$, the influence value $\omega_{i} $ can be derived: 
\begin{equation}
    \omega_{i} = \left(\sum_{k=1}^{K}|g_{ik}|\right)\left(\sum_{d=1}^{D}|h_{id}|\right) = \|\mathbf{g}_{i}\|_1\cdot\|\mathbf{h}_i\|_1.
\label{eq:omega}
\end{equation}
Detailed proof can be found in the supplement. We define the loss weight of sample $x_i$ as the moving mean of influence values within the neighboring region of $\|\mathbf{g}_i\|_1$:
\begin{equation}
    \tilde{\omega}_i^{(t)} = \epsilon \cdot \tilde{\omega}_i^{(t-1)} + (1-\epsilon)\cdot \frac{1}{|\Omega_m|} \sum \Omega_m,
\label{eq:mm_omega}
\end{equation}
where $\Omega_m$ is a subset of $\omega_i$ whose gradient norm $\|\mathbf{g}_i\|_1$ falls into the $m$-th bin out of total $M$ bins in the region $[0,1]$, i.e., $\Omega_m = \{\omega_i | \|\mathbf{g}_i\|_1 \in [\frac{m-1}{M}, \frac{m}{M}], m=1,\ldots,M\}$. The $\epsilon$ is a momentum factor within $[0,1]$, $M$ is a constant, and $t$ is the training iteration. We set the initial weight $\tilde{\omega}_i^{(0)}$ as the 1.0. A larger $\epsilon$ means the set of influence values $\omega_i$ are less considered, while $M$ controls the granularity of the neighborhood of the gradient norm. Eventually, the proposed Momentum Importance-Balanced (MIB) EDL loss is defined as:
\begin{equation}
    \mathcal{L}_{\text{MIB-EDL}} = \frac{1}{N} \sum_{i=1}^N \tilde{\omega}_i^{(t)}  \mathcal{L}_{\text{EDL}}^{(i)}(\boldsymbol{\alpha}_i).
\label{eq:ghedl}
\end{equation}

The proposed MIB-EDL loss encourages the model to smoothly focus on important samples as the training iteration increases. In practice, to stabilize the training, the re-weighting is applied after $T_0$ training iterations. Different to~\cite{Park_2021_ICCV} that uses the inverse of $\omega_i$ to down-weight the influential samples for a balanced closed-set recognition, we use Eq.~\eqref{eq:omega} to up-weight these samples for open-set recognition, and~\eqref{eq:mm_omega} to achieve a smooth update of the sample weight.

\subsection{Actionness Prediction} 
\label{sec:act}

Due to the mixture of unknown actions and pure background frames, it is not sufficient to distinguish between them by the evidential uncertainty over $K$ known classes. Therefore, predicting the actionness that indicates how likely a sample is a foreground action is critical. We notice the fact that data from known classes are \emph{positive} data while the samples from the ``background" mixture are \emph{unlabeled}. This intrinsically reduces to a semi-supervised learning problem called positive-unlabeled (PU) learning~\cite{BekkerML2020}. In this paper, we propose a simple yet effective PU learning method to predict the actionness.

Let $\hat{a}_i \in [0,1]$ be the predicted actionness score of the sample $x_i$, the actionness in a training batch $\hat{\mathcal{A}}=\{\hat{a}_i\}$ can be splitted into the positive set $\hat{\mathcal{P}}=\{\hat{a}_i|y_i\geq 1\}$ and the unlabeled background set $\hat{\mathcal{U}}=\{\hat{a}_i|y_i = 0\}$. In this paper, we propose to ascendingly sort the $\hat{\mathcal{U}}$ and select top-$M$ samples to form the most likely negative set $\hat{\mathcal{N}}=\{\hat{a}_i | \hat{a}_i \in sort(\hat{\mathcal{U}})_{1,\ldots,M}$\}. Then, a binary cross-entropy (BCE) loss could be applied to the $\hat{\mathcal{P}}$ and $\hat{\mathcal{N}}$:
\begin{equation}
    \mathcal{L}_{\text{ACT}}(\hat{\mathcal{P}}, \hat{\mathcal{N}}) =- \frac{1}{|\hat{\mathcal{P}}|}\sum_{\hat{a}_i\in \hat{\mathcal{P}}} \log\hat{a}_i - \frac{1}{|\hat{\mathcal{N}}|}\sum_{\hat{a}_i\in \hat{\mathcal{N}}}\log(1-\hat{a}_i).
\label{eq:act}
\end{equation}
Here, to achieve a balanced BCE training, we set the size of negative set to $M=|\hat{\mathcal{N}}|\defeq \min (|\hat{\mathcal{P}}|, |\hat{\mathcal{U}}|)$ considering that in most training batches we have $|\hat{\mathcal{U}}| \gg |\hat{\mathcal{P}}|$. This BCE loss will push the probably pure background samples far away from positive actions. Though this method is straightforward, the learned actionness scores are found discriminative enough to distinguish between the foreground actions and background frames in the OSTAL setting (see Fig.~\ref{fig:dist_a}).

\subsection{Location Regression}

To maintain the flexibility of our method on existing TAL models, the temporal location regression follows the design of the TAL models. Take the state-of-the-art TAL model AFSD~\cite{AFSD_CVPR2021} as an example, it consists of a coarse stage to predict the location proposals $\hat{l}_i=[\hat{s}_i,\hat{e}_i]$ and a refined stage to predict the temporal offset $\hat{\delta}_i=[\hat{\delta}_i^{(s)},\hat{\delta}_i^{(e)}]$ with respect to the $\hat{l}_i$. The coarse stage is learned by temporal Intersection-over-Union (tIoU) loss, while the refined stage is learned by an $L_1$ loss:
\begin{equation}
\left\{
\begin{aligned}
    & \mathcal{L}_{\text{LOC}}(\{\hat{l_i}\}) = \frac{1}{N_C} \sum_i \mathbb{I}[y_i \geq 1]\left(1-\frac{|\hat{l}_i \cap l_i|}{|\hat{l}_i \cup l_i|}\right) \\
    & \mathcal{L}_{\text{LOC}}(\{\hat{\delta_i}\}) = \frac{1}{N_R} \sum_i \mathbb{I}[y_i \geq 1](|\hat{\delta}_i - \delta_i|),
\end{aligned}
\right.
\label{eq:loc}
\end{equation}
where $N_C$ and $N_R$ are corresponding number of samples that are matched with the ground truth action locations by an IoU threshold. The indicator function $\mathbb{I}[y_i\geq 1]$ filters out the unmatched samples which are treated as the ``background" data. In testing, the predicted location is recovered by $l^*_i = [\hat{s}_i + 0.5(\hat{e}_i-\hat{s}_i)\hat{\delta}_i^{(s)}, \hat{e}_i + 0.5(\hat{e}_i-\hat{s}_i)\hat{\delta}_i^{(e)}]$. Note that our OpenTAL framework is not limited to specific TAL models but general in design.

\subsection{IoU-aware Uncertainty Calibration}

Though the loss functions defined by Eqs.~\eqref{eq:ghedl}\eqref{eq:act}\eqref{eq:loc} are sufficient for a complete OSTAL task, the learned uncertainty in the classification module is not calibrated by considering the localization performance. Intuitively, an action proposal of high temporal overlap with the ground truth location should contain more evidence and thus low uncertainty. To this end, we propose a novel IoU-aware uncertainty calibration method:
\begin{equation}
    \mathcal{L}^{(i)}_{\text{IoUC}}(\hat{l}_i, u_i) = -w_{\hat{l}_i, l_i}\log (1-u_i) - (1-w_{\hat{l}_i, l_i})\log(u_i)
\label{eq:iouc}
\end{equation}
where the weight $w$ is a clipped form of the temporal IoU between the predicted and ground truth locations:
\begin{equation}
    w_{\hat{l}_i, l_i} = \max\left(\gamma, \text{IoU}(\hat{l}_i, l_i)\right)
\label{eq:gamma}
\end{equation}
where the $\gamma$ is a small non-negative constant. The cross-entropy form in Eq.~\eqref{eq:iouc} and~\eqref{eq:gamma} will encourage the model to produce high uncertainty ($u_i\!\rightarrow\! 1$) for action proposals with low localization quality ($w\!\rightarrow\! \gamma$).

The motivation behind the clipping by $\max ()$ is that, given with the ground truth of known actions, both the proposals of background frames and unknown actions are not overlapped with the ground truth such that $\text{IoU}(\hat{l}_i, l_i)\leq 0$, the clipping could avoid reversing the loss value from positive to negative, while still maintaining a low localization quality $\gamma$. Besides, it is reasonable to encourage high uncertainty $u_i$ by small $\gamma$ for the location proposals of the background and the unknown actions in the OSTAL setting.

\input{alg}

\subsection{Training and Inference}

The training procedure is to minimize the weighted sum of losses defined by Eqs.~\eqref{eq:ghedl}\eqref{eq:act}\eqref{eq:loc}\eqref{eq:iouc}:
\begin{equation}
    \mathcal{L} = \mu \mathcal{L}_{\text{MIB-EDL}} + \mathcal{L}_{\text{ACT}} + \mathcal{L}_{\text{LOC}} + \mathbb{E}[\mathcal{L}^{(i)}_{\text{IoUC}}],
\label{eq:total}
\end{equation}
where $\mu$ is a hyperparameter, and $\mathbb{E}[\cdot]$ is to take the mean loss values over the input samples.

In inference, the untrimmed video input is fed into a TAL model, and our OpenTAL method trained on the TAL model could produce multiple action locations $\{l_i^*\}$, classification labels $\hat{y}_i=\arg\max_{j\in [1,\ldots,K]} \mathbb{E}[\mathbf{p}_{ij}]$, classification uncertainty $u_i$, and actionness score $\hat{a}_i$. Together with the $u_i$ and $\hat{a}_i$, a positively localized foreground action $x_i$, i.e., $a_i > 0.5$, can be accepted as known class $\hat{y}_i$, or rejected as the unknown by the following simple scoring function:
\begin{equation}
    P(x_i|a_i>0.5) = \left\{ 
    \begin{aligned}
    & unknown, & \text{if} \; u_i > \tau, \\
    & \hat{y}_i, & \text{otherwise}.
    \end{aligned}
    \right.
\label{eq:rule}
\end{equation}
The complete inference procedure is shown in Algorithm~\ref{alg:test}. In addition to this two-level decision, one-level decisions by the functional formulas of $P(x_i)$ w.r.t. to $u_i$ and $a_i$ are also plausible (see Table~\ref{tab:oodscore}). However, we empirically found that Eq.~\eqref{eq:rule} is the most effective formula while maintaining the explainable nature of decision-making.

\section{Experiment}

\subsection{Implementation Details} 
Our method is implemented on the AFSD~\cite{AFSD_CVPR2021} model\footnote{https://github.com/TencentYoutuResearch/ActionDetection-AFSD}, which is a state-of-the-art TAL model. Pre-trained I3D~\cite{I3DCVPR2017} backbone is used in AFSD. The proposed OpenTAL is applied to both the coarse and refined stages of AFSD. Specifically, the proposed MIB re-weighting is applied after $10$ training epochs. We empirically set the momentum $\epsilon$ to $0.99$ and the number of bins $M$ to $50$. The small constant $\gamma$ in Eq.~\eqref{eq:gamma} is set to $0.001$. The loss weight $\mu$ in Eq.~\eqref{eq:total} is set to 10. We trained the model $25$ epochs to ensure full convergence. The rest settings in AFSD are kept unchanged. 

\subsection{Datasets} THUMOS14~\cite{THUMOS14} and ActivityNet1.3~\cite{ANetCVPR2015} are two commonly-used datasets for TAL evaluation. THUMOS14 dataset contains 200 training videos and 212 testing videos. ActivityNet1.3 dataset contains about 20K videos with 200 human activity categories. Since our method is not limited by data modality, we use RGB videos for training and testing by default. To enable OSTAL evaluation, we randomly select 3/4 THUMOS14 categories of the training videos as the known data. This random selection is repeated to generate three THUMOS14 open set splits between the known and the unknown. Considering that ActivityNet1.3 is newer and covers most THUMOS14 categories, the ActivityNet1.3 is not suitable to be the closed-set training data when the model is tested on THUMOS14. Therefore, we train models on the THUMOS14 known split, and use the THUMOS14 unknown split and the disjoint categories of ActivityNet1.3 as two sources open set testing data. To get the disjoint categories from ActivityNet1.3, we manually removed 14 semantically overlapping categories by referring to the THUMOS14 categories. Detailed dataset information could be found in our supplement. 

\subsection{Evaluation Protocols}

The mean Average Precision (\textbf{mAP}) is typically used for the evaluation of closed set TAL performance. To enable OSTAL performance evaluation, the Area Under the Receiver Operating Characteristic (\textbf{AUROC}) curve and the Area Under the Precision-Recall (\textbf{AUPR}) are introduced to evaluate the performance of detecting the unknown from the known actions for positively localized actions. To address the operational meaning in practice, we additionally report the False Alarm Rate at True Positive Rate of 95\% (\textbf{FAR@95}), by which smaller value indicates better performance. However, we noticed the metrics above are insufficient for the OSTAL task because the multi-class classification performance of the known classes in the OSTAL setting is ignored. Inspired by the Open Set Classification Rate~\cite{dhamija2018reducing,neal2018open,chenPAMI2021}, we propose the Open Set Detection Rate (\textbf{OSDR}), which is defined as the area under the curve of Correct Detection Rate (CDR) and False Positive Rate (FPR). Given an operation point $\tau$ of the scoring function $P(x)$ for detecting the unknown and an operation point $t_0$ of the tIoU for localizing the foreground actions, CDR and FPR are defined as:
\begin{equation}
\left\{
\begin{aligned}
    & \text{CDR} (\tau, t_0) = \frac{|\{x|(x \in \mathcal{F}_k) \wedge (\hat{f}_{x|y}=y) \wedge P(x) < \tau \}|}{| \mathcal{F}_k|} \\
    & \text{FPR}(\tau, t_0) = \frac{|\{x|(x \in \mathcal{F}_u) \wedge P(x) < \tau\}|}{|\mathcal{F}_u|}
\end{aligned}
\right.
\end{equation}
where $\mathcal{F}_k$ is the set of positively localized known actions, i.e., $\mathcal{F}_k = \{x|(\text{tIoU} > t_0)\wedge (y\in [1,\ldots,K]) \}$, and $\mathcal{F}_u$ is the set of positively localized unknown actions, i.e., $\mathcal{F}_u = \{x|(\text{tIoU} > t_0)\wedge (y=0)\}$. The CDR indicates the fraction of known actions that are positively localized and correctly classified into their known classes, while the FPR denotes the fraction of unknown actions that are positively localized but falsely accepted as an arbitrary known class. Higher OSDR indicates better performance for the OSTAL task.

For stable evaluation, all results are reported by averaging the results of each evaluation metric over the three THUMOS14 splits. Results are reported at tIoU threshold 0.3 for THUMOS14 and 0.5 for ActivityNet1.4, and results by other thresholds are in the supplement.

\begin{table*}[t]
\centering
\captionsetup{font=small,aboveskip=3pt}
\caption{\textbf{OSTAL Results (\%).} Models trained on the THUMOS14 closed set are tested on the open sets by including the unknown classes from THUMOS14 and ActivityNet1.3, respectively. The mAP is provided as the reference of the TAL results on THUMOS14 closed set.}
\label{tab:thumos}
\small
\setlength{\tabcolsep}{3.1mm}
\setlength{\extrarowheight}{0.5mm}
\begin{tabular}{l|cccc|cccc|c}
\toprule
\multirow{2}{*}{Methods} &\multicolumn{4}{c|}{THUMOS14 as the Unknown} &\multicolumn{4}{c|}{ActivityNet1.3 as the Unknown} &\multirow{2}{*}{mAP} \\
\cline{2-9}
& FAR@95 ($\downarrow$) & AUROC & AUPR & OSDR  & FAR@95 ($\downarrow$) & AUROC & AUPR & OSDR & \\
\hline
SoftMax & 85.58 & 54.70 & 31.85 & 23.40 & 85.05 & 56.97 & 53.54 & 27.63 & 55.81 \\
OpenMax~\cite{BendaleCVPR2016} & 90.34 & 53.26 & 33.17 & 13.66 & 91.36 & 51.24 & 54.88 & 15.73 & 36.36 \\
EDL~\cite{BaoICCV2021} & 81.42 & 64.05 & 40.05 & 36.26 & 84.01 & 62.82 & 53.97 & 38.56  & 52.24 \\
\textbf{OpenTAL} & \textbf{70.96} & \textbf{78.33} & \textbf{58.62} & \textbf{42.91} & \textbf{63.11} & \textbf{82.97}  & \textbf{80.41} & \textbf{50.49} & 55.02 \\
\bottomrule
\end{tabular}
\end{table*}

\subsection{Comparison with State-of-the-arts}

The OpenTAL method is compared with the following baselines based on the AFSD: (1) \textbf{SoftMax}: use the softmax confidence score to identify the unknown. (2) \textbf{OpenMax}: use OpenMax~\cite{BendaleCVPR2016} in testing to append the softmax scores with unknown class. (3) \textbf{EDL}: similar to~\cite{BaoICCV2021}, vanilla EDL method is used to replace the traditional cross-entropy loss for uncertainty quantification. Models are tested using both the THUMOS14 unknown spits and the ActivityNet1.3 disjoint subset. Results are reported in Table~\ref{tab:thumos}.

The results show that the OpenTAL outperforms the baselines by large margins on all OSTAL metrics, while still keeping comparable closed set TAL performance (less than 1\% mAP decrease). The results also show that OpenMax does not work well on the OSTAL task, especially when the large-scale ActivityNet1.3 dataset is used as the unknown. The EDL works well but still far behind the proposed OpenTAL. 
Fig.~\ref{fig:curves} shows the detailed evaluation by the curves of AUROC and OSDR on one THUMOS14 split. Figures on other splits are in the supplement. They clearly show that the proposed OpenTAL on different operation points of scoring values and different open set splits is consistently better than the baselines. 

\input{curves}

\subsection{Ablation Study}

\paragraph{Component Ablation.} By individually removing the major components of OpenTAL, three model variants are compared. (1) Without MIB: the proposed MIB re-weighting is removed so that the vanilla EDL loss (Eq.~\eqref{eq:edl}) is used. (2) Without ACT: the actionness prediction is removed so that the $(K+1)$-way classification in  $\mathcal{L}_{\text{MIB-EDL}}$ (Eq.~\eqref{eq:ghedl}) is adopted. (3) Without IoUC: the loss $\mathcal{L}_{\text{IoUC}}$ (Eq.~\eqref{eq:iouc}) is removed from the training. Results are reported in Table~\ref{tab:ablation}. They show that OpenTAL achieves the best performance. Specifically, the MIB re-weighting strategy contributes the most to the OSDR performance gain by around 30\%. The actionness prediction (ACT) contributes the most to the FAR@95, AUROC, and AUPR metrics. Besides, the proposed IoUC loss also leads to significant performance gains on all metrics. These observations demonstrate the effectiveness of the three components for the OSTAL task.

\begin{table}
\centering
\captionsetup{font=small,aboveskip=3pt}
\caption{\textbf{Ablation Results (\%).} The proposed EDL re-weighting method (MIB), the actionness prediction (ACT), and the IoUC loss are individually ablated from the OpenTAL.}
\label{tab:ablation}
\small
\setlength{\tabcolsep}{0.4mm}
\setlength{\extrarowheight}{0.5mm}
\begin{tabular}{c|ccc|cccc}
\toprule
Variants & MIB & ACT & IoUC & FAR@95 ($\downarrow$) & AUROC & AUPR & OSDR \\
\hline
(1) & &\checkmark &\checkmark & 77.20 & 76.41 & 56.65 & 12.10 \\
(2) &\checkmark & &\checkmark & 82.85 & 58.12 & 31.80 & 37.89 \\
(3) &\checkmark &\checkmark & & 79.64 & 62.73 & 37.86 & 39.39 \\
\hline
OpenTAL &\checkmark &\checkmark &\checkmark & \textbf{70.96} & \textbf{78.33} & \textbf{58.62} & \textbf{42.91} \\
\bottomrule
\end{tabular}
\end{table}

\vspace{-4mm}
\paragraph{Choices of Re-weighting Methods.} We compare the proposed MIB re-weighting method (MIB (soft)) with the MIB (hard) and existing literature on sample re-weighting in Table~\ref{tab:reweight}. The results show that the focal loss (Focal)~\cite{LinICCV2017} does not work well with the OpenTAL framework. GHM~\cite{LiAAAI2019} and IB~\cite{Park_2021_ICCV} methods could achieve comparable FAR@95, AUROC, and AUPR performance, but their OSDR results are still largely far behind ours. Note that these methods are all designed for closed set recognition, thus the proposed MIB is more suitable for open set scenarios.  Besides, the hard version of MIB that the momentum mechanism is removed by setting the $\epsilon$ to 0, could improve about 4\% FAR@95 while sacrifice the AUROC, AUPR, and OSDR.

\begin{table}[t]
\centering
\captionsetup{font=small,aboveskip=3pt}
\caption{\textbf{Results of Different Re-weightings (\%).} MIB (hard) means the momentum factor $\epsilon=0$ in Eq.~\eqref{eq:mm_omega} such that the sample weight is updated in a hard manner, while the MIB (soft) sets the $\epsilon$ to 0.99 to enable a soft update, and wo. Re-weight means $\epsilon=1.0$.}
\label{tab:reweight}
\small
\setlength{\tabcolsep}{2mm}
\setlength{\extrarowheight}{0.5mm}
\begin{tabular}{l|cccc}
\toprule
Methods & FAR@95 ($\downarrow$) & AUROC & AUPR & OSDR \\
\hline
wo. Re-weight. & 77.20 & 76.41 & 56.65 & 12.10 \\
Focal~\cite{LinICCV2017} & 91.05 & 56.67  & 35.55  & 2.04 \\
GHM~\cite{LiAAAI2019} & 78.33 & 73.52 & 54.03 & 1.41 \\
IB~\cite{Park_2021_ICCV} & 80.23 & 75.91 & 58.00 & 2.18\\
\hline
MIB (hard) & \textbf{66.34} & 78.16 & 57.66 & 38.90 \\
MIB (soft) & 70.96 & \textbf{78.33} & \textbf{58.62} & \textbf{42.91} \\
\bottomrule
\end{tabular}
\end{table}

\vspace{-4mm}
\paragraph{Choices of Scoring Function.} The scoring function is critical to identify the known and unknown actions, as well as the background frames in model inference. In addition to the proposed two-level decision by~\eqref{eq:rule}, we compare it with four reasonable one-level decision methods by utilizing actionness $a_i$ and uncertainty $u_i$. The results in Table~\ref{tab:oodscore} show that using the maximum classification confidence (the 1st row) or other compositions of $u_i$ and $a_i$ (the 2nd and 3rd rows) cannot achieve favorable performance. The proposed method (the last row) is slightly better than the product between $u_i$ and $a_i$ (the 4-th row) with comparable FAR@95 performance. Though there are certainly other alternatives, our scoring function achieves the best performance while maintaining a good decision-making explanation, which means that the foreground actions are identified first by $a_i$, based on which the known and unknown actions are further distinguished by $u_i$. 

\begin{figure}[t]
    \centering
    \subcaptionbox{Actionness\label{fig:dist_a}}{
    \includegraphics[width=0.47\linewidth]{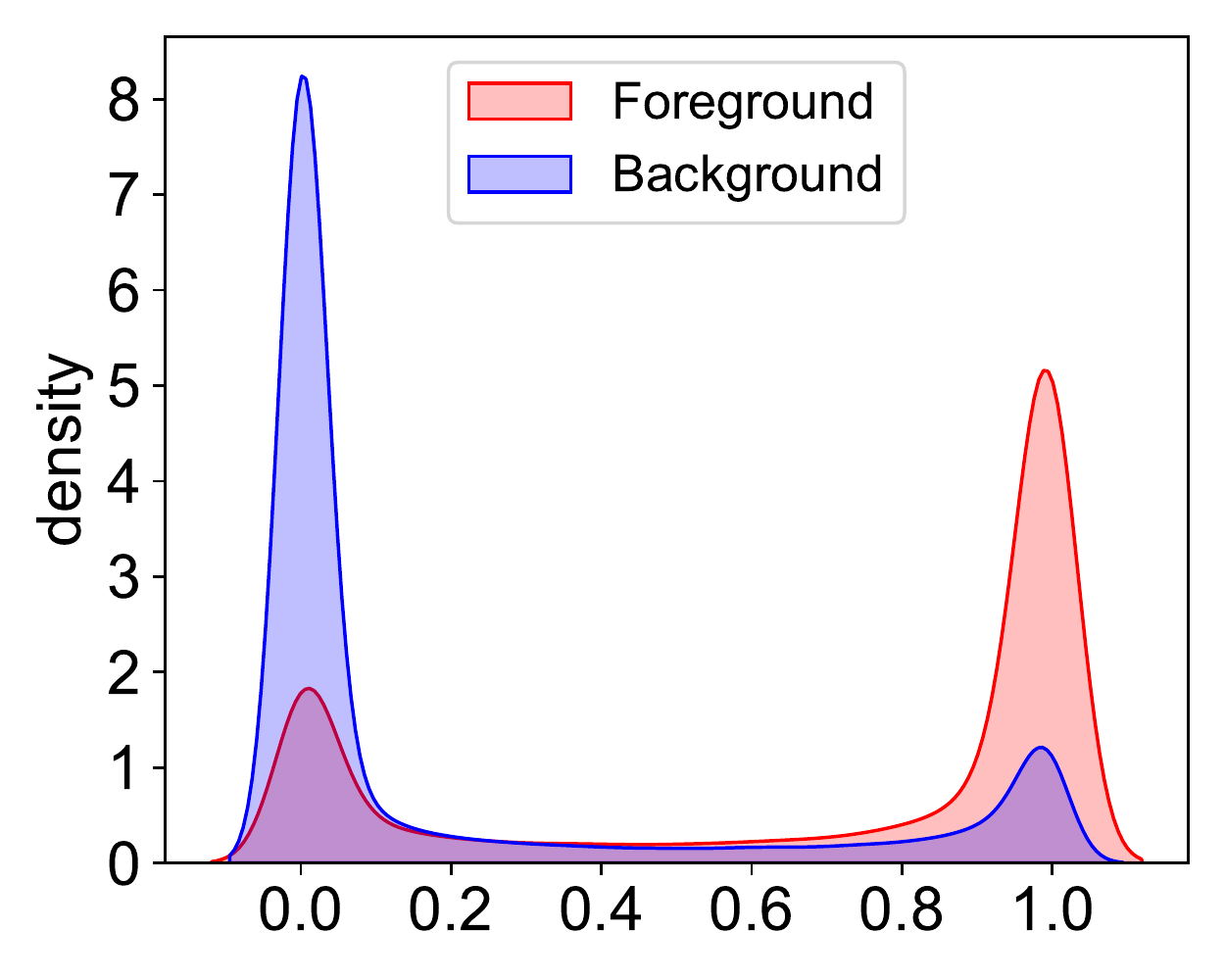}
    }
    \subcaptionbox{Uncertainty\label{fig:dist_u}}{
    \includegraphics[width=0.47\linewidth]{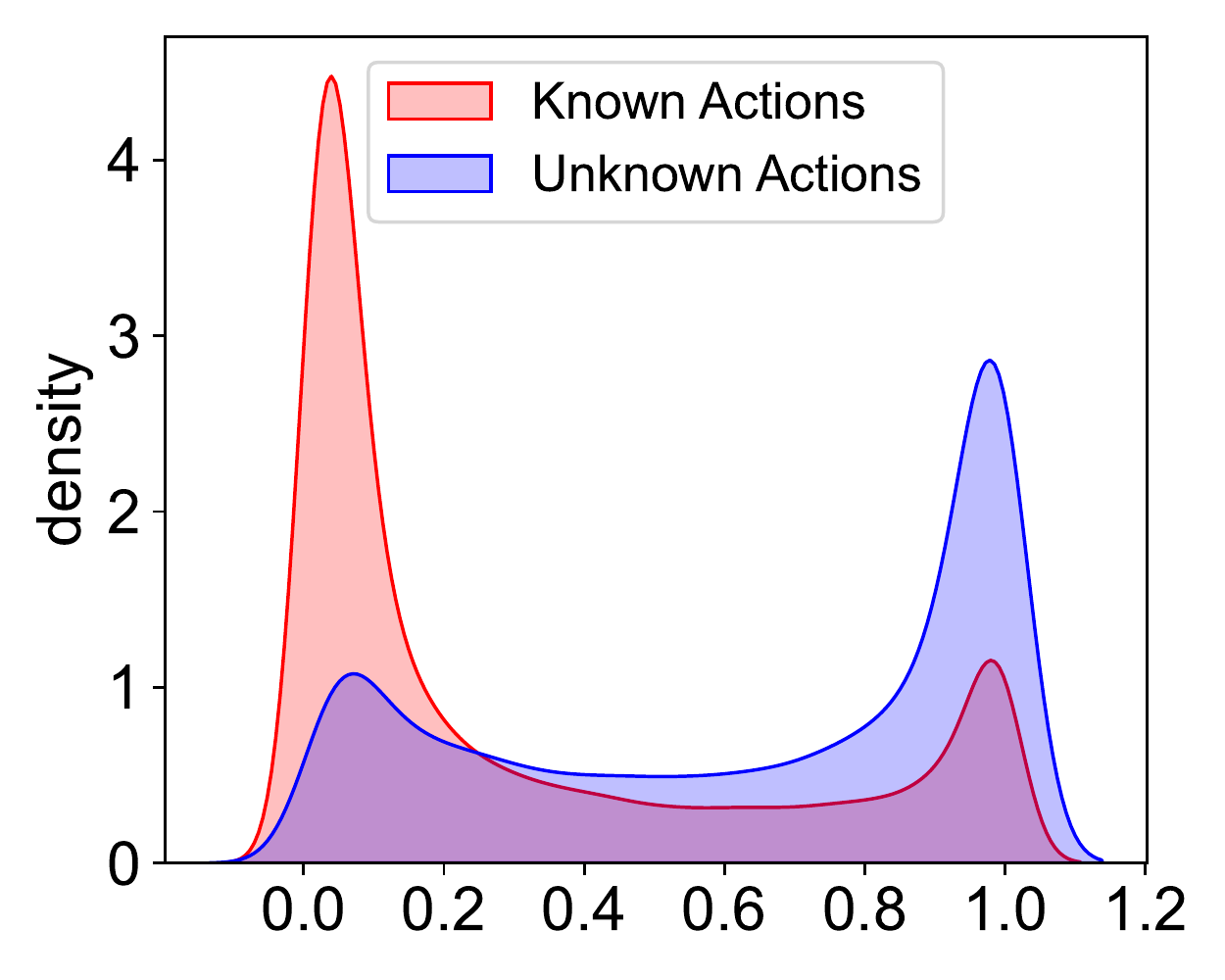}
    }
    \captionsetup{font=small,aboveskip=0pt}
    \caption{\textbf{Distributions of Actionness and Uncertainty.} The two figures show significant separation between the foreground actions and background frames by actionness score, as well as the separation between the known and unknown actions by uncertainty.}
    \label{fig:dist}
\end{figure}

\vspace{-4mm}
\paragraph{Distributions of Actionness and Uncertainty.} To show the quality of the learned actionness and uncertainty, we visualized their distributions on the test set in Fig.~\ref{fig:dist}. Specifically, the dominant modes in Fig.~\ref{fig:dist_a} show that foreground actions are majorly assigned with high actionness while the background frames are with low actionness, and the dominant modes in Fig~\ref{fig:dist_u} show that the actions of known classes are majorly assigned with low uncertainty while those of the unknowns are with high uncertainty. These observations align well with the expectation of our OpenTAL method.

\begin{table}[t]
\centering
\captionsetup{font=small,aboveskip=3pt}
\caption{\textbf{Scoring Functions.} It shows when conditioned on $a_i > 0.5$, uncertainty $u_i$ is the best scoring function for the OSTAL task.}
\label{tab:oodscore}
\small
\setlength{\tabcolsep}{0.5mm}
\setlength{\extrarowheight}{0.5mm}
\begin{tabular}{c|cccc}
\toprule
Scoring Functions & FAR@95 ($\downarrow$) & AUROC & AUPR & OSDR \\
\hline
\footnotesize $P(x_i)=1-\max_{j}(\boldsymbol{\alpha}_i/S_i)$   & 77.90 & 59.50 & 35.82 & 31.38 \\
\footnotesize $P(x_i)=u_i / (1-a_i)$   & 79.16 &  61.94 & 38.52  & 30.64 \\
\footnotesize $P(x_i)=a_i / (1-u_i)$   & 90.39 & 72.71 & 56.19 & 38.24 \\
\footnotesize $P(x_i)=u_i\cdot a_i$   & \textbf{70.64} & 77.52 & 58.17 & 42.44 \\
\hline
\footnotesize $P(x_i|a_i>0.5)=u_i$   & 70.96 & \textbf{78.33} & \textbf{58.62} & \textbf{42.91} \\
\bottomrule
\end{tabular}
\end{table}

\begin{figure}[t]
    \centering
    \includegraphics[width=\linewidth]{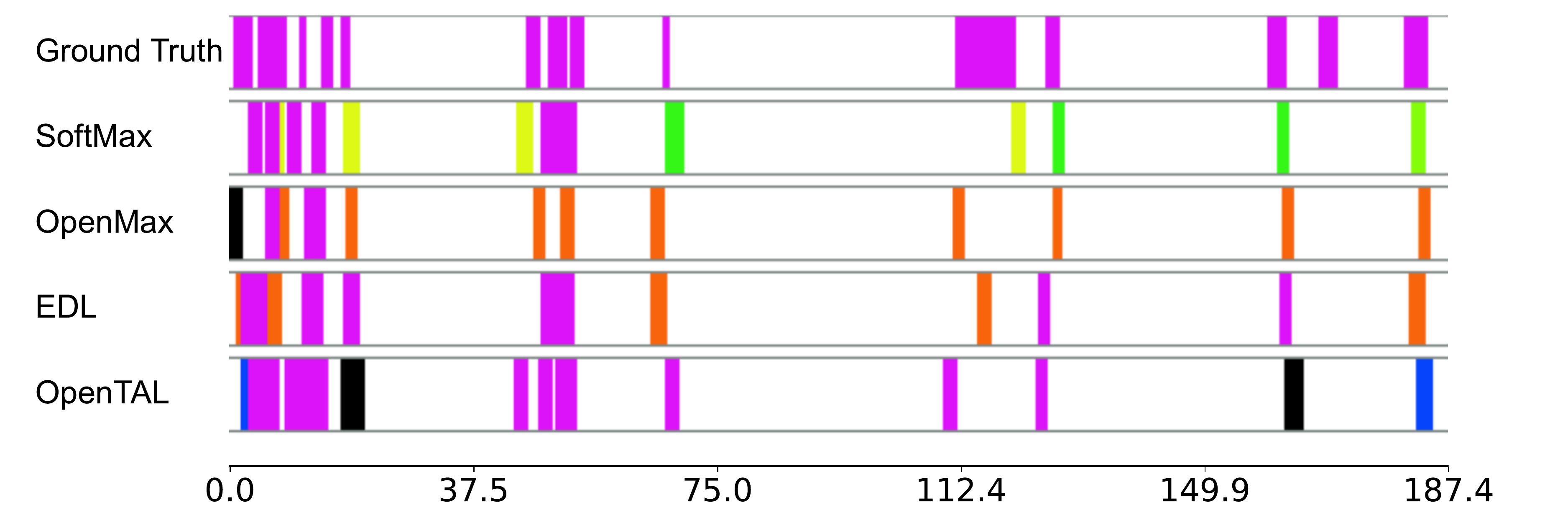}
    \includegraphics[width=\linewidth]{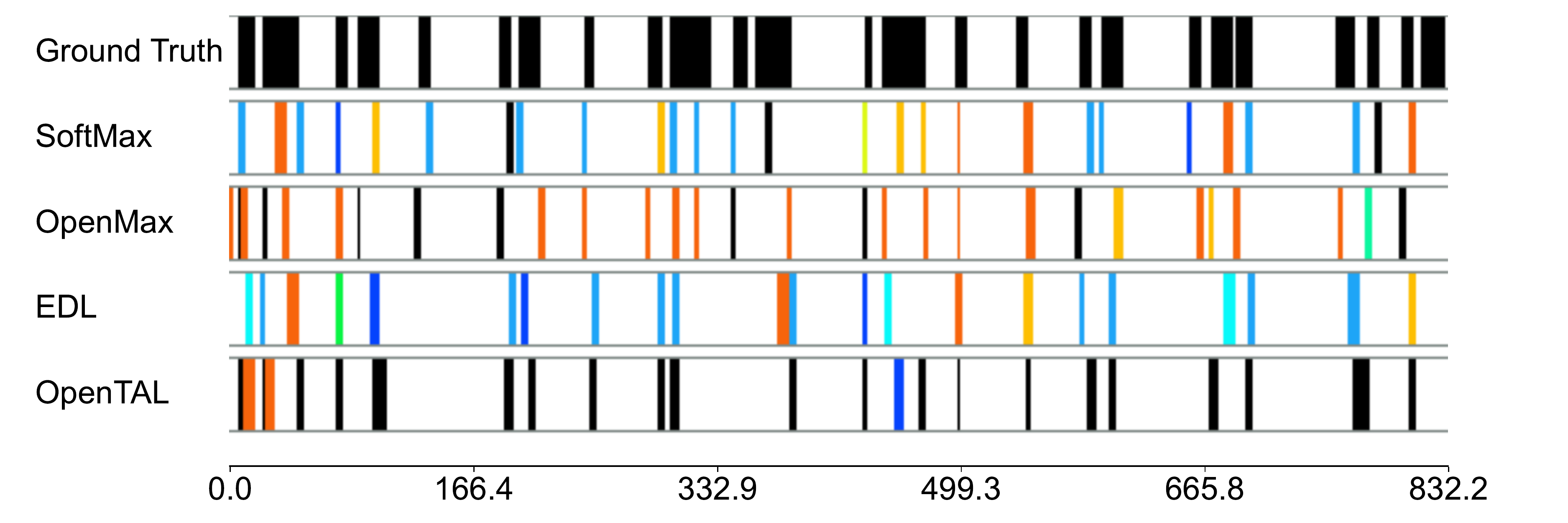}
    \includegraphics[width=\linewidth]{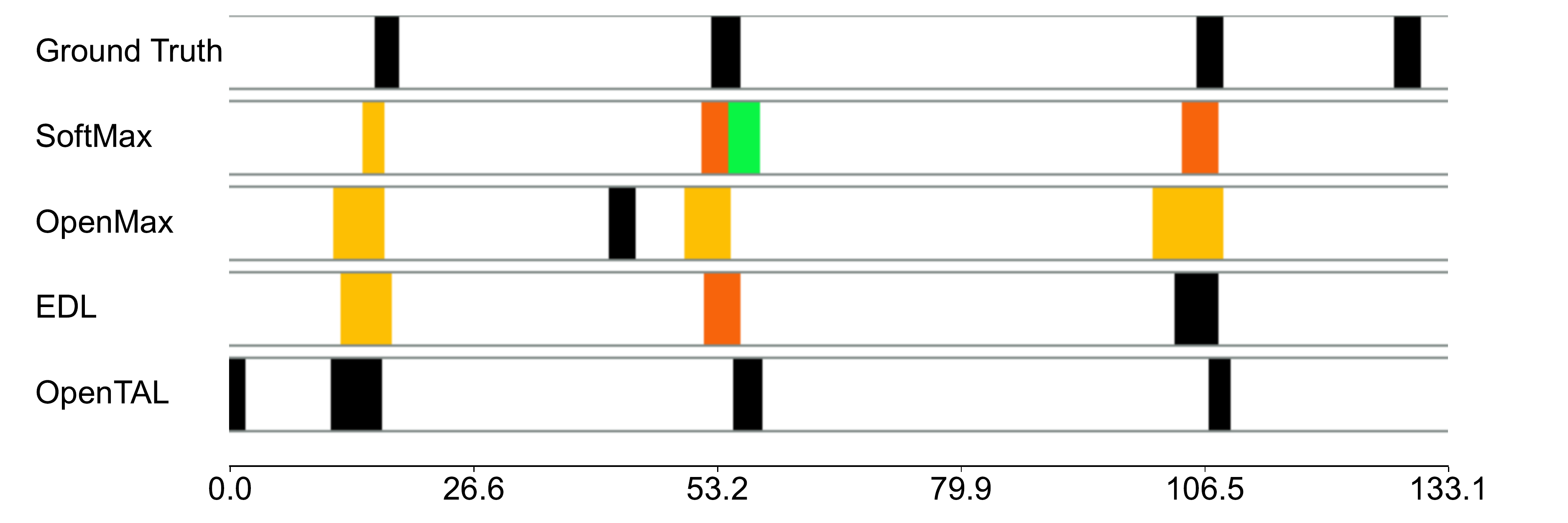}
    \caption{\textbf{Qualitative Results.} We show the actions of unknown classes with black color, while the rest colors are actions of known classes. The $x$-axis represents the timestamps (seconds).}
    \label{fig:vis_demo}
\end{figure}

\vspace{-4mm}
\paragraph{Qualitative Results.} Fig.~\ref{fig:vis_demo} shows the qualitative results of the proposed OpenTAL and baseline approaches. The three video samples are from the THUMOS14 dataset. The results clearly show that OpenTAL is superior to baselines in terms of both recognizing the known actions (colored segments in the 1st video), and rejecting the unknown actions (black segments in the 2nd and 3rd videos).

\vspace{2mm}
\noindent\textbf{Limitations.} We note that all those methods are not showing remarkable high OSDR performance, which indicates the challenging nature of the OSTAL task and there exists large room for improvement in the OpenTAL.

\section{Conclusion}

In this paper, we introduce the Open Set Temporal Action Localization (OSTAL) task. It aims to simultaneously localize and recognize human actions, and to reject the unknown actions from untrimmed videos in an open-world. The unique challenge lies in discriminating between known and unknown actions as well as background video frames. To this end, we propose a general OpenTAL framework to enable existing TAL models for the OSTAL task. 
The OpenTAL predicts the locations, classifications with uncertainties, and actionness to jointly achieve the goal. For comprehensive OSTAL evaluation, the Open Set Detection Rate is introduced. The OpenTAL is empirically demonstrated to be effective and significantly outperform existing baselines. 
We believe the generality of the OpenTAL design could inspire relevant research fields such as spatio-temporal action detection, video object detection, and video grounding toward open set scenarios.

\vspace{2mm}
\small\noindent\textbf{Acknowledgement}. This research is supported by Office of Naval Research (ONR) grant N00014-18-1-2875 and the Army Research Office (ARO) grant W911NF-21-1-0236. The views and conclusions contained in this document are those of the authors and should not be interpreted as representing the official policies, either expressed or implied, of the ONR, the ARO, or the U.S. Government. Besides, we sincerely thank Junwen Chen for TAL literature survey, data pre-processing, and methodology discussion. 

{\small
\bibliographystyle{ieee_fullname}
\bibliography{egbib}
}

\clearpage
\input{supp_v1}

\end{document}

%% file: alg.tex
\begin{algorithm}[tb]
\caption{Inference Procedure}
\label{alg:test}
\begin{algorithmic}[1]
\Require Untrimmed test video.
\Require Trained OpenTAL model.
\Require Threshold $\tau$ from training data by Eq.~\eqref{eq:rule}
\State Data pre-processing (if applicable).
\State Predict proposals $\mathcal{G}\!=\!\{l^*_i,\hat{y}_i,u_i,\hat{a}_i\}|_{i=1}^N$ by OpenTAL.
\State Post-processing (if applicable).
\For{each proposal $\mathcal{G}_i \in \mathcal{G}$}
    \If{$\hat{a}_i < 0.5$} \Comment{Background}
        \State $\mathcal{G}_i$ is a \textit{Background}; \textbf{continue}.
    \EndIf
    \If{$u_i > \tau$} \Comment{Unknown Action}
        \State  $\mathcal{G}_i$ is \textit{Unknown}. 
    \Else \Comment{Known Action}
        \State $\mathcal{G}_i$ is \textit{Known} by $\hat{y}_i=\arg\max_{j} \mathbb{E}[\mathbf{p}_{ij}]$. 
    \EndIf
\EndFor
\end{algorithmic}
\end{algorithm}

%% file: curves.tex
\begin{figure}[t]
    \centering
    \subcaptionbox{ROC Curves}{
        \includegraphics[width=0.47\linewidth]{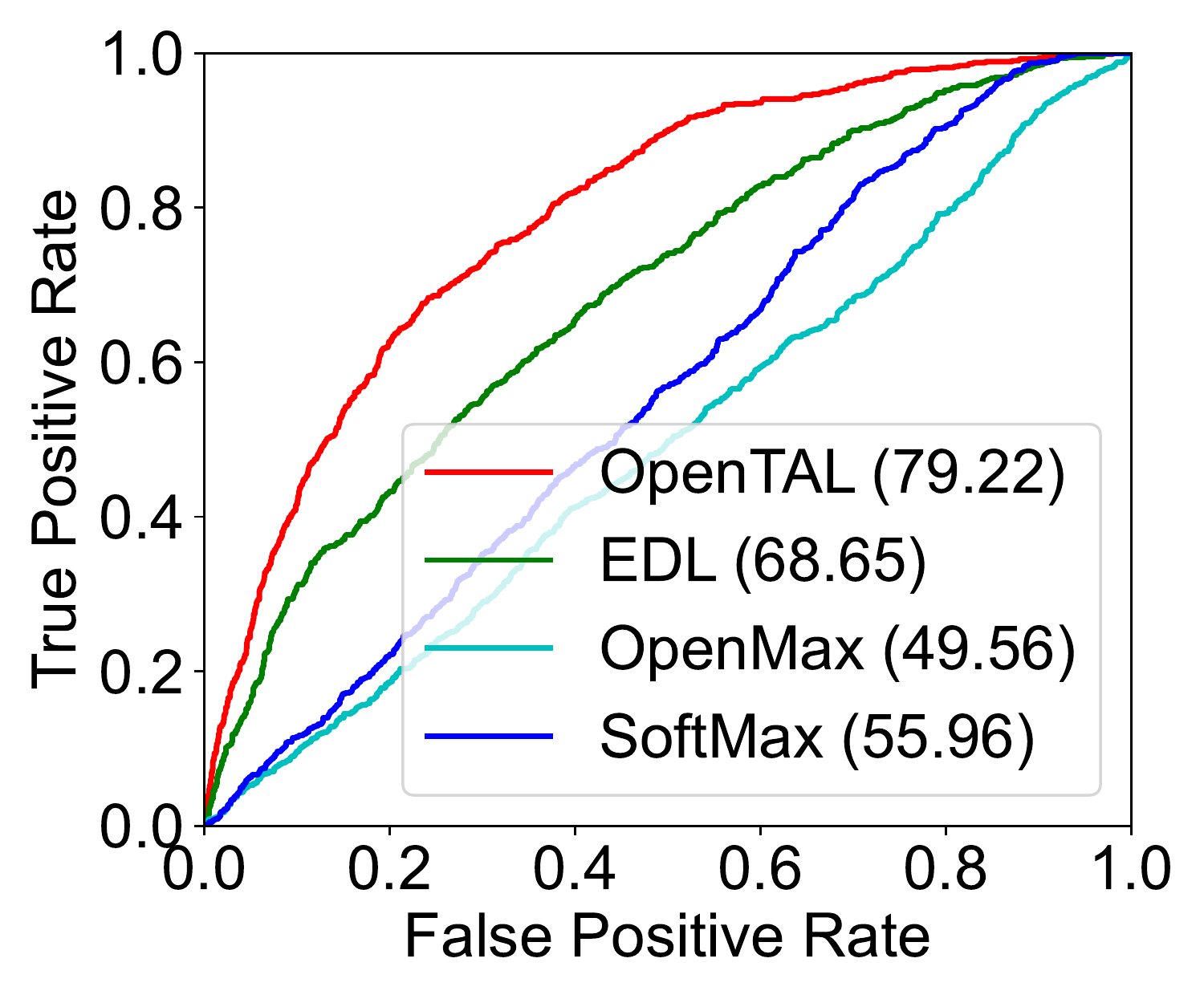}
    }
    \subcaptionbox{OSDR Curves}{
        \includegraphics[width=0.47\linewidth]{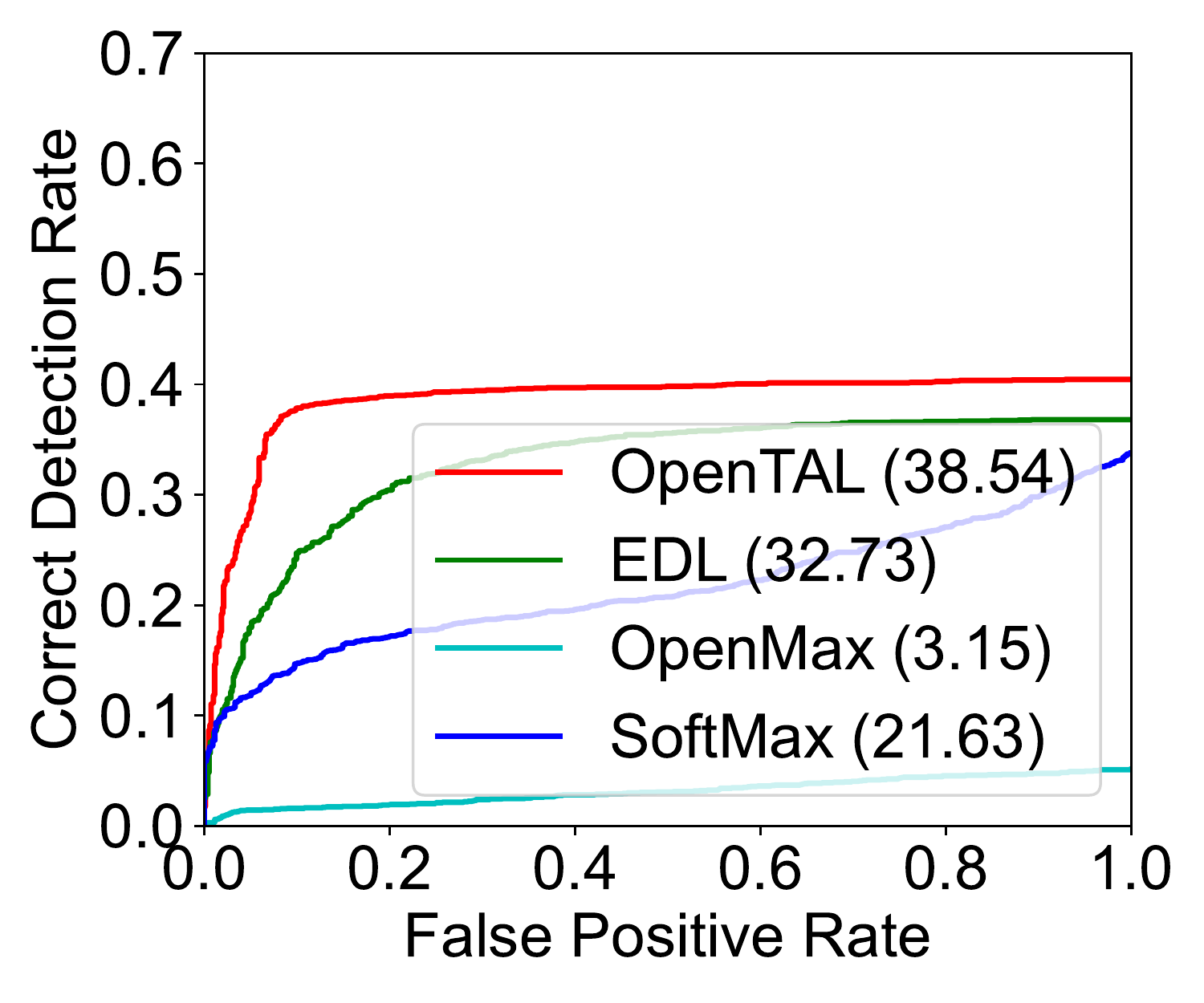}
    }
    \captionsetup{font=small,aboveskip=5pt}
    \caption{\textbf{ROC and OSDR curves on one THUMOS14 split.} Numbers in the brackets are AUROC or OSDR values.}
    \label{fig:curves}
\end{figure}

%% file: supp_v1.tex
\begin{appendix}

\section*{Supplementary Material}

In this document, we provide the detailed proof of the gradient of the EDL loss (Sec.~\ref{sec:edl_grad}), the dataset description of the open set setting (Sec.~\ref{sec:dataset}), implementation details (Sec.~\ref{sec:impl}), additional results and discussions (Sec.~\ref{sec:res_exp}).

\section{Gradient of EDL}
\label{sec:edl_grad}

Given the DNN logits $\mathbf{z}_i\in\mathbb{R}^K$ of sample $x_i$, an evidence function defined by $\exp$ is applied to the logits to get the class-wise evidence prediction, i.e., $\mathbf{e}_i=\exp(\mathbf{z}_i)$. Following the maximum likelihood loss form of Evidential Deep Learning (EDL)~\cite{SensoyNIPS2018}, we have the EDL loss:
\begin{equation}
    \mathcal{L}_{\text{EDL}}^{(i)}(\boldsymbol{\alpha}_i) = \sum_{j=1}^K t_{ij}(\log(S_{i}) - \log(\alpha_{ij})),
\label{supp_eq:edl}
\end{equation}
where $t_{ij}=1$ iff. the class label $y_i=j$, otherwise $t_{ij}=0$. The total Dirichlet strength $S_i=\sum_j \alpha_{ij}$ and the class-wise strength $\boldsymbol{\alpha}_i = \mathbf{e}_i + 1$. Therefore, according to the simple chain rule, we have the partial derivative:
\begin{equation}
    \frac{\partial \alpha_{ij}}{\partial z_{ij}} = \frac{\partial \alpha_{ij}}{\partial e_{ij}}\cdot \frac{\partial e_{ij}}{\partial z_{ij}} = e_{ij} 
\end{equation}
Then, the gradient of the $j$-th entry in Eq.~\eqref{supp_eq:edl}, i.e.,  $\mathcal{L}_{\text{EDL}}^{(ij)}$, w.r.t. the logits $z_{ij}$ can be derived as follows:
\begin{equation}
    \begin{split}
        g_{ij} = \frac{\partial \mathcal{L}_{\text{EDL}}^{(ij)}}{\partial z_{ij}} &= t_{ij} \left[\frac{1}{S_i} \frac{\partial S_i}{\partial z_{ij}} - \frac{1}{\alpha_{ij}} \frac{\partial \alpha_{ij}}{\partial z_{ij}} \right] \\
        &= t_{ij} \left[\frac{1}{S_i} \sum_{k=1}^{K} \frac{\partial \alpha_{ik}}{\partial z_{ij}} - \frac{e_{ij}}{\alpha_{ij}}\right] \\
        &= t_{ij} \left[\frac{1}{S_i} \sum_{k=1}^{K} e_{ik} - \frac{e_{ij}}{\alpha_{ij}}\right] 
    \end{split}
\end{equation}
Consider that $S_i=\sum_k \alpha_{ik}=\sum_j e_{ik} + K$, and the evidential uncertainty $u_i=K/S_i$, we further simplify the $g_{ij}$ as follows:
\begin{equation}
    \begin{split}
        g_{ij} &= t_{ij} \left[\frac{S_i - K}{S_i} - \frac{\alpha_{ij} - 1}{\alpha_{ij}}\right]  \\
        &= t_{ij} \left[\frac{S_i - K\alpha_{ij}}{S_i\alpha_{ij}}\right] \\
        &= t_{ij} \left[\frac{1}{\alpha_{ij}} - u_i\right],
    \end{split}
\end{equation}
which has proved the equation of $g_{ij}$ in our main paper. From this conclusion, when considering that $\alpha_{ij}\in (1,\infty)$ and $u_i \in (0,1)$, we have the property $|g_{ij}|\in [0,1)$.

Furthermore, consider the last DNN layer parameters $\mathbf{w}\in \mathbb{R}^{D\times K}$ such that $\mathbf{z}_i=\mathbf{w}^T\mathbf{h}_i$ where $\mathbf{h}_i\in \mathbb{R}^D$ is the high-dimensional feature of $x_i$, we can derive the gradient of EDL loss w.r.t. parameters $\mathbf{w}$: \begin{equation}
    \nabla_{w}\mathcal{L} = \frac{\partial \mathcal{L}_{\text{EDL}}^{(ij)}}{\partial w_{dk}} = \frac{\partial \mathcal{L}_{\text{EDL}}^{(ik)}}{\partial z_{ik}} \cdot \frac{\partial z_{ik}}{\partial w_{dk}} = g_{ik}\cdot h_{id},
\end{equation}
where $w_{dk}$ and $h_{id}$ are elements of the matrix $\mathbf{w}$ and the vector $\mathbf{h}_i$. Similar to~\cite{Park_2021_ICCV}, we consider the influence function~\cite{koh2017understanding} by ignoring the inverse of Hessian and using the magnitude ($L_1$ norm) of the gradient:
\begin{equation}
\begin{split}
    \omega_{i} = \|\nabla_{w}\mathcal{L}\|_1 & =\sum_{k=1}^{K}\sum_{d=1}^{D} |g_{ik}\cdot h_{id}| \\
    &= \left(\sum_{k=1}^{K}|g_{ik}|\right)\left(\sum_{d=1}^{D}|h_{id}|\right) \\
    &= \|\mathbf{g}_{i}\|_1\cdot\|\mathbf{h}_i\|_1,
\end{split}
\end{equation}
which has proved the equation of $\omega_i$ in our main paper.

\section{Dataset Details}
\label{sec:dataset}

To enable the existing Temporal Action Localization (TAL) datasets such as THUMOS14~\cite{THUMOS14} and ActivityNet1.3~\cite{ANetCVPR2015} for the open set TAL setting, a subset of action categories has to be reserved as the unknown used in open set testing. In practice, we randomly splitted the THUMOS14 three times into known and unknown subsets of categories. For each split, a model will be trained on the closed set (which only contains known categories), and tested on the open set that contains both known and unknown categories. Table~\ref{tab:splits} shows the detailed information of the three dataset splits from THUMOS14. 

To further increase the openness in testing, we incorporate activity categories from ActivityNet1.3 that are non-overlapped with THUMOS14 into the open set testing. Specifically, the following 14 overlapping activity categories are removed: \textit{Table soccer, Javelin throw, Clean and jerk, Springboard diving, Pole vault, Cricket, High jump, Shot put, Long jump, Hammer throw, Snatch, Volleyball, Plataform diving, Discus throw}. Note that we did not use ActivityNet1.3 for similar model training as the THUMOS14, e.g., train a model on multiple random splits of ActivityNet1.3, due to the limited computational resource.

\begin{table}[t]
\centering
\captionsetup{font=small,aboveskip=3pt}
\caption{\textbf{THUMOS14 Splits for Open Set TAL.} For each split, five out of twenty action categories are randomly selected as the unknown (\textbf{U}) used in open set testing, while the rest fifteen categories are the known (\textbf{K}) used in model training.}
\label{tab:splits}
\small
\setlength{\tabcolsep}{4.5mm}
\setlength{\extrarowheight}{0.5mm}
\begin{tabular}{l|ccc}\toprule
&Split 1 &Split 2 &Split 3 \\
\hline
\emph{BaseballPitch} &K &K &K \\
\emph{BasketballDunk} &K &K &K \\
\emph{Billiards} &K &K &K \\
\emph{CricketBowling} &K &U &K \\
\emph{CricketShot} &K &K &U \\
\emph{FrisbeeCatch} &K &K &K \\
\emph{GolfSwing} &K &K &K \\
\emph{HammerThrow} &K &U &K \\
\emph{HighJump} &K &K &K \\
\emph{JavelinThrow} &K &U &U \\
\emph{PoleVault} &K &K &U \\
\emph{Shotput} &K &K &U \\
\emph{TennisSwing} &K &K &K \\
\emph{ThrowDiscus} &K &K &K \\
\emph{VolleyballSpiking} &K &K &K \\
\emph{CleanAndJerk} &U &K &K \\
\emph{CliffDiving} &U &U &K \\
\emph{Diving} &U &U &K \\
\emph{LongJump} &U &K &U \\
\emph{SoccerPenalty} &U &K &K \\
\bottomrule
\end{tabular}
\end{table}

\input{supp_tex/supp_tables}

\section{Implementation Details}
\label{sec:impl}

\paragraph{Detailed Architecture} The proposed OpenTAL is primarily implemented on the AFSD~\cite{AFSD_CVPR2021} framework. It uses a pre-trained I3D~\cite{I3DCVPR2017} as the feature extraction backbone and a 6-layer temporal FPN architecture is applied to the I3D for action classification and localization. Each level consists of a coarse stage, a saliency-based proposal refinement module, and a refined stage. The first two pyramid levels use 3D convolutional (Conv3D) block while the rest four levels use 1D convolutional (Conv1D) block. Group Normalization and ReLU activation are utilized in each block. The temporal localization head and action classification head are implemented by a shared Conv1D block across all 6 levels. To implement OpenTAL method, the $(K+1)$-way classification head is replaced with $K$-way evidential neural network head, while the localization head is kept unchanged. We additionally add an actionness prediction branch which consists of a Conv1D block for both the coarse and the refined stages. 

\paragraph{Training and Testing} In training, the proposed classification loss $\mathcal{L_{\text{MIB-EDL}}}$ and actionness prediction loss $\mathcal{L}_{\text{ACT}}$ are applied to both the coarse and refined stages in AFSD, while the IoU-aware uncertainty calibration loss $\mathcal{L}_{\text{IoUC}}$ is only applied to the refined stage because this loss function is dependent of the pre-computed temporal IoU using the predicted action locations in the coarse stage. Similar to AFSD, we used temporal IoU threshold 0.5 in the training to identify the foreground actions from the proposals. Besides, we reduced the weight of triplet loss in AFSD to 0.001 since the contrastive learning loss would not work well when there are unknown action clips in the background. The whole model is trained by Adam optimizer with base learning rate 1e-5 and weight decay 1e-3. All models are trained with 25 epochs to ensure full convergence and the model snapshot of the last epoch is used for testing and evaluation.

In testing, the actionness score is multiplied to the confidence score before the soft-NMS post-processing module. The $\sigma$ and top-$N$ hyperparameters are set to 0.5 and 5000, which are recommended by the AFSD.

\section{Additional Results}
\label{sec:res_exp}

\paragraph{Impact of tIoU Thresholds} Since the proposed OSTAL task cares not only the classification but also the temporal localization, we present the experimental results under different temporal IoU (tIoU)
thresholds. Following existing TAL literature, we set five tIoU thresholds $[0.3:0.1:0.7]$ when the unknown classes are from THUMOS14 and ten tIoU thresholds $[0.5:0.05:0.95]$ when the unknown classes are from ActivityNet1.3, respectively. Evaluation results by AUROC, AUPR, and OSDR are reported in Table~\ref{tab:auroc}, \ref{tab:aupr}, and \ref{tab:osdr}, respectively. The results show that AUROC performances are stable across different tIoU thresholds, while the AUPR and OSDR performances vary significantly as the tIoU threshold changes. Besides, as the tIoU threshold increasing, AUROC and OSDR values would increase accordingly. For all those tIoU thresholds and evaluation metrics, the proposed OpenTAL could consistently outperform baselines.

\vspace{-2mm}
\paragraph{Impact of Dataset Splits} For open set problems, splitting an existing fully annotated dataset into known and unknown part plays an important role in performance evaluation. In this document, we comprehensively show the ROC, PR, and OSDR curves on three different THUMOS14 open set splits in Fig.~\ref{fig:roc_curves}, Fig~\ref{fig:pr_curves}, and Fig.~\ref{fig:osdr_curves}, respectively. From these figures, we can find that the ROC curves vary much more across different splits than tIoU thresholds, while the PR and OSDR curves vary significantly both across the splits and tIoU thresholds. Besides, for all sub-figures, the proposed OpenTAL could significantly outperform baselines.

\paragraph{More Visualizations} In this document, we add more visualizations for comparing the proposed OpenTAL with baselines in Fig~\ref{fig:supp_demo}. The first 4 examples (in the first 2 rows) show that OpenTAL could well localize and recognize known actions (colorful segments). The rest of 12 examples show that OpenTAL can roughly localize and reject the unknown actions (black segments).

\input{supp_tex/roc_curves}

\input{supp_tex/pr_curves}

\input{supp_tex/osdr_curves}

\input{supp_tex/demo}

\end{appendix}

%% file: supp_tex/supp_tables.tex
\begin{table*}[!htp]
\centering
\captionsetup{font=small,aboveskip=3pt}
\caption{\textbf{AUROC Results (\%) vs. Different tIoU Thresholds}. Models trained on the THUMOS14 closed set are tested by including the unknown classes from THUMOS14 and ActivityNet1.3, respectively. Results are averaged over the three dataset splits. }
\label{tab:auroc}
\small
\setlength{\tabcolsep}{3.9mm}
\setlength{\extrarowheight}{0.5mm}
\begin{tabular}{l|cccccc|cccc}
\toprule
\multirow{2}{*}{Methods} &\multicolumn{6}{c|}{THUMOS14 as the Unknown} &\multicolumn{4}{c}{ActivityNet1.3 as the Unknown} \\
\cline{2-11}
&0.3 &0.4 &0.5 &0.6 &0.7 &Avg. &0.5 &0.75 &0.95 &Avg. \\
\hline
SoftMax &54.70 &55.46 &56.41 &57.12 &57.11 &56.16 & 56.97 & 58.41 & 55.97 & 57.77 \\
OpenMax~\cite{BendaleCVPR2016} &53.26 &52.1 &52.13 &51.89 &52.53 &52.38 & 51.24 & 52.39 & 49.13  & 51.59\\
EDL~\cite{BaoICCV2021} &64.05 &64.27 &65.13 &66.21 &66.81 &65.29 & 62.82 & 66.23 & 67.92 & 65.69 \\
OpenTAL &\textbf{78.33} &\textbf{79.04} &\textbf{79.30} &\textbf{79.40} &\textbf{79.82} &\textbf{79.18} & \textbf{82.97} & \textbf{83.21} & \textbf{83.38} & \textbf{83.22} \\
\bottomrule
\end{tabular}
\end{table*}

\begin{table*}[!htp]
\centering
\captionsetup{font=small,aboveskip=3pt}
\caption{\textbf{AUPR Results (\%) vs. Different tIoU Thresholds}. Models trained on the THUMOS14 closed set are tested by including the unknown classes from THUMOS14 and ActivityNet1.3, respectively. Results are averaged over the three dataset splits. }
\label{tab:aupr}
\small
\setlength{\tabcolsep}{3.9mm}
\setlength{\extrarowheight}{0.5mm}
\begin{tabular}{l|cccccc|cccc}
\toprule
\multirow{2}{*}{Methods} &\multicolumn{6}{c|}{THUMOS14 as the Unknown} &\multicolumn{4}{c}{ActivityNet1.3 as the Unknown} \\
\cline{2-11}
&0.3 &0.4 &0.5 &0.6 &0.7 &Avg. &0.5 &0.75 &0.95 &Avg. \\
\hline
SoftMax &31.85 &31.81 &31.11 &29.78 &27.99 &30.51 & 53.54 & 44.15 & 34.54 & 44.77 \\
OpenMax~\cite{BendaleCVPR2016} &33.17 &31.61 &30.59 &29.15 &28.45 &30.60 & 54.88 & 48.37 & 40.07  & 48.48 \\
EDL~\cite{BaoICCV2021} &40.05 &39.45 &38.05 &37.58 &36.35 &38.30 & 53.97 & 47.22 & 45.59 & 48.46 \\
OpenTAL &\textbf{58.62} &\textbf{59.40} &\textbf{58.78} &\textbf{57.54} &\textbf{55.88} &\textbf{58.04} & \textbf{80.41} & \textbf{74.20} &\textbf{73.92} & \textbf{75.54} \\
\bottomrule
\end{tabular}
\end{table*}

\begin{table*}[!htp]
\centering
\captionsetup{font=small,aboveskip=3pt}
\caption{\textbf{OSDR Results (\%) vs. Different tIoU Thresholds}. Models trained on the THUMOS14 closed set are tested by including the unknown classes from THUMOS14 and ActivityNet1.3, respectively. Results are averaged over the three dataset splits. }
\label{tab:osdr}
\small
\setlength{\tabcolsep}{3.9mm}
\setlength{\extrarowheight}{0.5mm}
\begin{tabular}{l|cccccc|cccc}
\toprule
\multirow{2}{*}{Methods} &\multicolumn{6}{c|}{THUMOS14 as the Unknown} &\multicolumn{4}{c}{ActivityNet1.3 as the Unknown} \\
\cline{2-11}
&0.3 &0.4 &0.5 &0.6 &0.7 &Avg. &0.5 &0.75 &0.95 &Avg. \\
\hline
SoftMax &23.40 &25.19 &27.43 &29.97 &32.08 &27.61 & 27.63 & 33.73 & 31.59 & 32.01 \\
OpenMax~\cite{BendaleCVPR2016} &13.66 &14.58 &15.91 &17.71 &20.41 &16.45 & 15.73 & 21.49 & 18.07 & 19.35\\
EDL~\cite{BaoICCV2021} &36.26 &37.58 &39.16 &41.18 &42.99 &39.43 & 38.56 & 43.72 & 42.20 & 42.18 \\
OpenTAL &\textbf{42.91} &\textbf{46.19} &\textbf{49.50} &\textbf{52.50} &\textbf{56.78} &\textbf{49.57} & \textbf{50.49} & \textbf{59.87} & \textbf{62.17} & \textbf{57.89} \\
\bottomrule
\end{tabular}
\end{table*}

%% file: supp_tex/roc_curves.tex
\newcommand{\framewidth}{0.195\linewidth}

\begin{figure*}[t]
\footnotesize
\centering
\renewcommand{\tabcolsep}{0.7pt} %
\begin{tabular}{cccccc}
& \textbf{tIoU=0.3} & \textbf{tIoU=0.4} & \textbf{tIoU=0.5} & \textbf{tIoU=0.6} & \textbf{tIoU=0.7}
\\
\parbox[c]{3mm}{\multirow{1}{*}[6.0em]{\rotatebox[origin=c]{90}{\textbf{split 1}}}} &
\includegraphics[width=\framewidth]{images/supp_curves/AUC_ROC_split0_tiou0.3.pdf} &
\includegraphics[width=\framewidth]{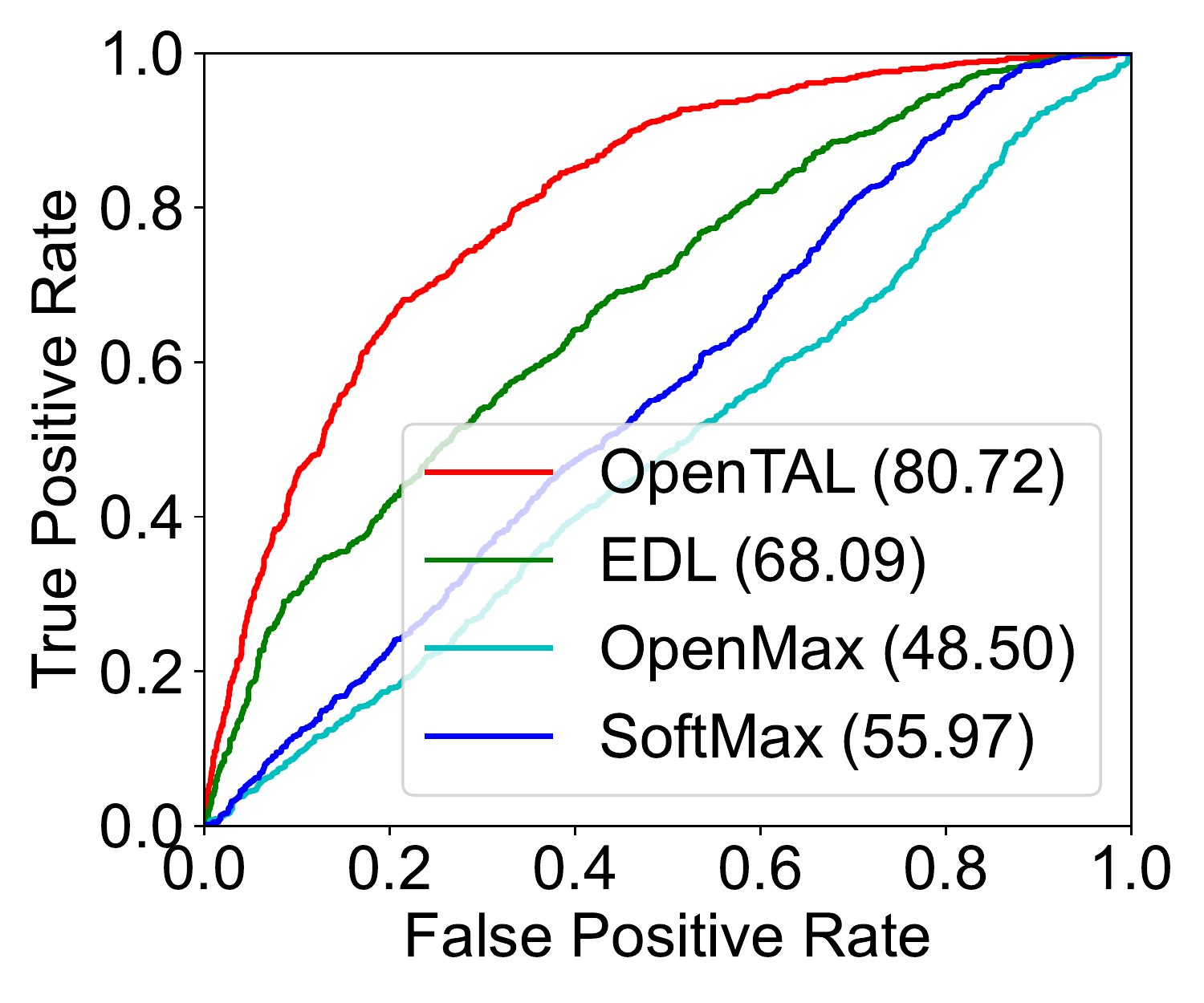} &
\includegraphics[width=\framewidth]{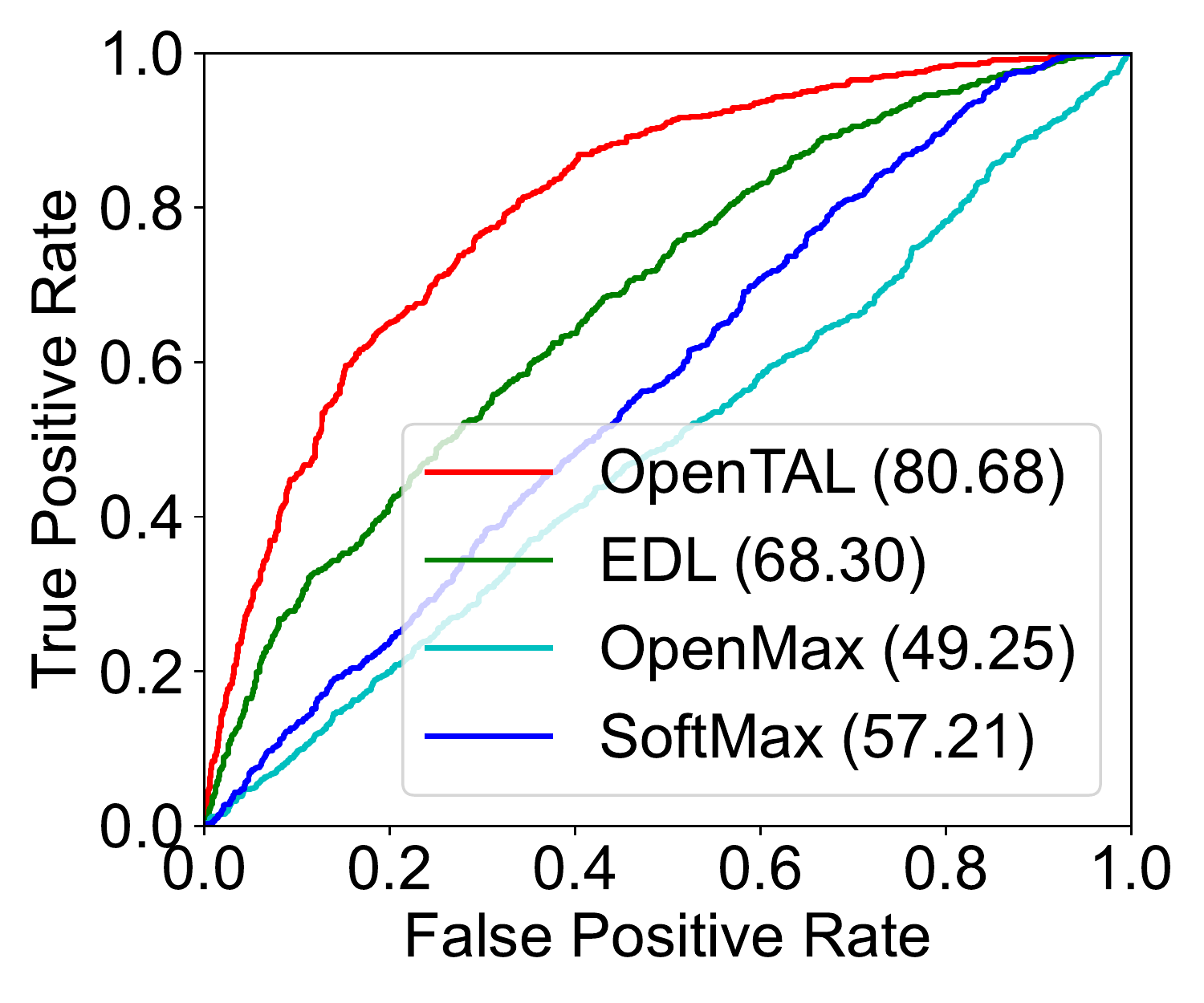} &
\includegraphics[width=\framewidth]{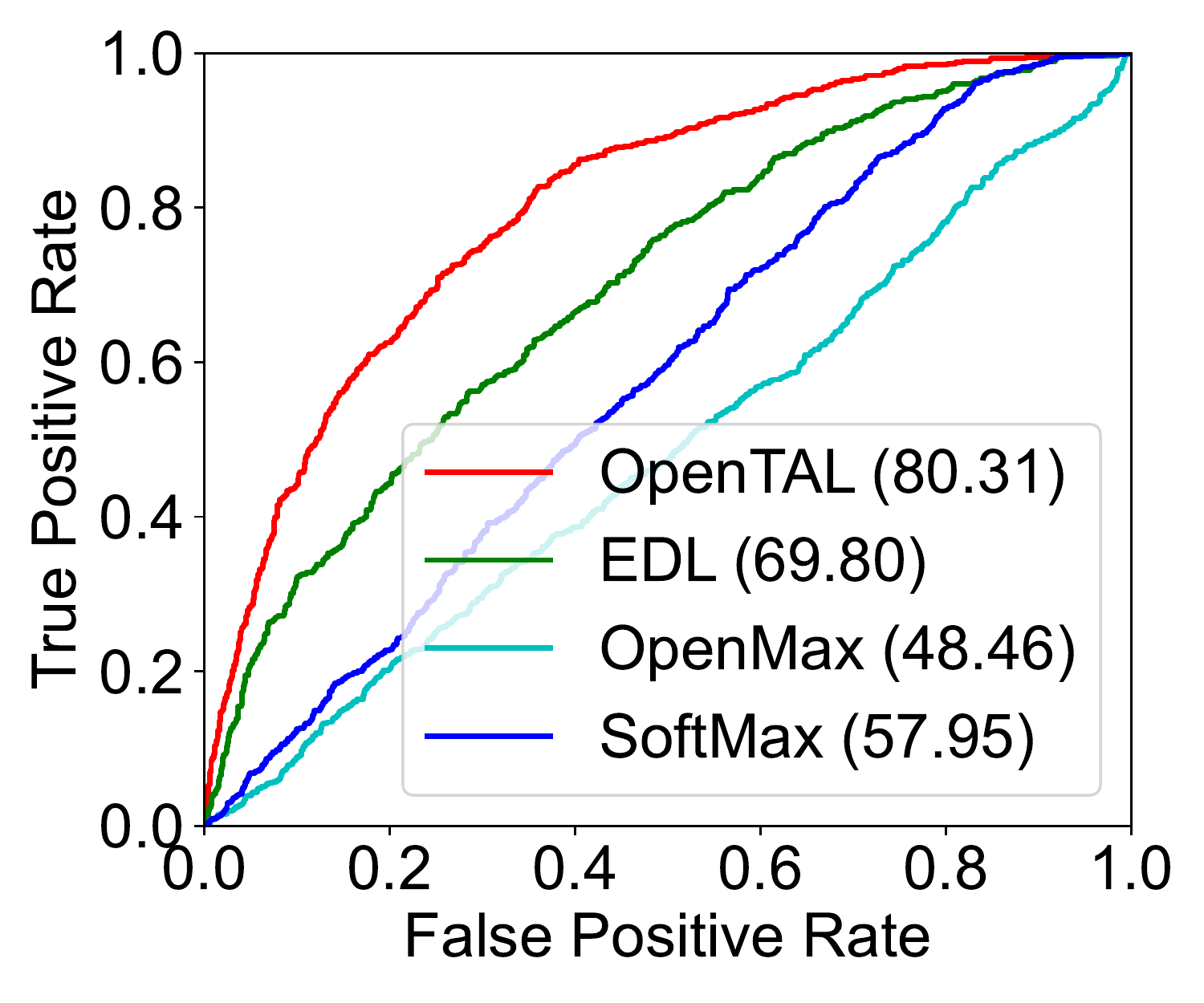} &
\includegraphics[width=\framewidth]{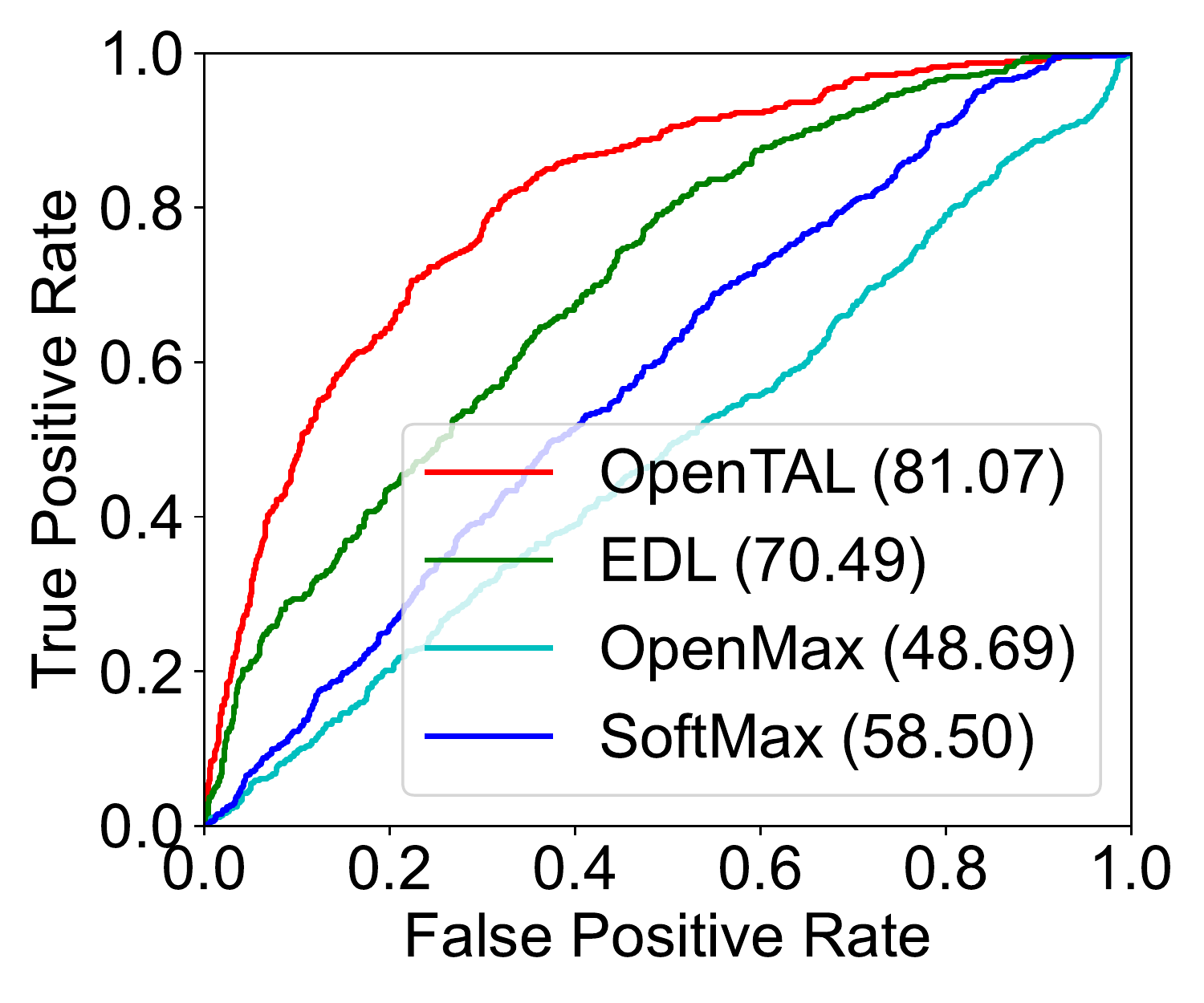}
\\
\parbox[c]{3mm}{\multirow{1}{*}[6.0em]{\rotatebox[origin=c]{90}{\textbf{split 2}}}} &
\includegraphics[width=\framewidth]{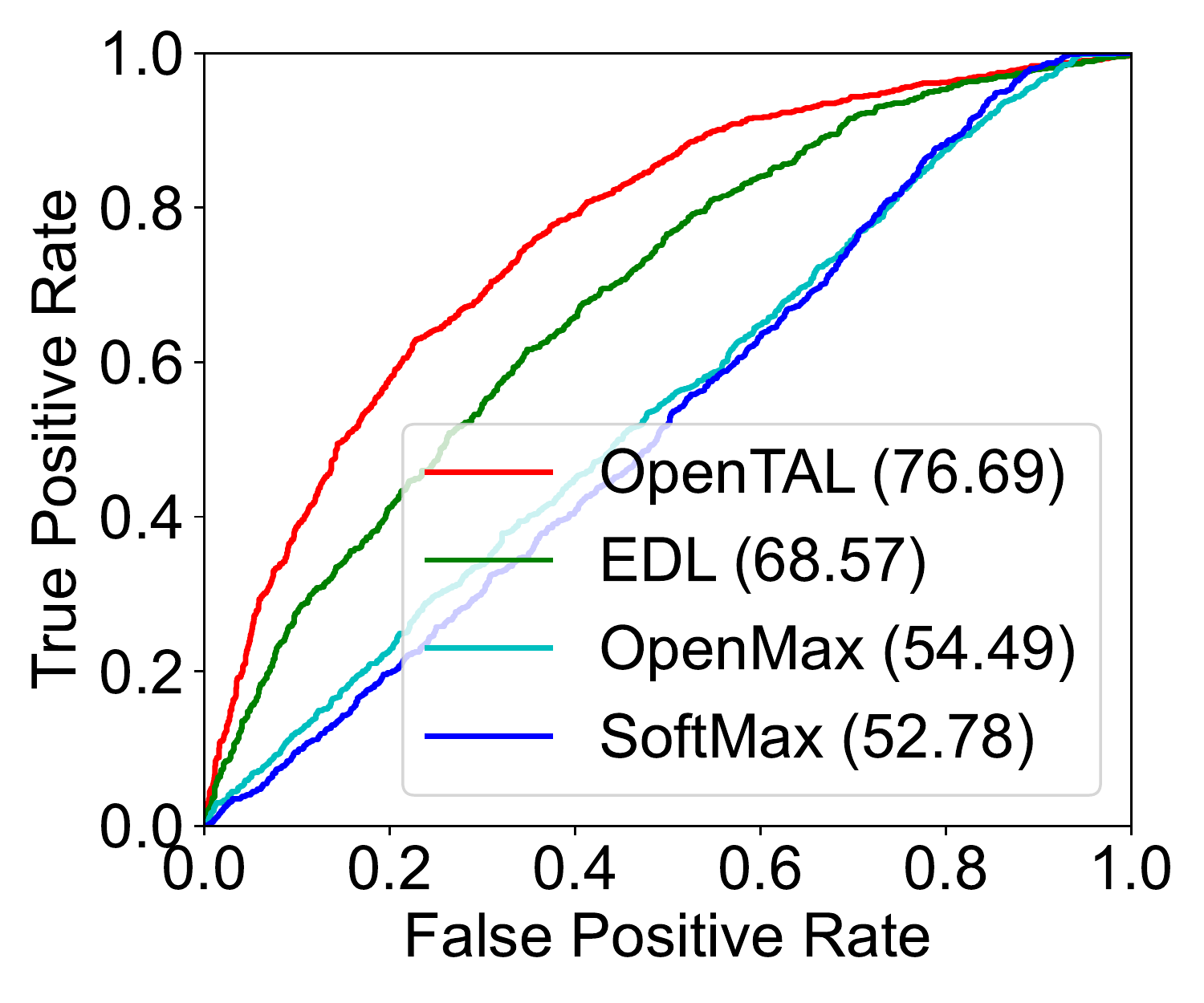} &
\includegraphics[width=\framewidth]{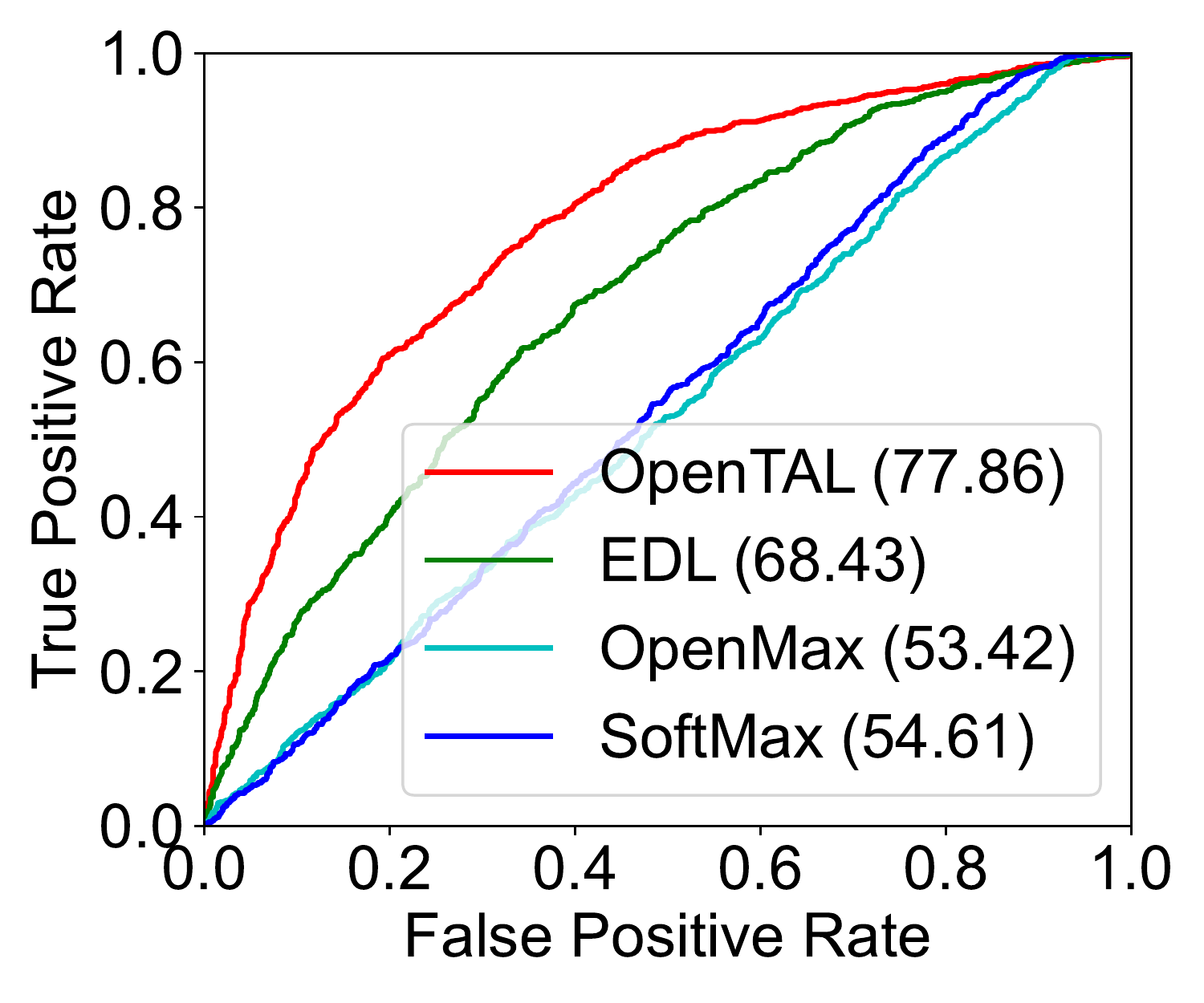} &
\includegraphics[width=\framewidth]{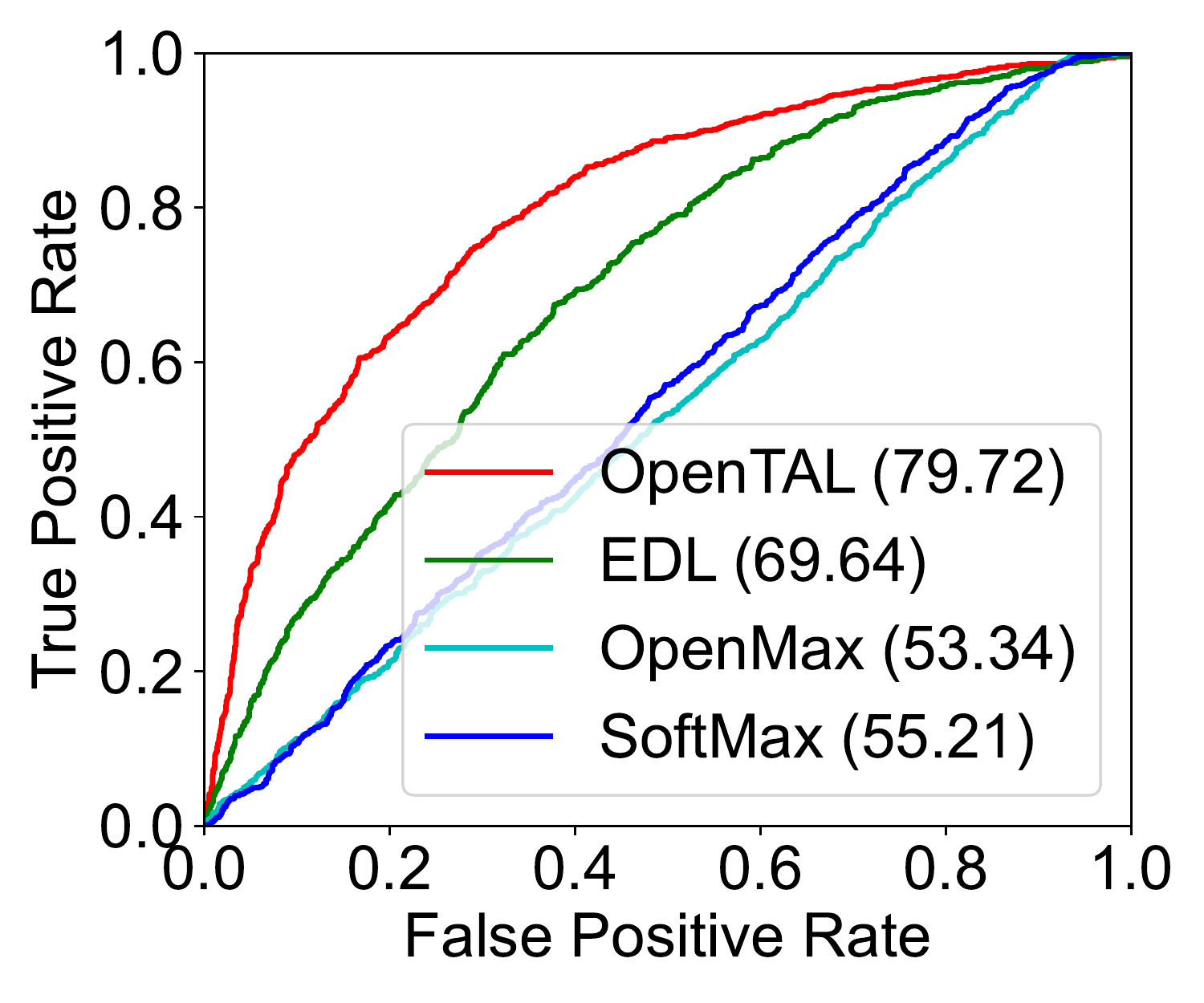} &
\includegraphics[width=\framewidth]{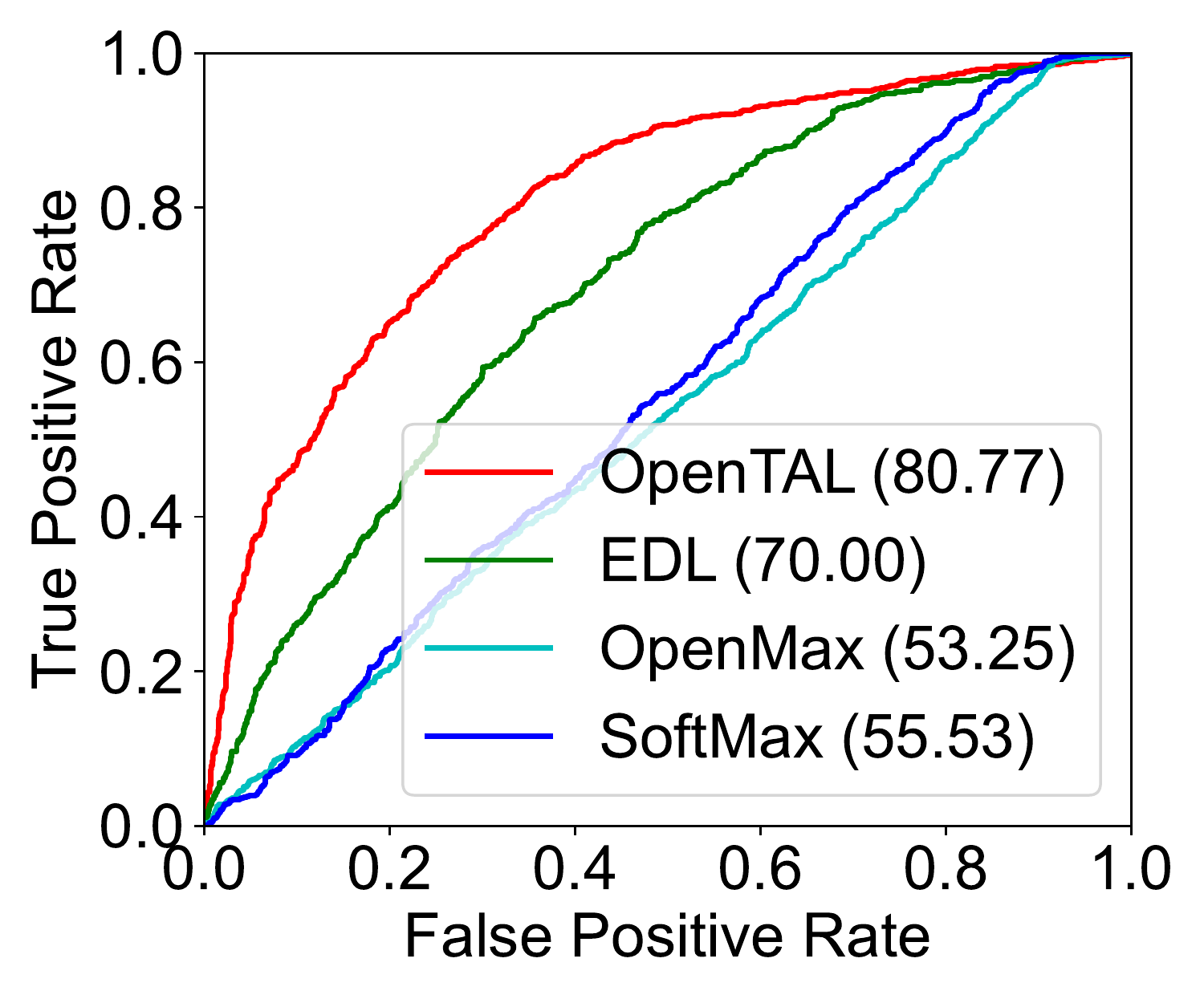} &
\includegraphics[width=\framewidth]{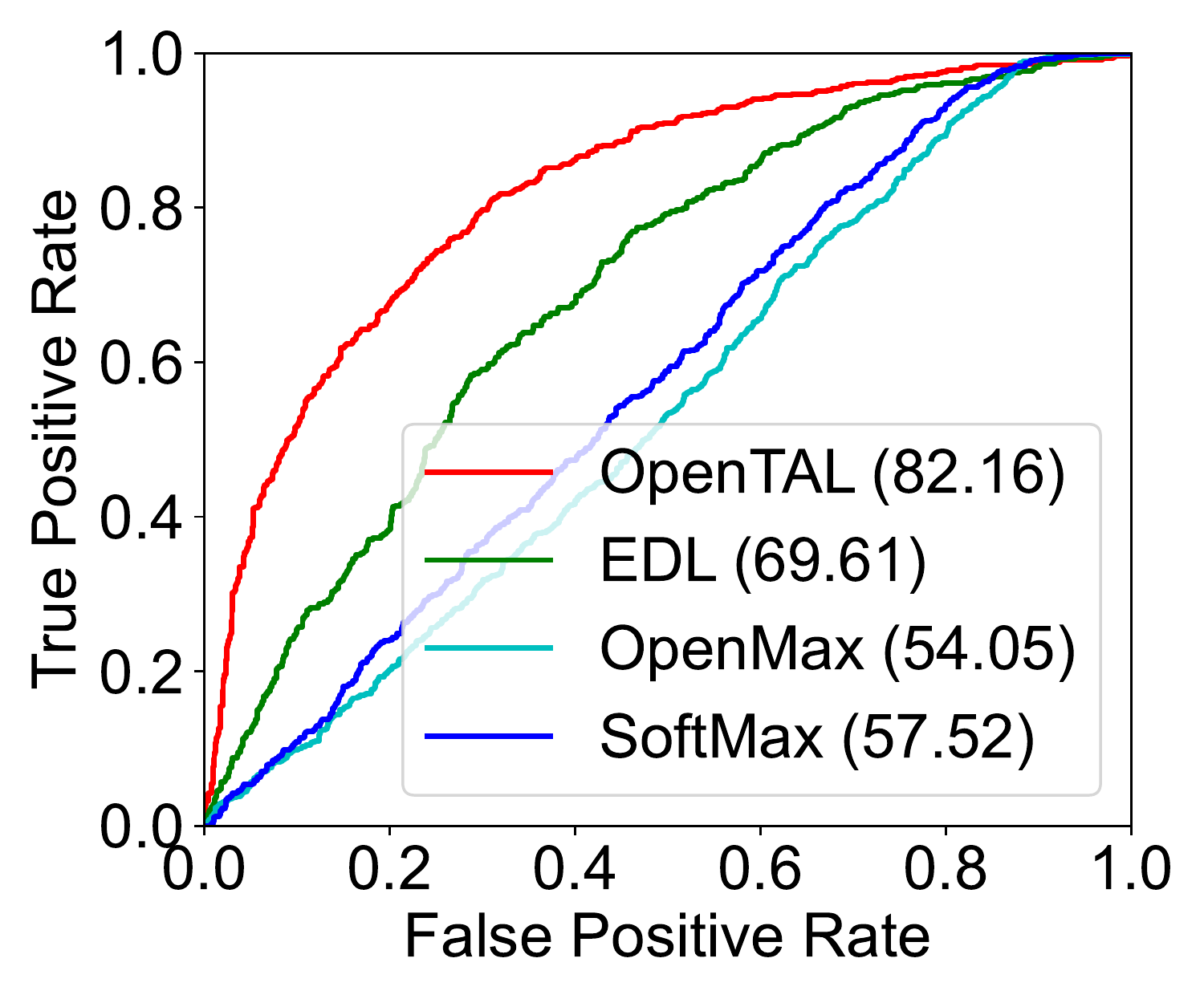}
\\
\parbox[c]{3mm}{\multirow{1}{*}[6.0em]{\rotatebox[origin=c]{90}{\textbf{split 3}}}} &
\includegraphics[width=\framewidth]{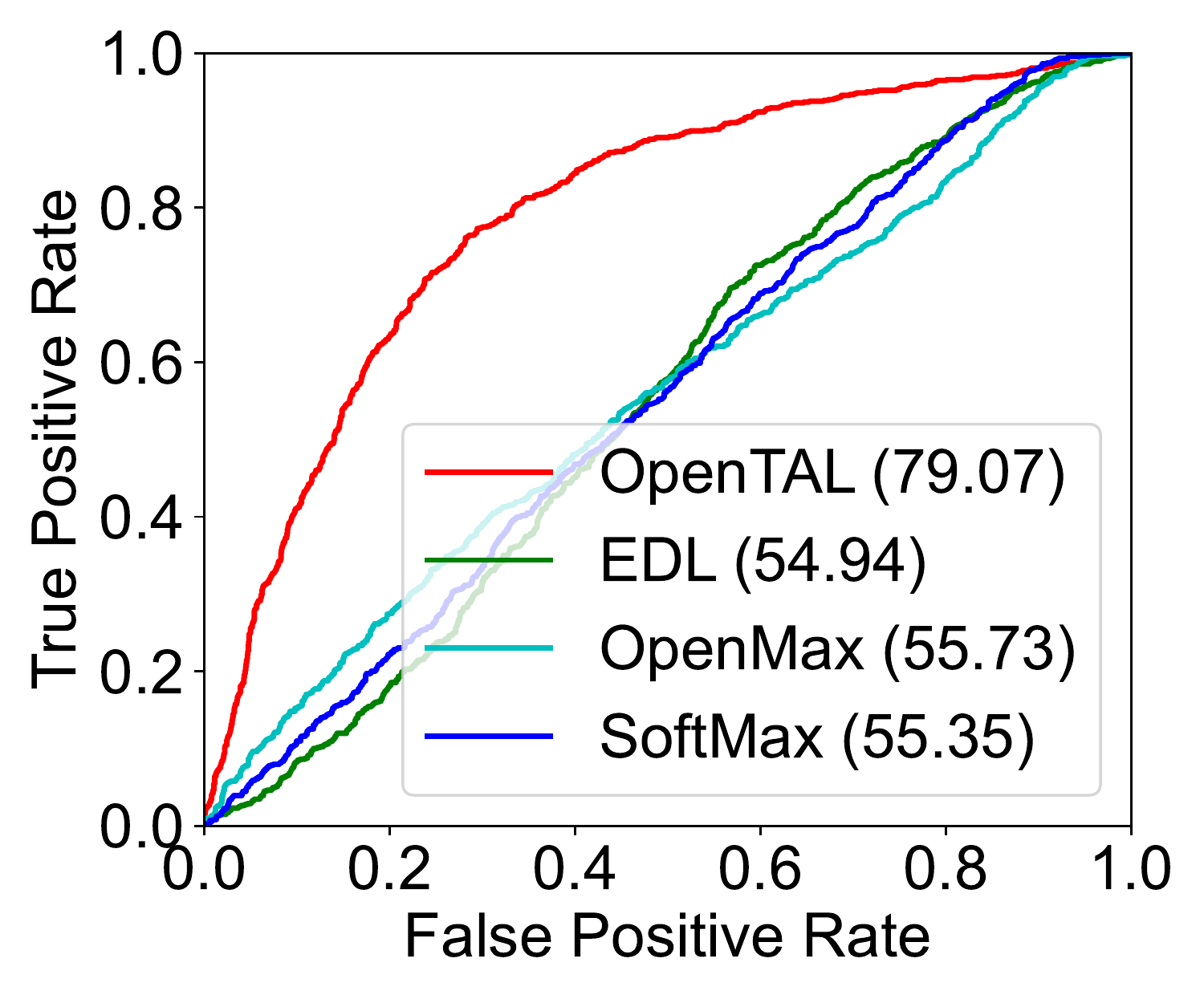} &
\includegraphics[width=\framewidth]{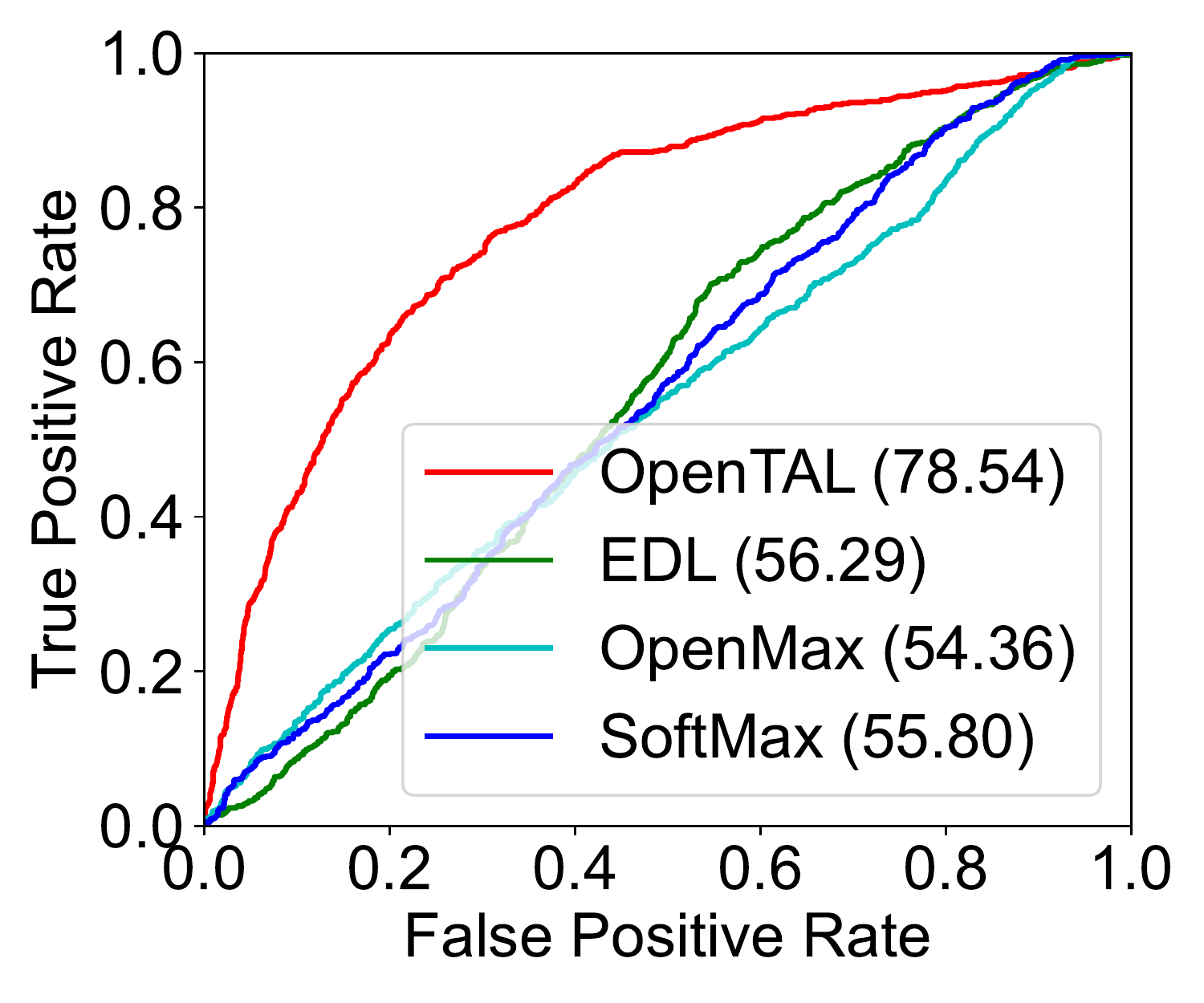} &
\includegraphics[width=\framewidth]{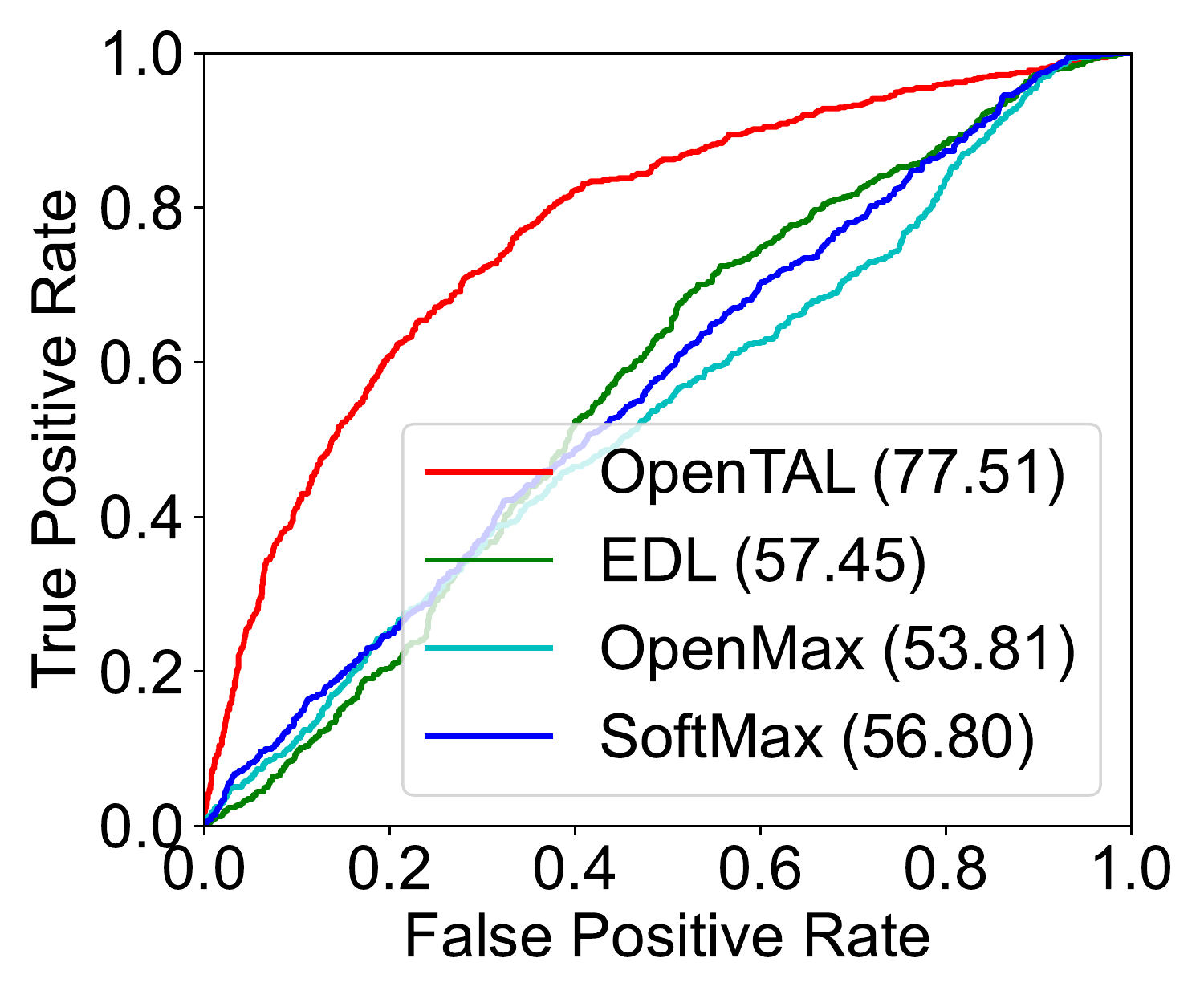} &
\includegraphics[width=\framewidth]{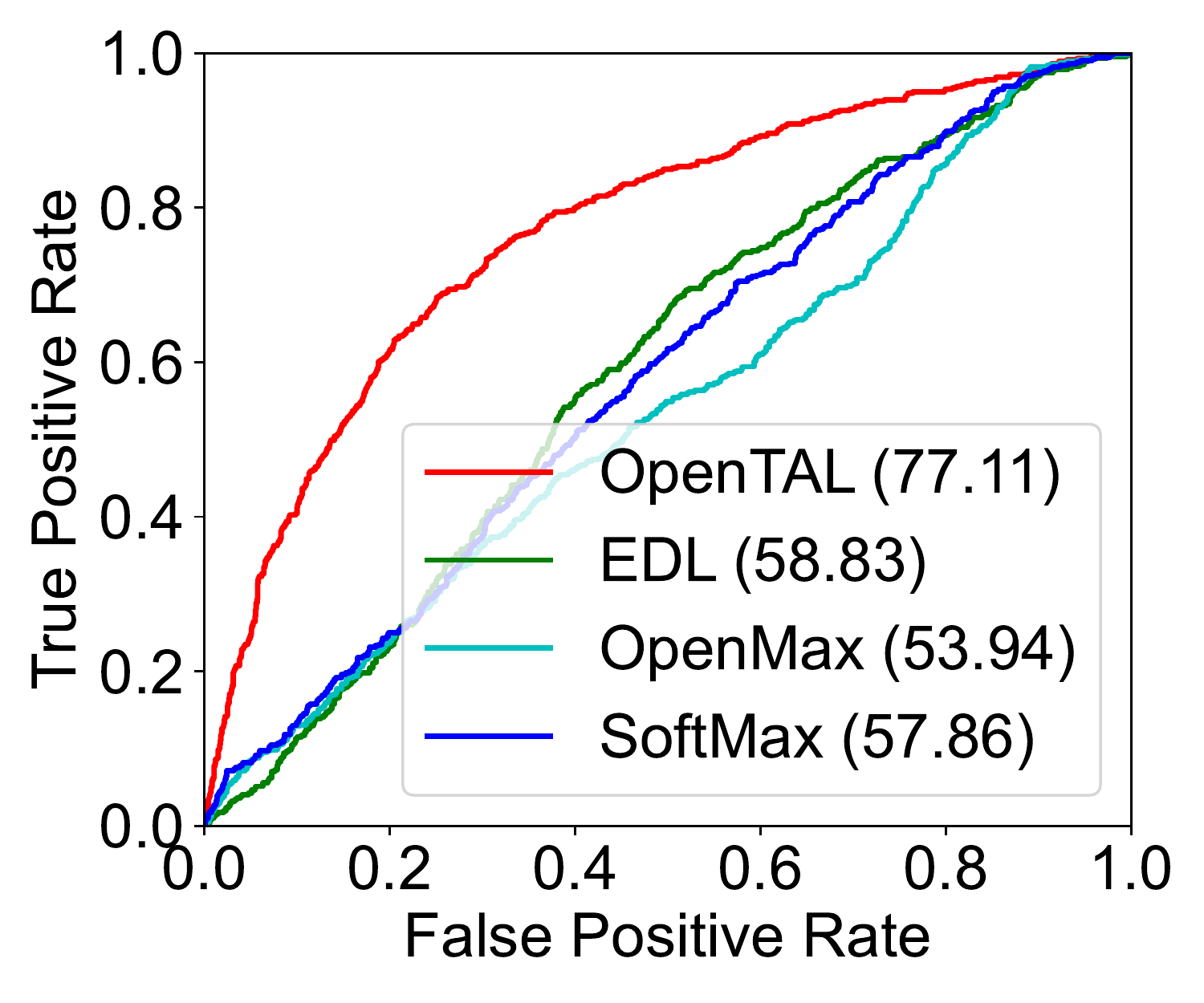} &
\includegraphics[width=\framewidth]{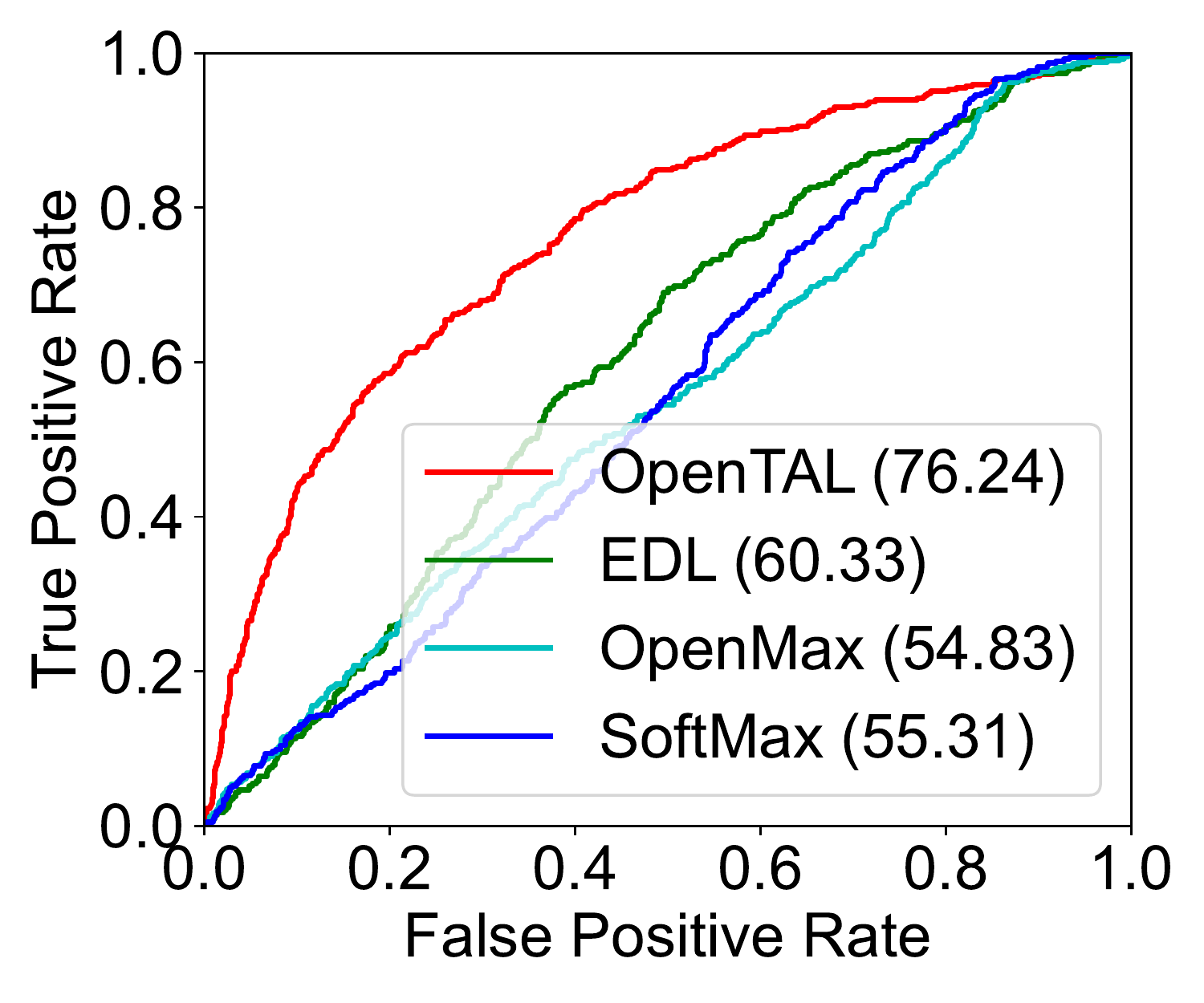}
\\
\end{tabular}
\captionsetup{font=small,aboveskip=3pt}
\caption{\textbf{ROC Curves.} These figures show the method comparison by ROC curves on THUMOS14 open set splits. Numbers in parentheses are AUROC values. They show that \ul{the ROC performance varies more across dataset splits than tIoU thresholds}, and our proposed OpenTAL could consistently outperform baselines on all the three splits and five thresholds.}
\label{fig:roc_curves}
\vspace{-10pt}
\end{figure*}

%% file: supp_tex/pr_curves.tex
\begin{figure*}[t]
\footnotesize
\centering
\renewcommand{\tabcolsep}{0.7pt} %
\begin{tabular}{cccccc}
& \textbf{tIoU=0.3} & \textbf{tIoU=0.4} & \textbf{tIoU=0.5} & \textbf{tIoU=0.6} & \textbf{tIoU=0.7}
\\
\parbox[c]{3mm}{\multirow{1}{*}[6.0em]{\rotatebox[origin=c]{90}{\textbf{split 1}}}} &
\includegraphics[width=\framewidth]{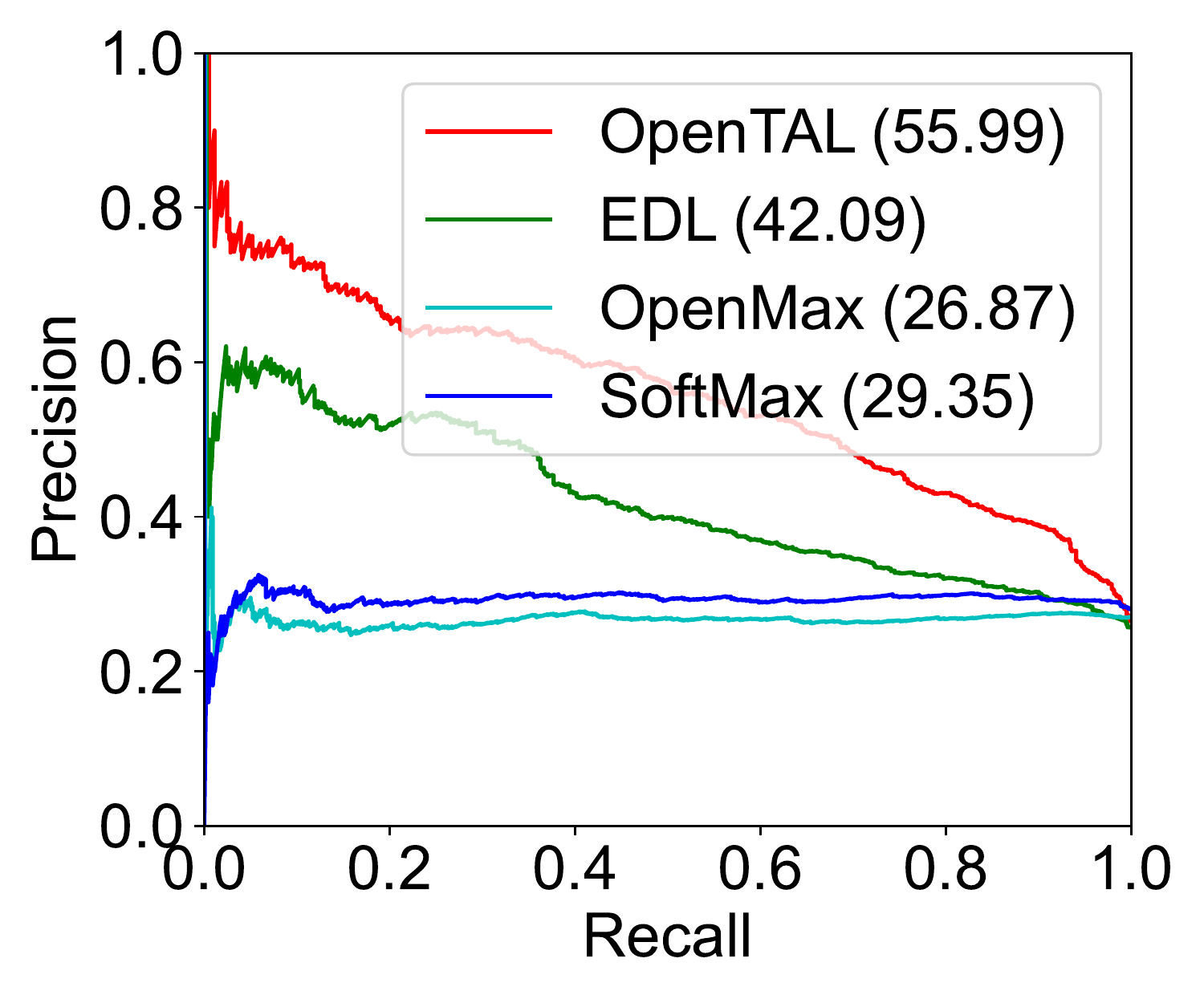} &
\includegraphics[width=\framewidth]{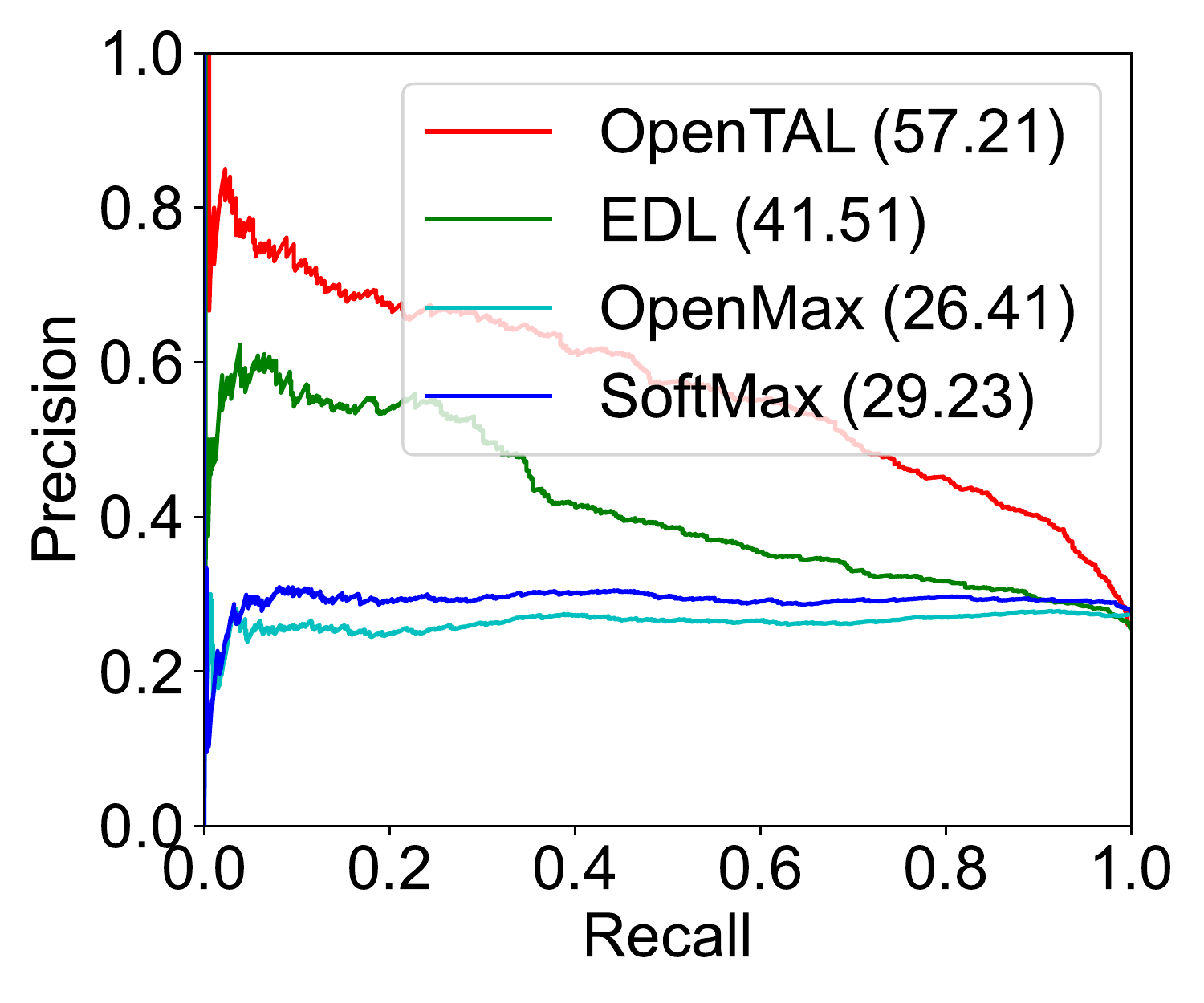} &
\includegraphics[width=\framewidth]{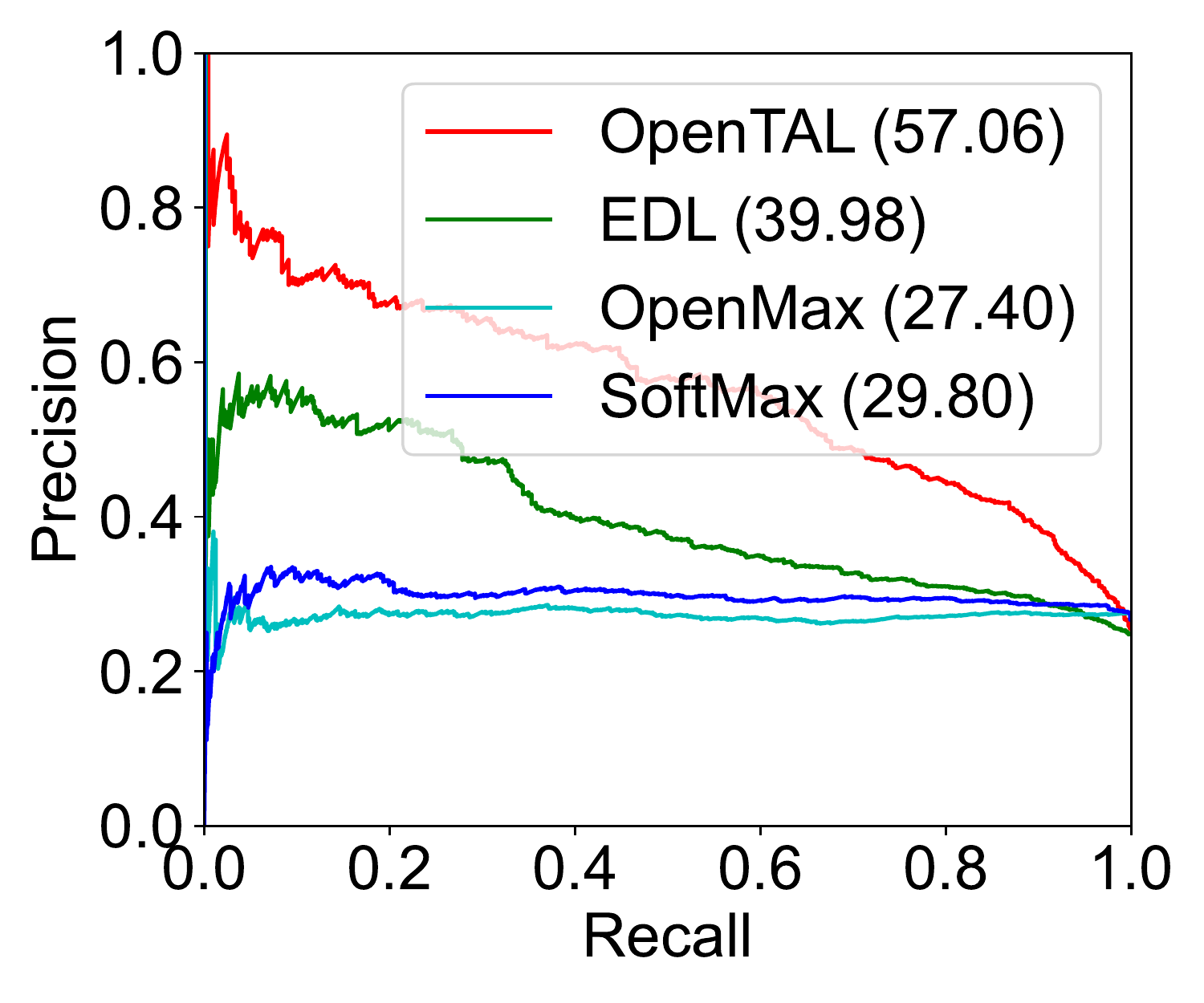} &
\includegraphics[width=\framewidth]{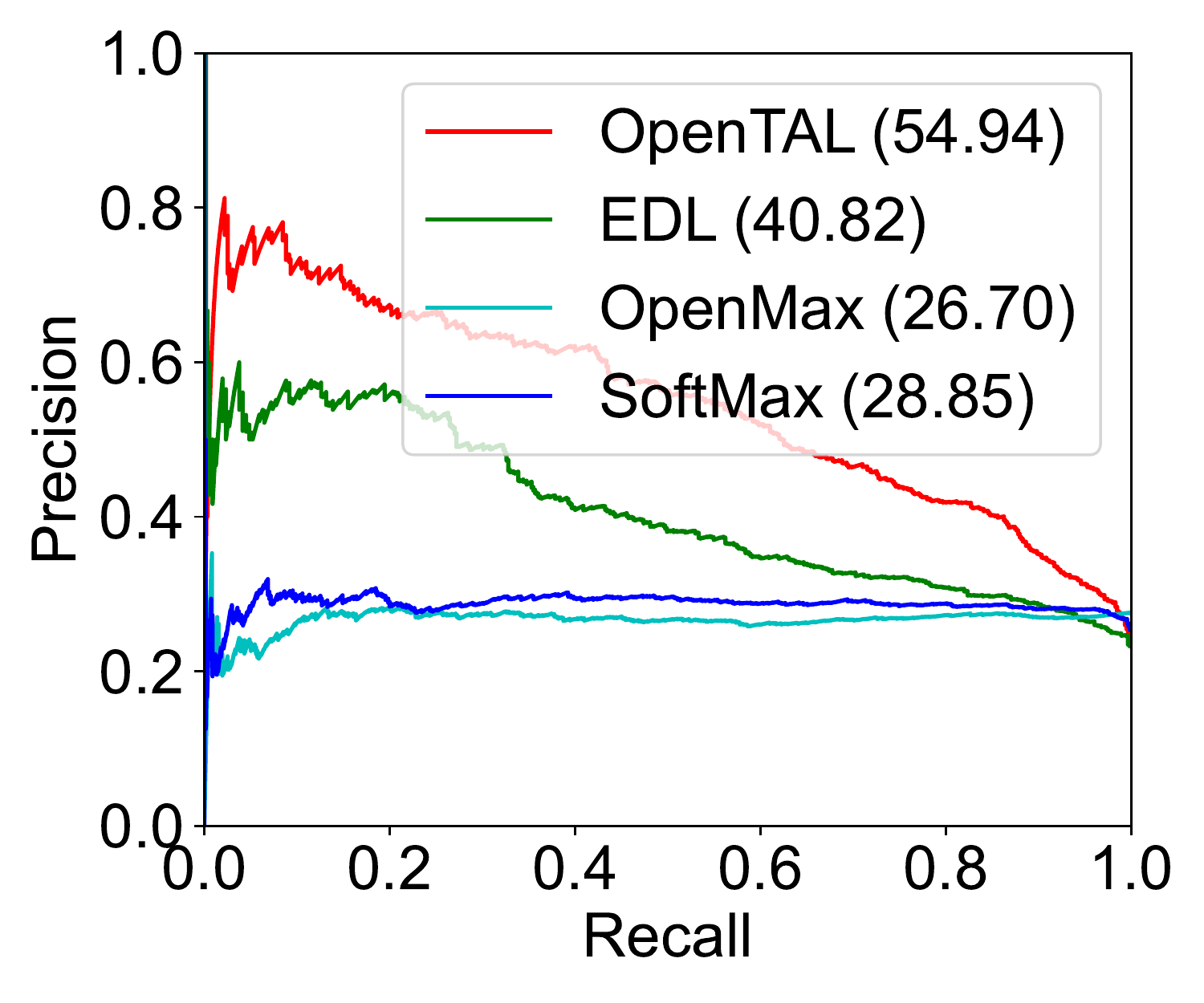} &
\includegraphics[width=\framewidth]{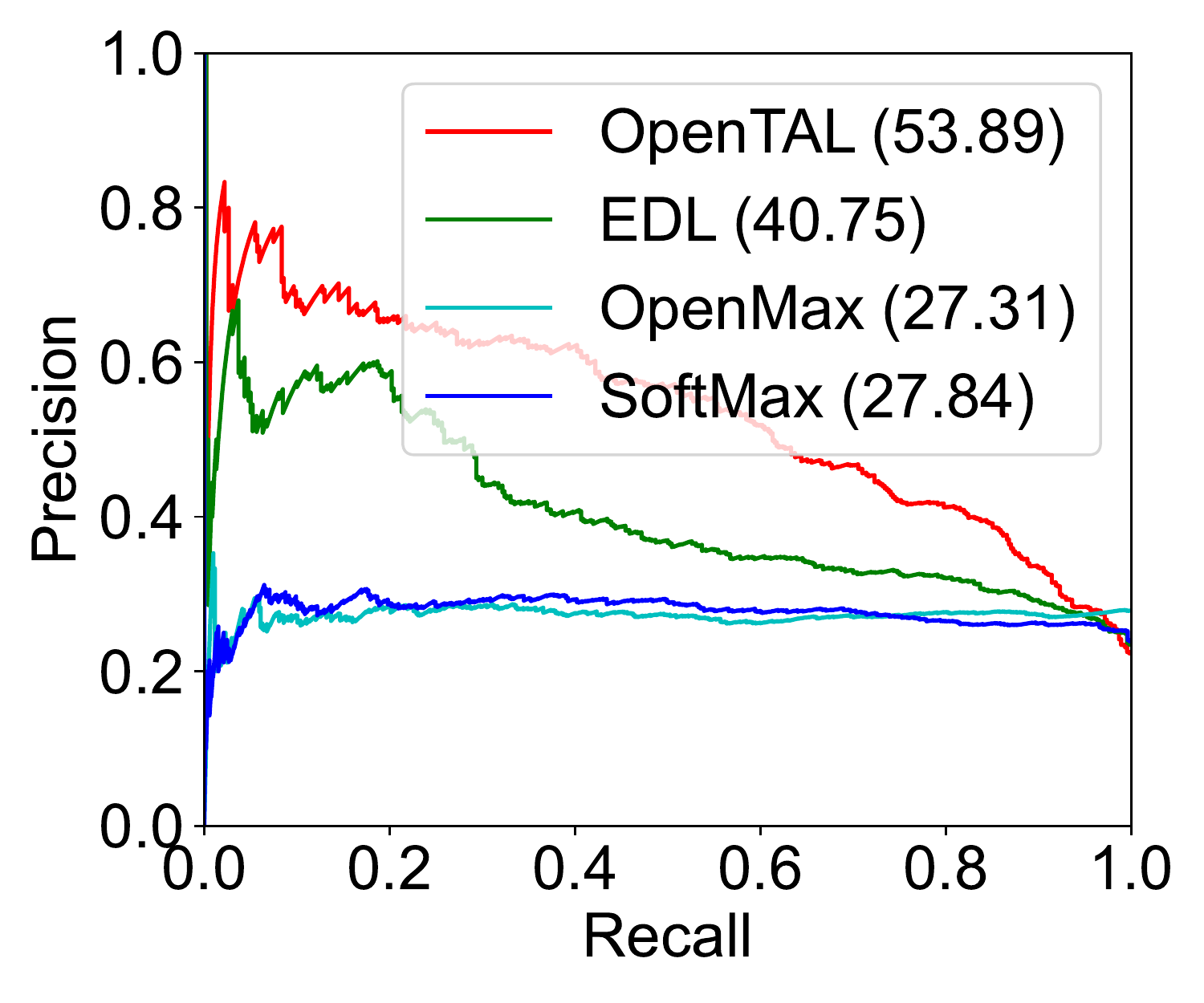}
\\
\parbox[c]{3mm}{\multirow{1}{*}[6.0em]{\rotatebox[origin=c]{90}{\textbf{split 2}}}} &
\includegraphics[width=\framewidth]{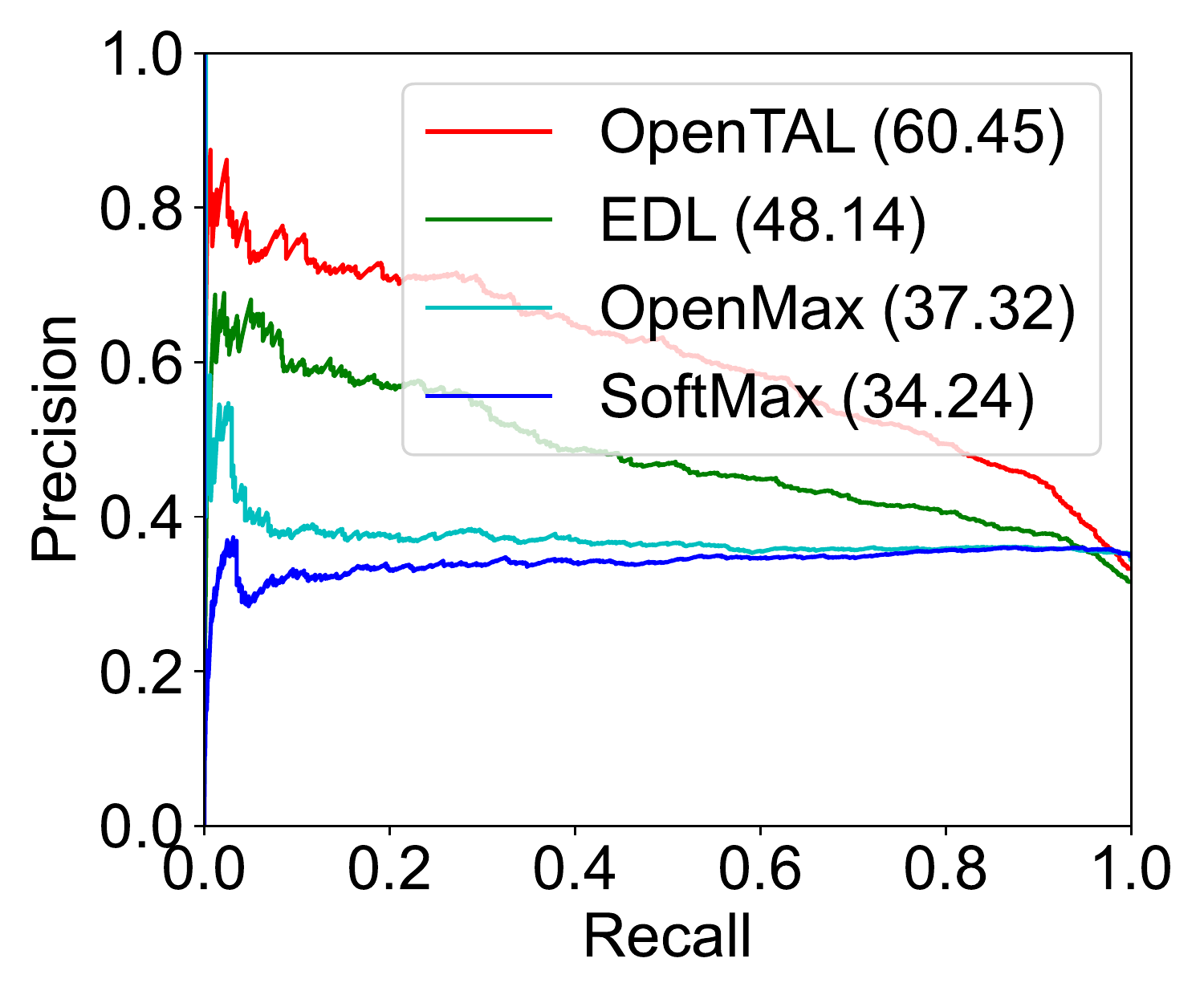} &
\includegraphics[width=\framewidth]{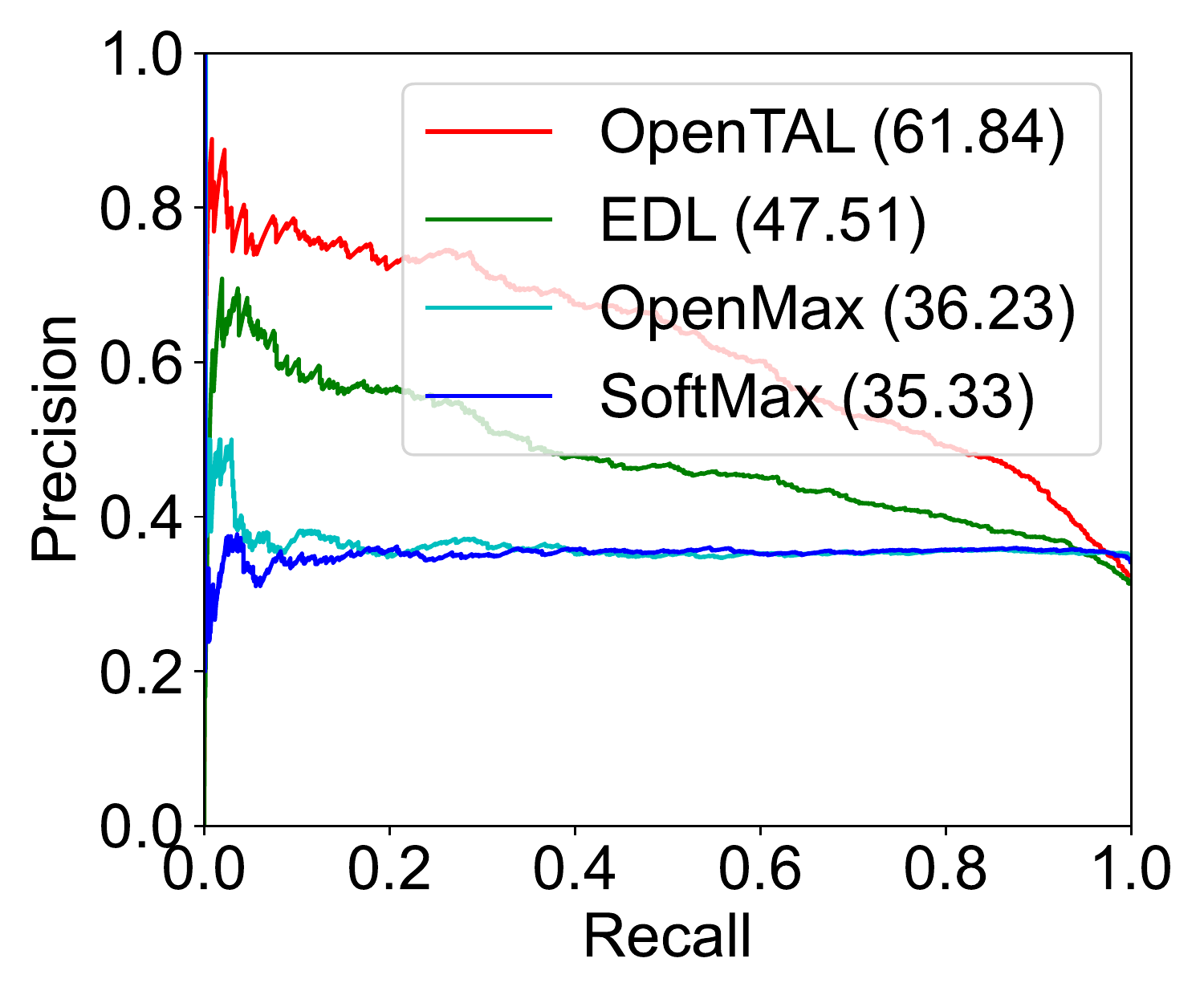} &
\includegraphics[width=\framewidth]{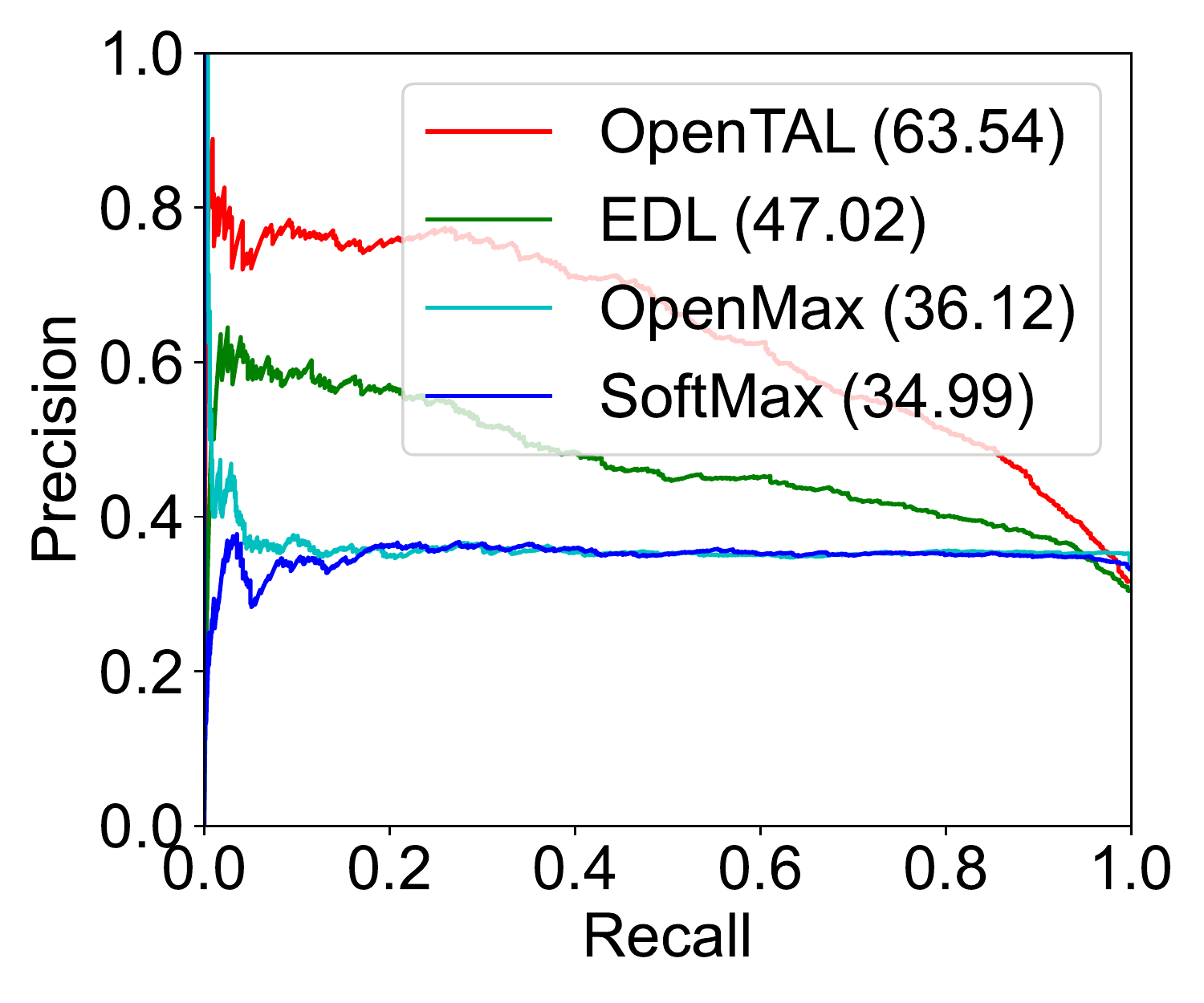} &
\includegraphics[width=\framewidth]{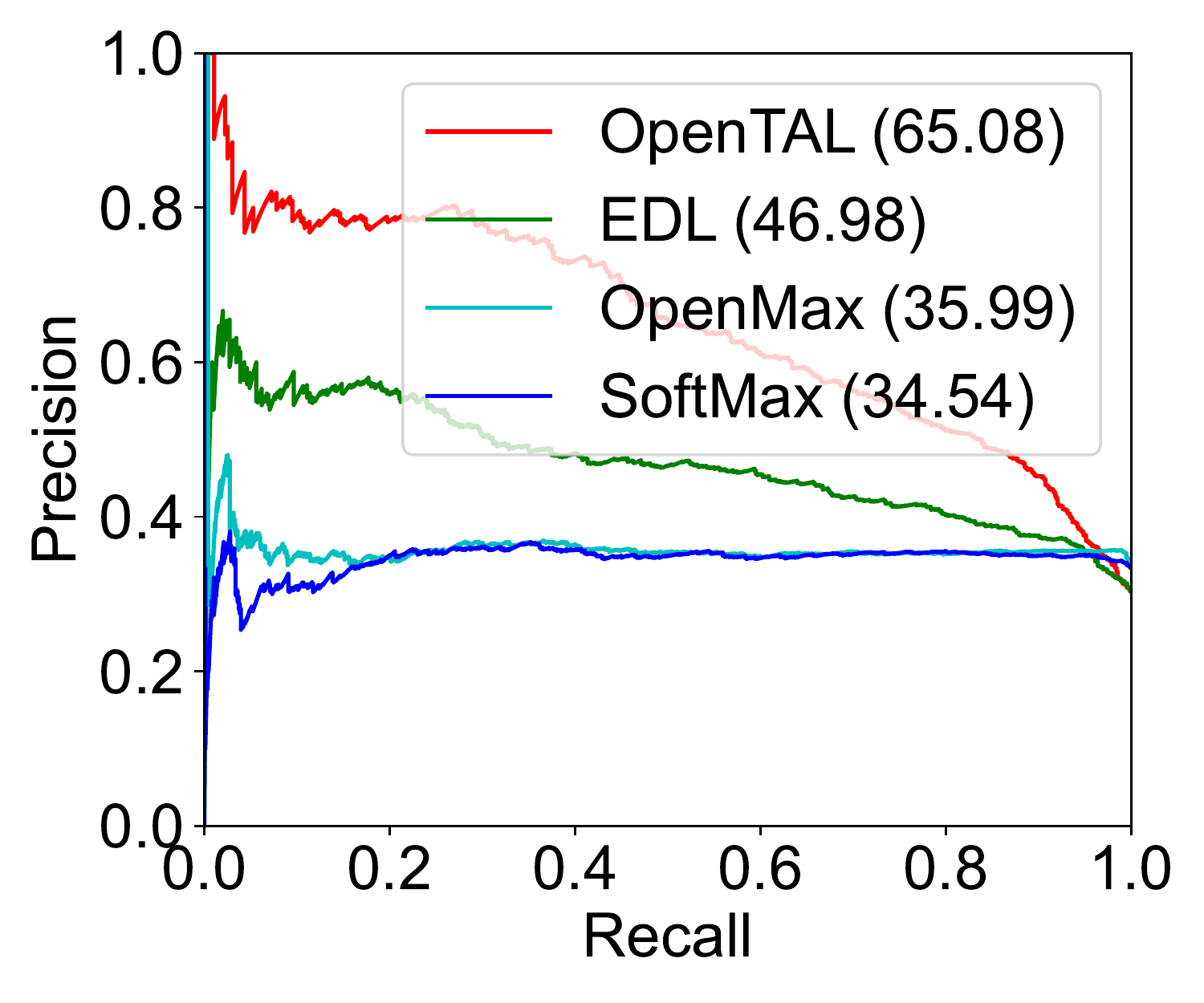} &
\includegraphics[width=\framewidth]{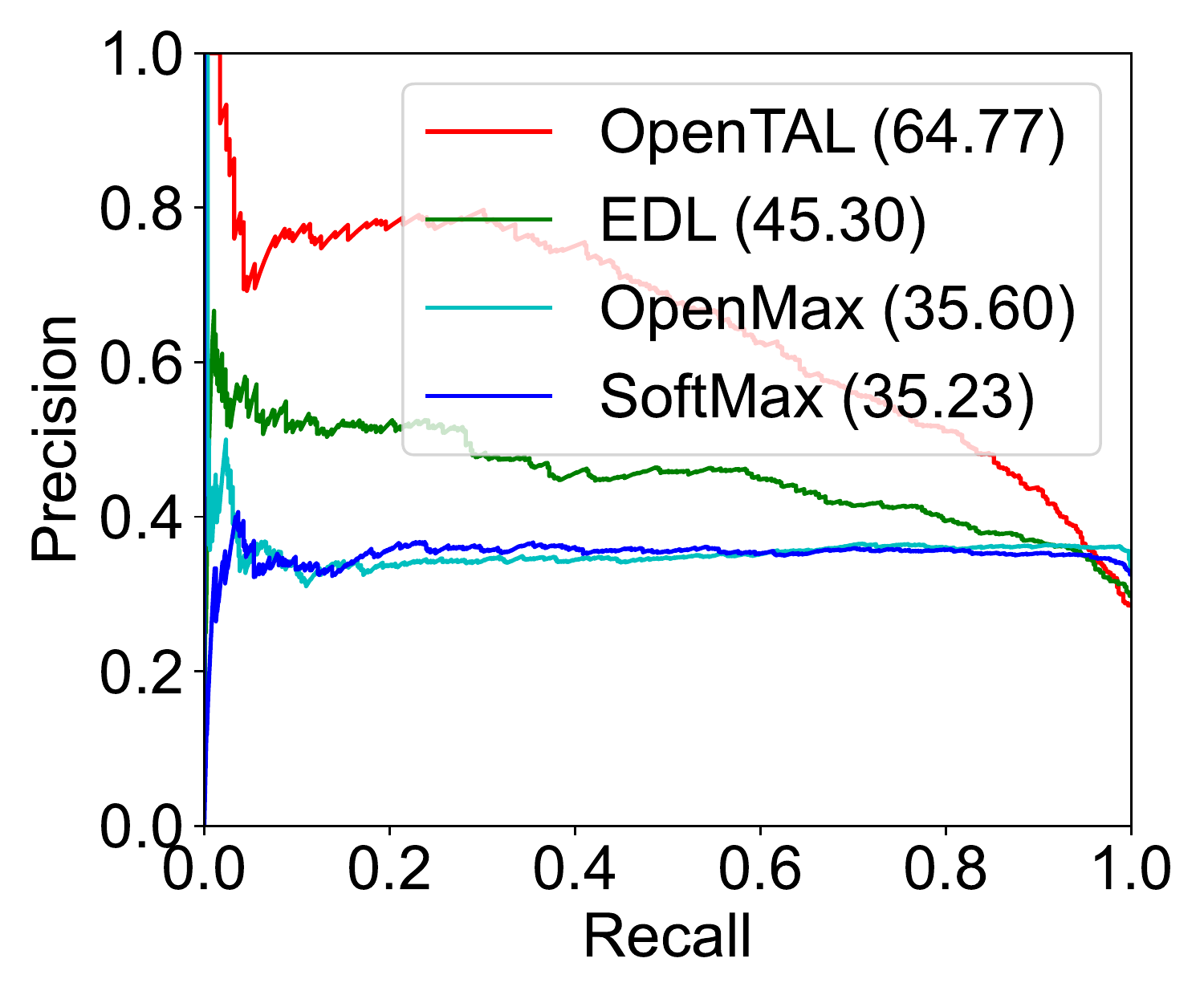}
\\
\parbox[c]{3mm}{\multirow{1}{*}[6.0em]{\rotatebox[origin=c]{90}{\textbf{split 3}}}} &
\includegraphics[width=\framewidth]{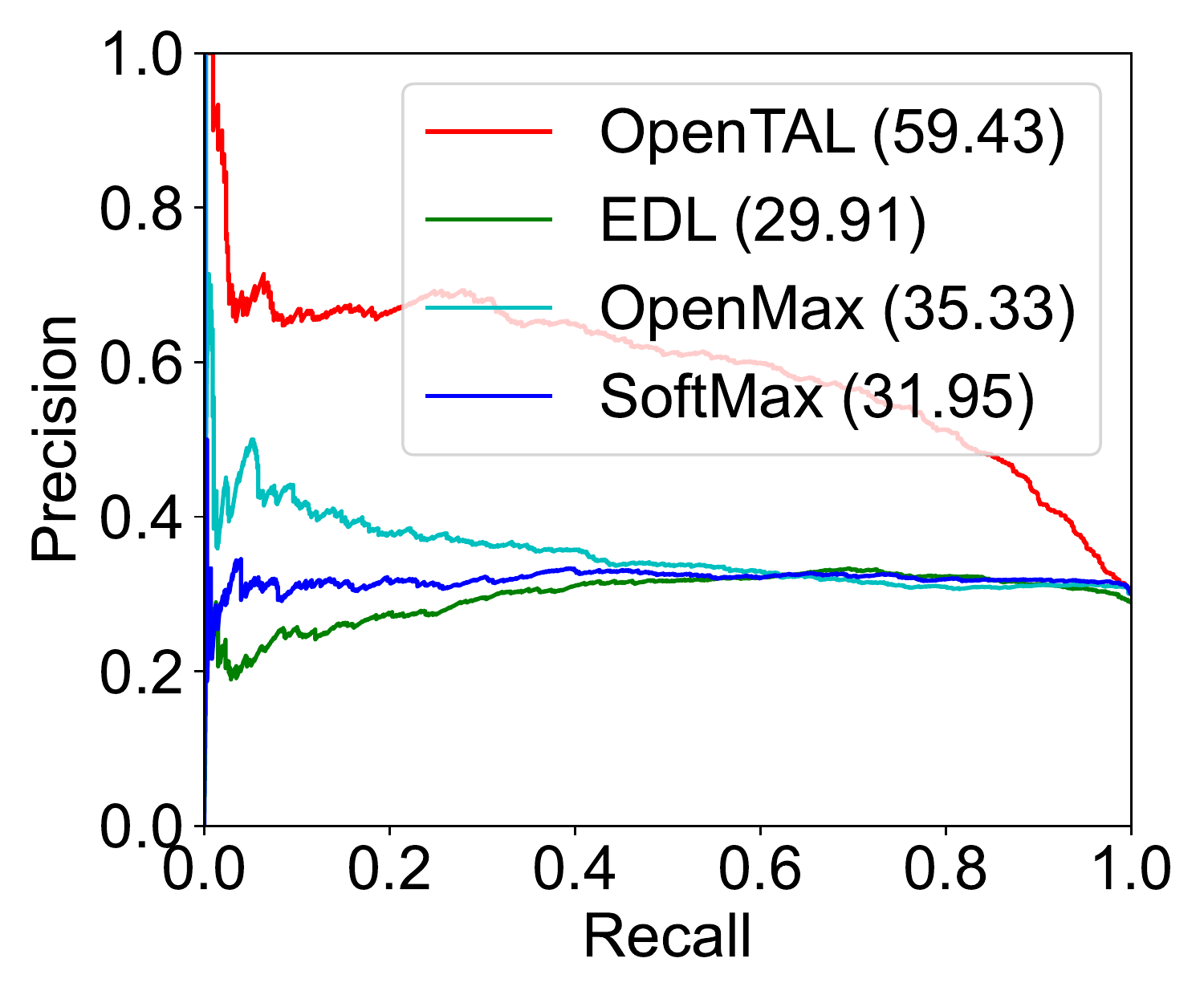} &
\includegraphics[width=\framewidth]{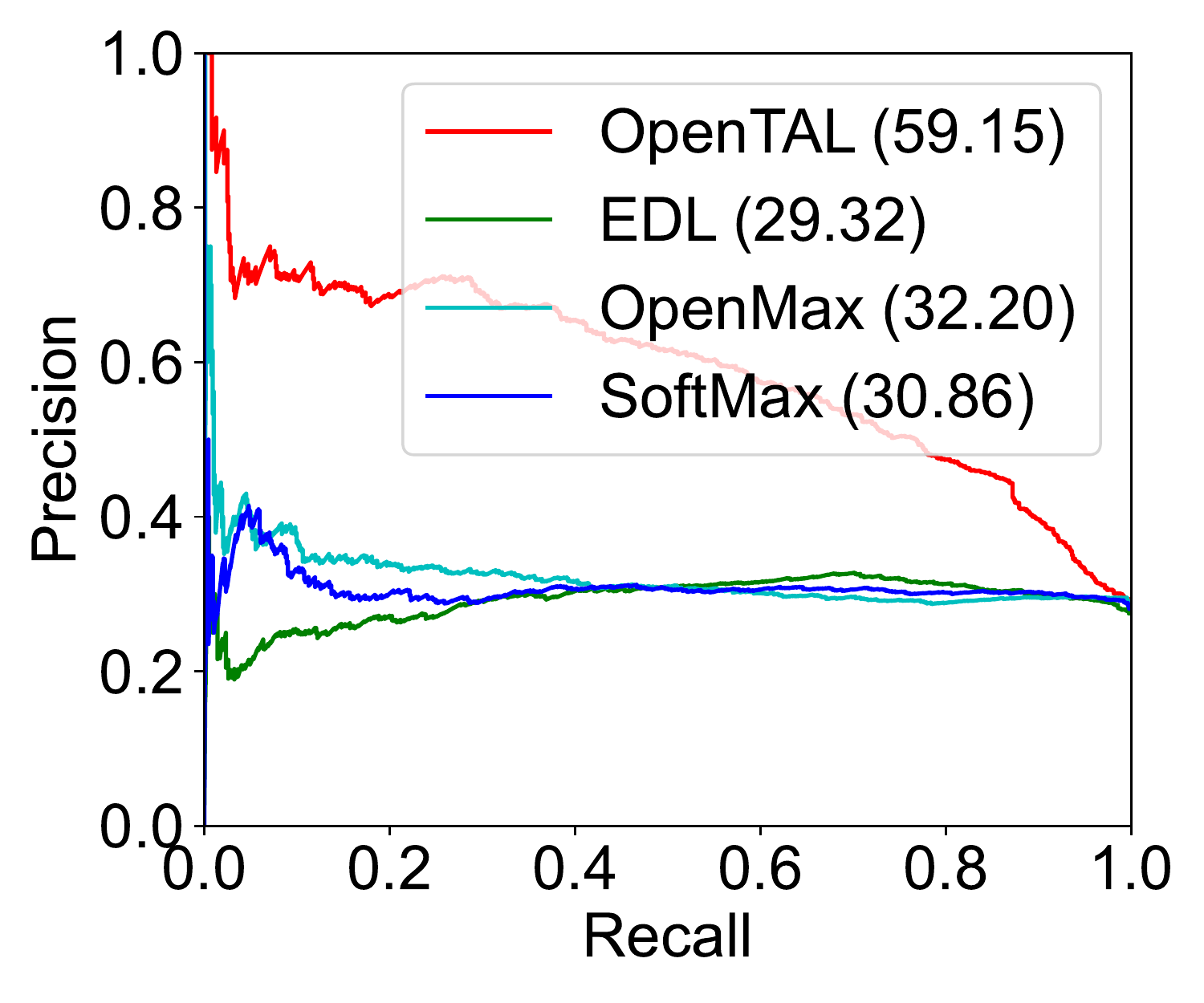} &
\includegraphics[width=\framewidth]{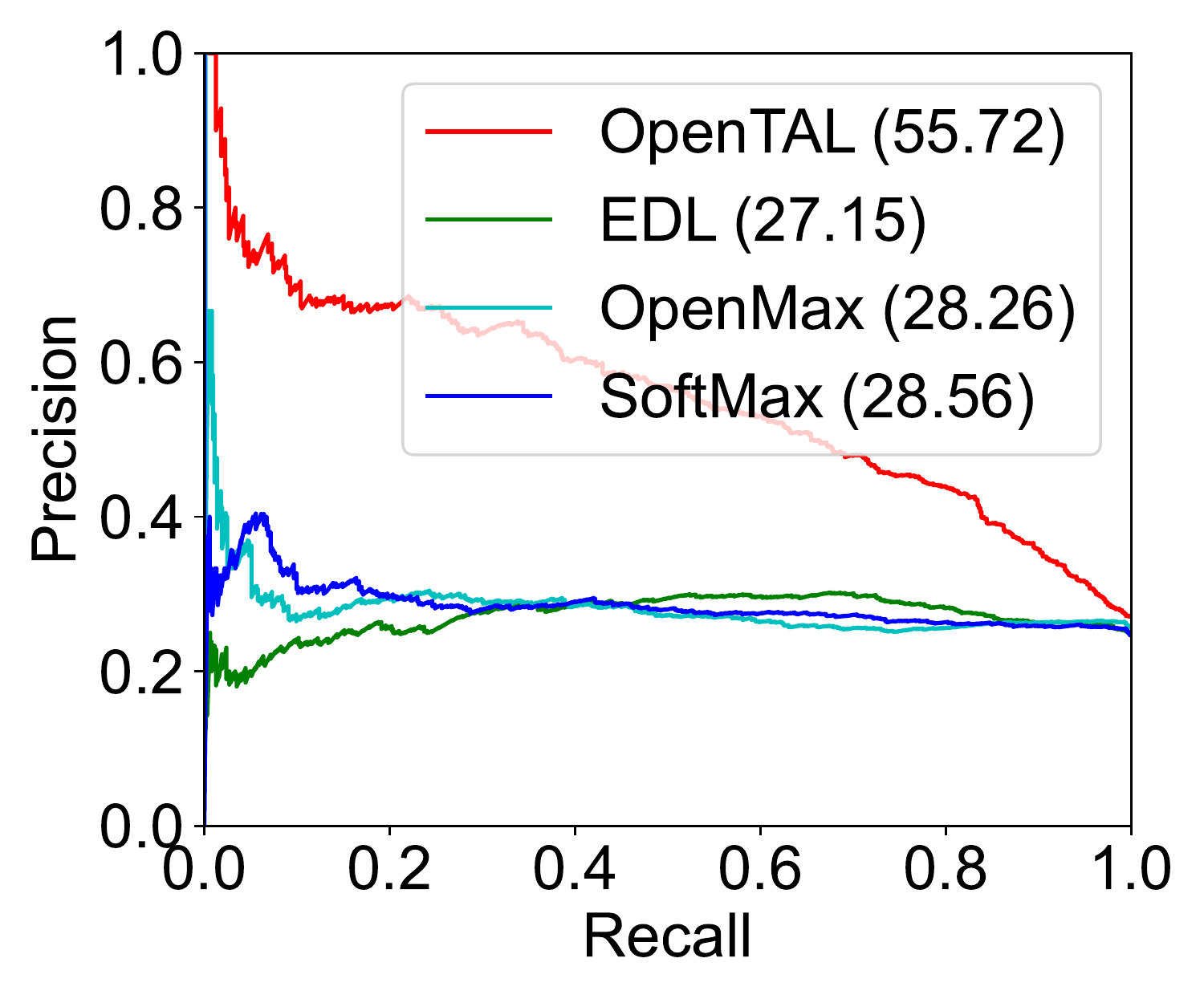} &
\includegraphics[width=\framewidth]{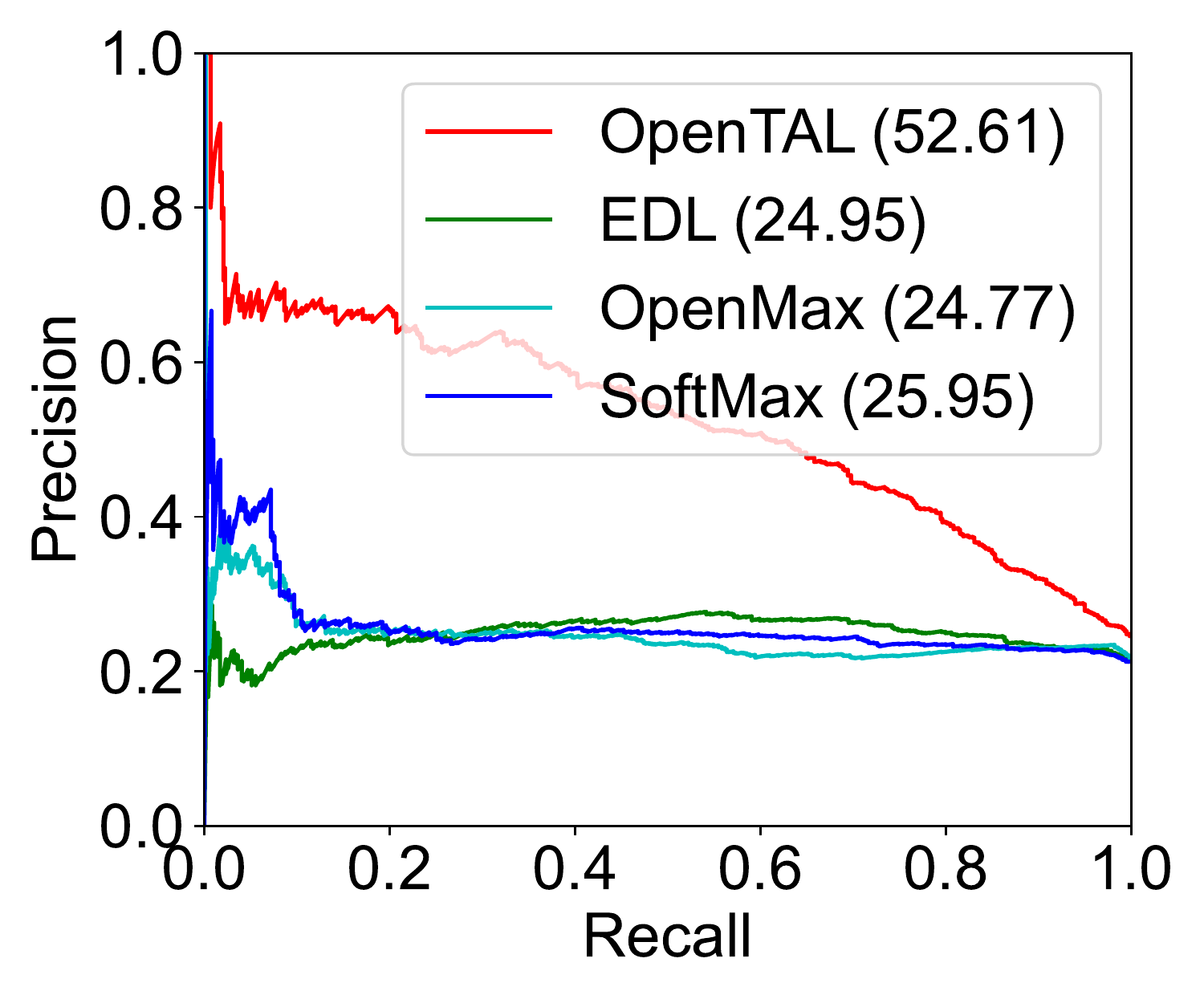} &
\includegraphics[width=\framewidth]{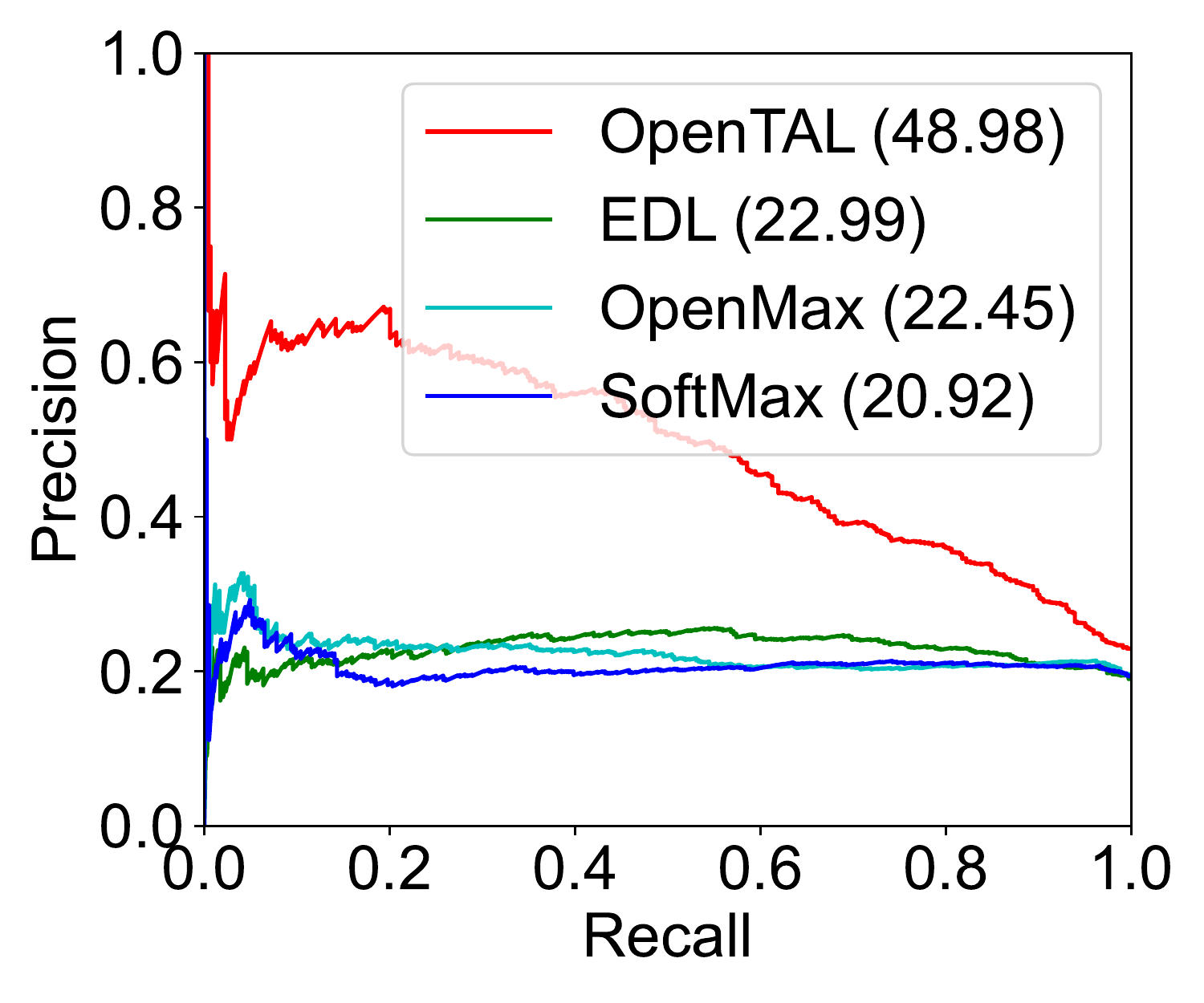}
\\
\end{tabular}
\captionsetup{font=small,aboveskip=3pt}
\caption{\textbf{PR Curves.} These figures show the method comparison by Precision Recall curves on THUMOS14 open set splits. Numbers in parentheses are AUPR values. They show that the~\ul{PR performance varies significantly both across dataset splits and tIoU thresholds}, and our proposed OpenTAL could consistently outperform baselines on all the three splits and five thresholds.}
\label{fig:pr_curves}
\vspace{-10pt}
\end{figure*}

%% file: supp_tex/osdr_curves.tex
\begin{figure*}[t]
\footnotesize
\centering
\renewcommand{\tabcolsep}{0.7pt} %
\begin{tabular}{cccccc}
& \textbf{tIoU=0.3} & \textbf{tIoU=0.4} & \textbf{tIoU=0.5} & \textbf{tIoU=0.6} & \textbf{tIoU=0.7}
\\
\parbox[c]{3mm}{\multirow{1}{*}[6.0em]{\rotatebox[origin=c]{90}{\textbf{split 1}}}} &
\includegraphics[width=\framewidth]{images/supp_curves/OSDR_split0_tiou0.3.pdf} &
\includegraphics[width=\framewidth]{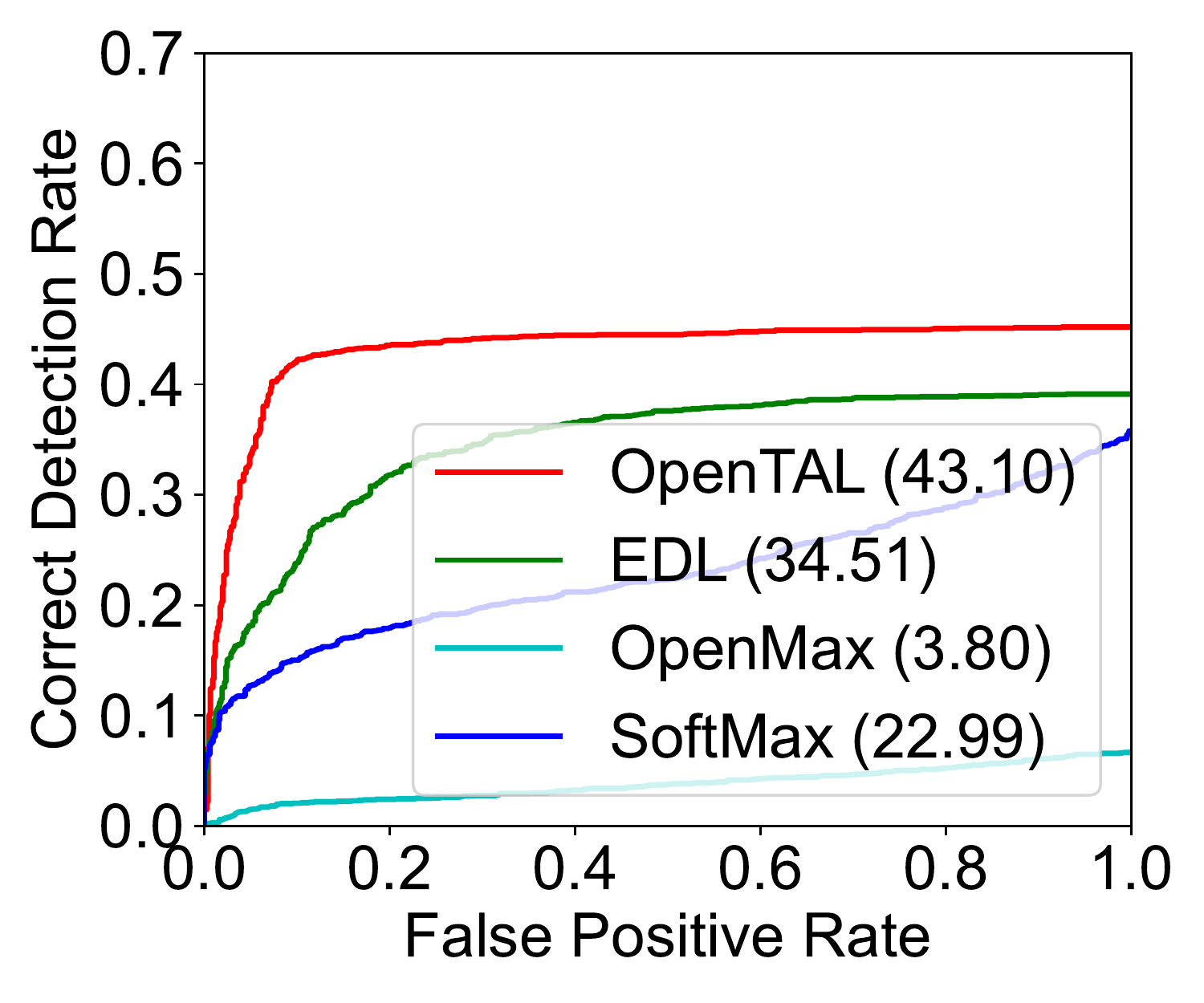} &
\includegraphics[width=\framewidth]{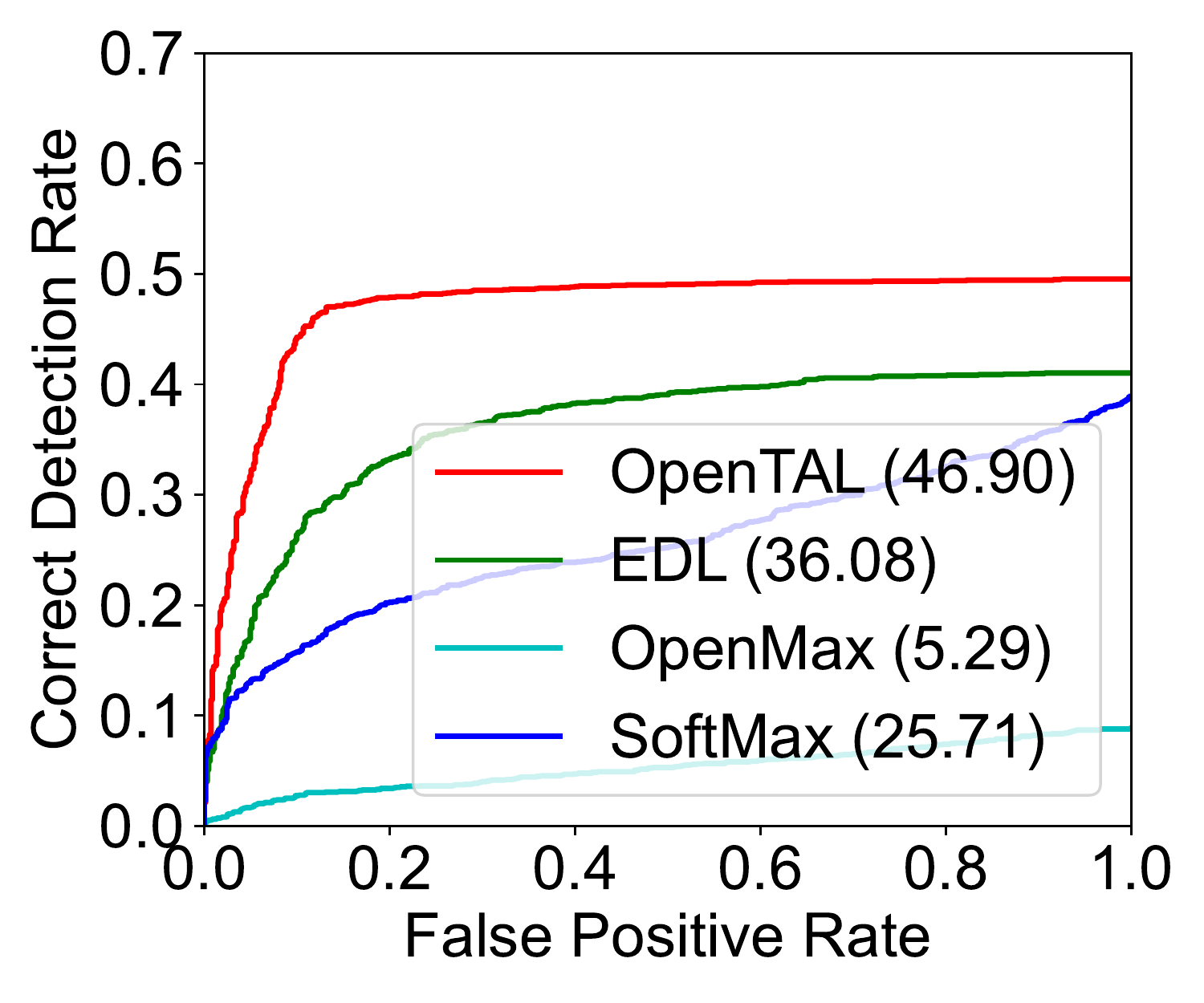} &
\includegraphics[width=\framewidth]{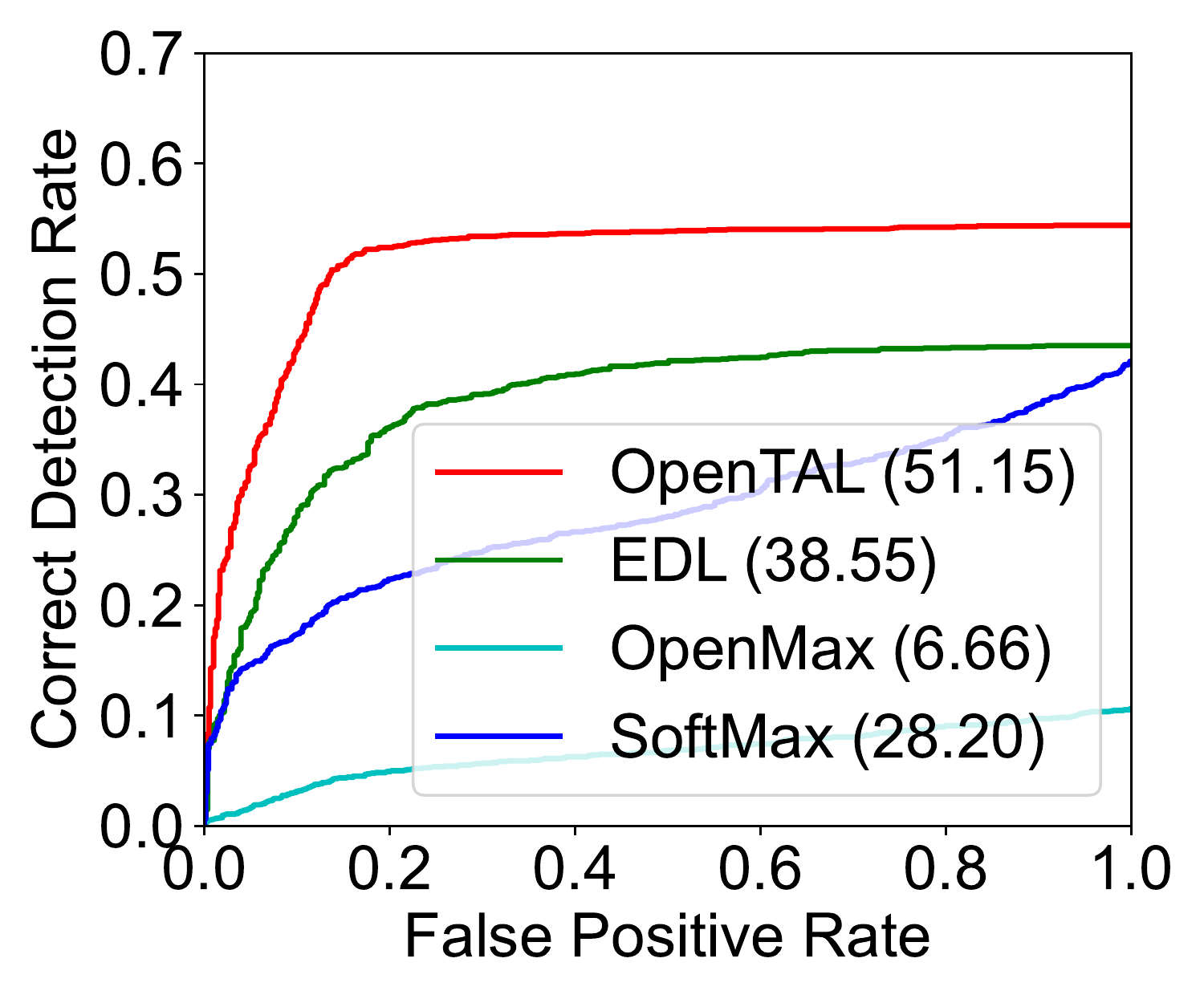} &
\includegraphics[width=\framewidth]{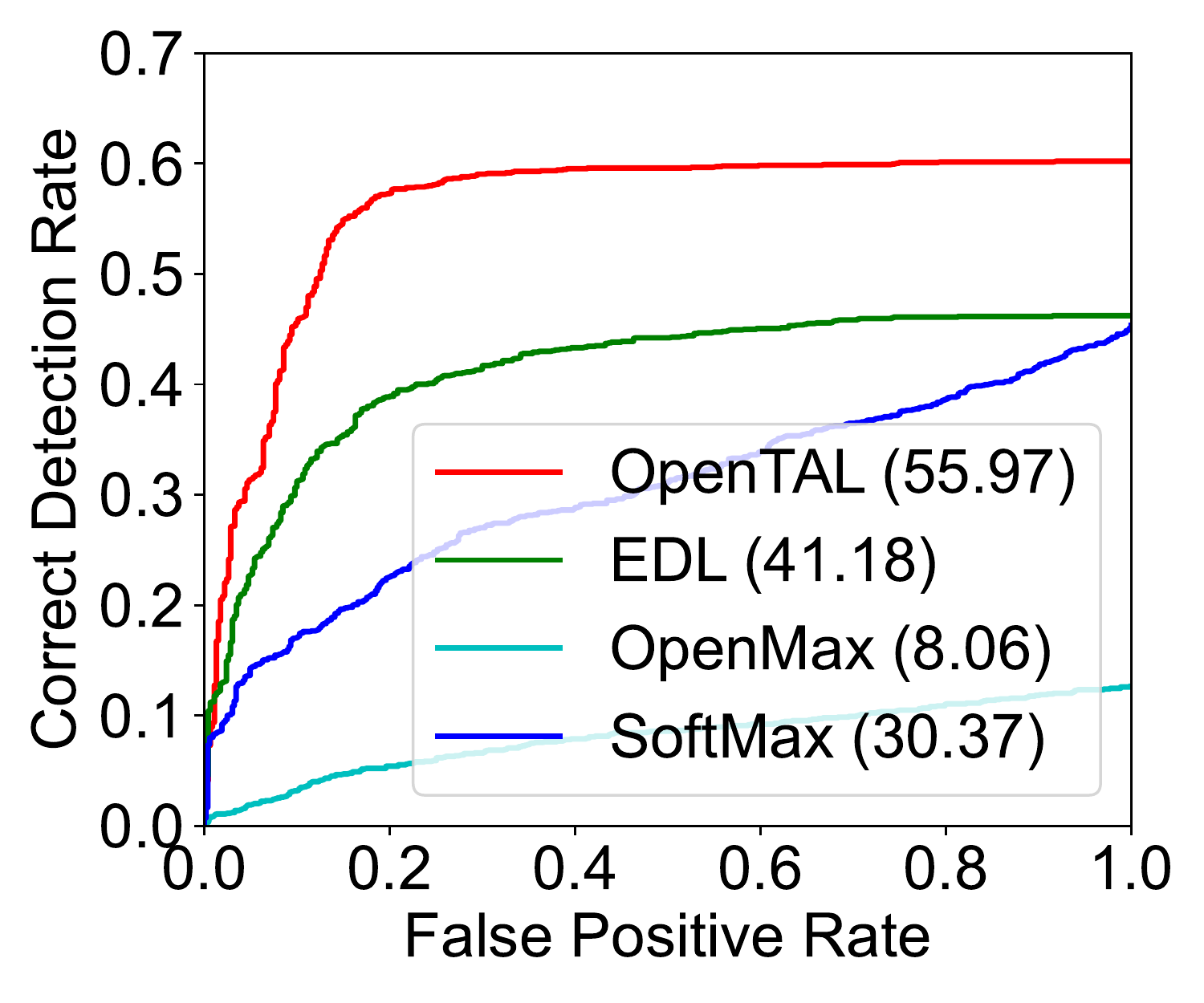}
\\
\parbox[c]{3mm}{\multirow{1}{*}[6.0em]{\rotatebox[origin=c]{90}{\textbf{split 2}}}} &
\includegraphics[width=\framewidth]{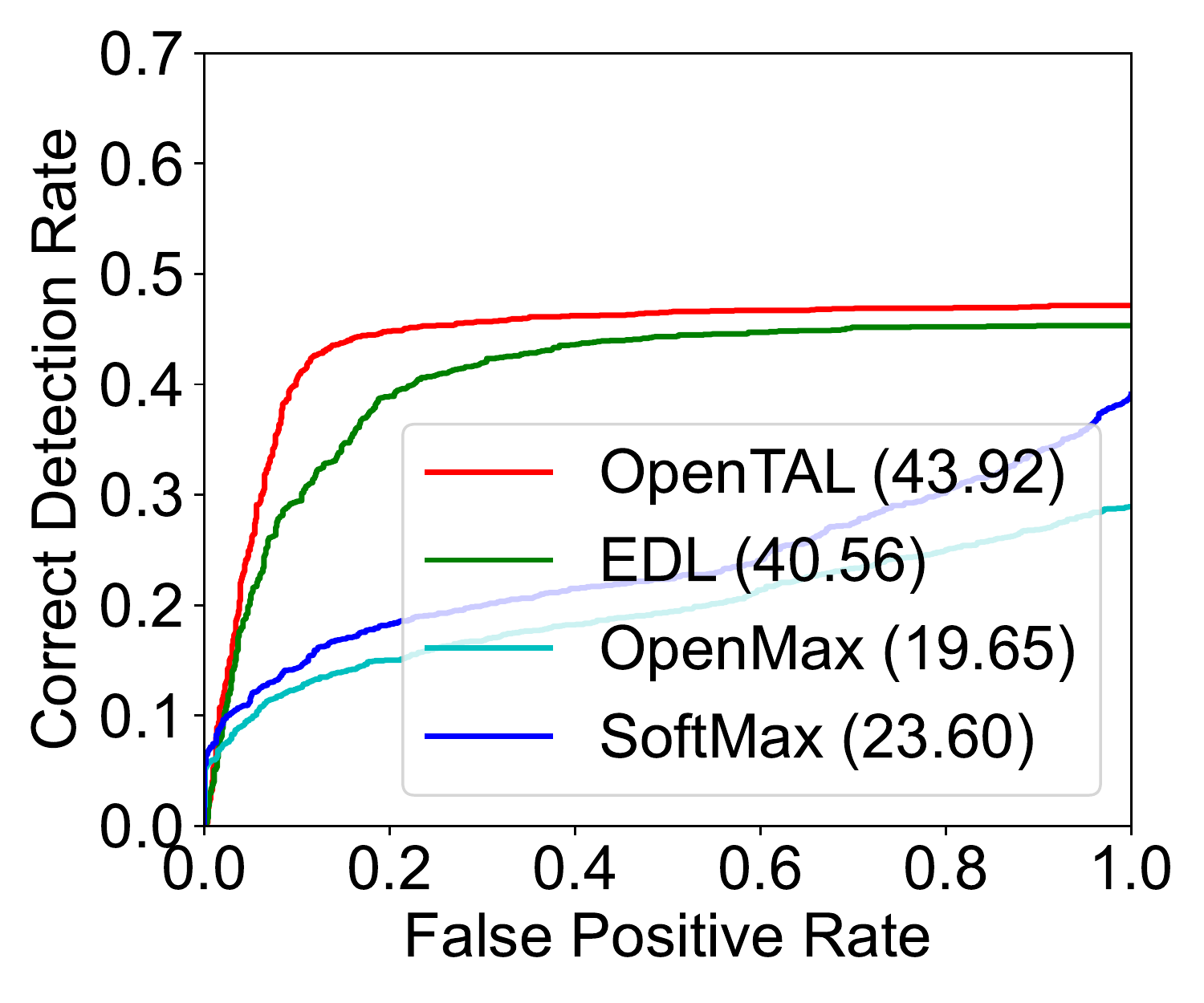} &
\includegraphics[width=\framewidth]{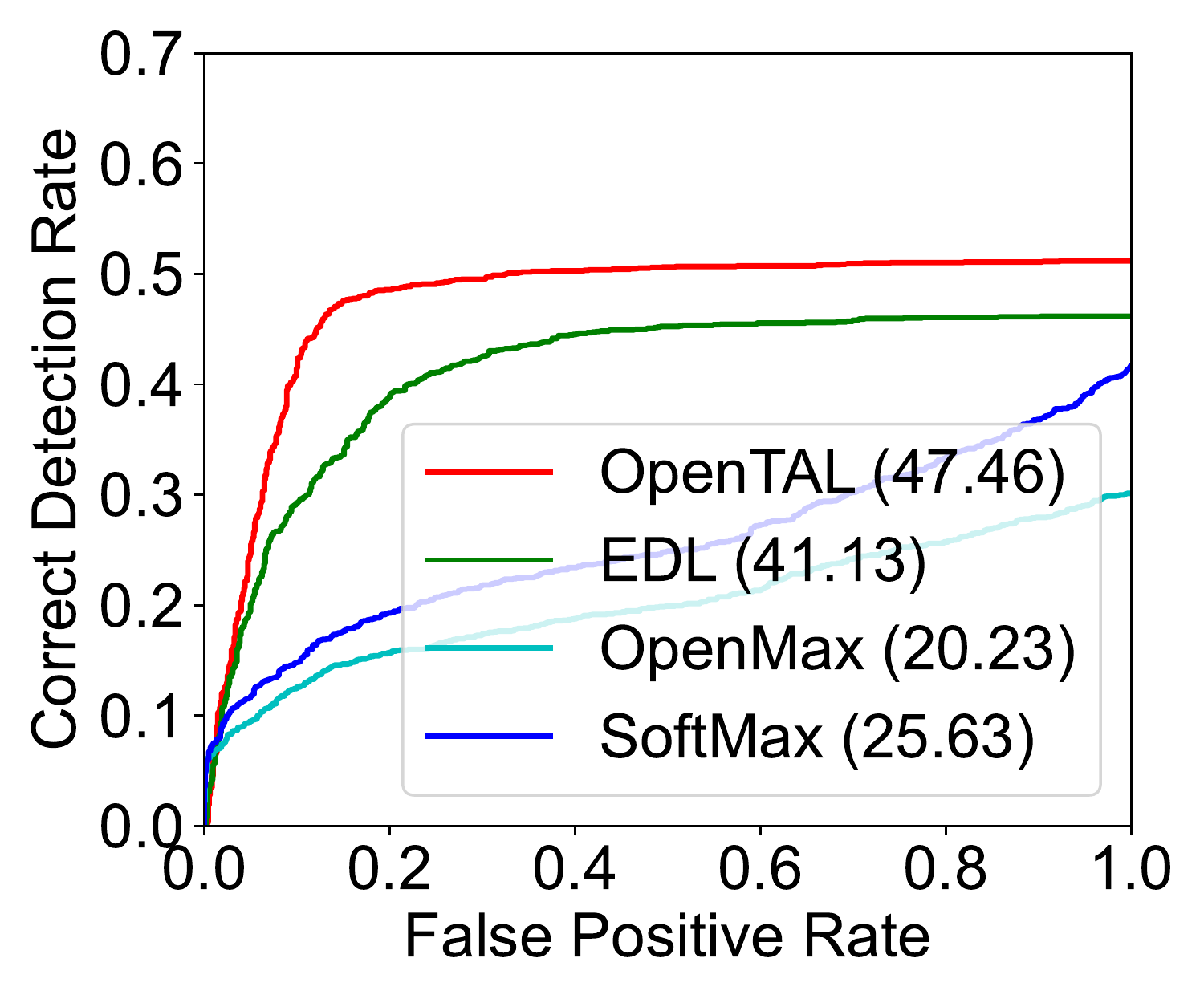} &
\includegraphics[width=\framewidth]{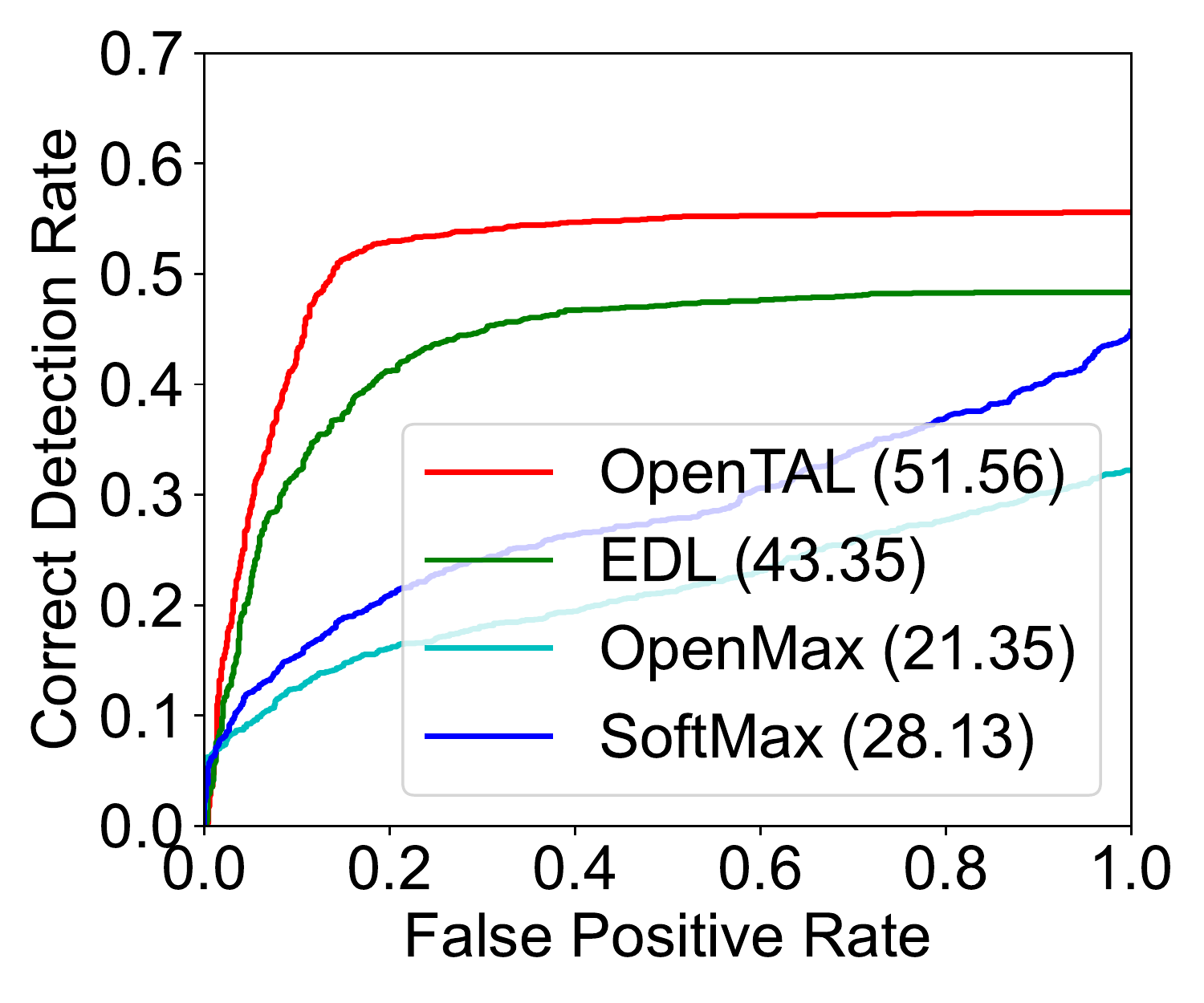} &
\includegraphics[width=\framewidth]{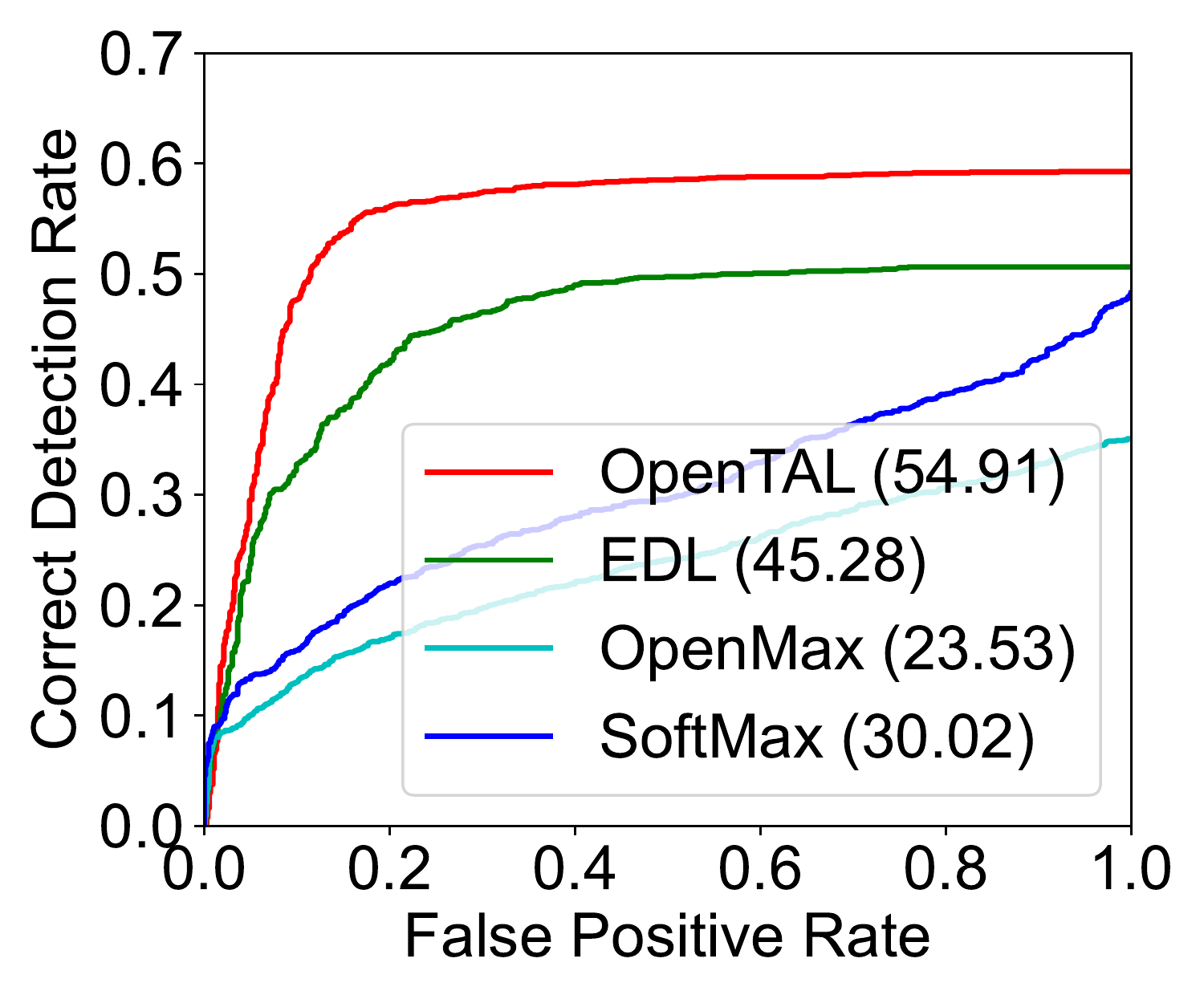} &
\includegraphics[width=\framewidth]{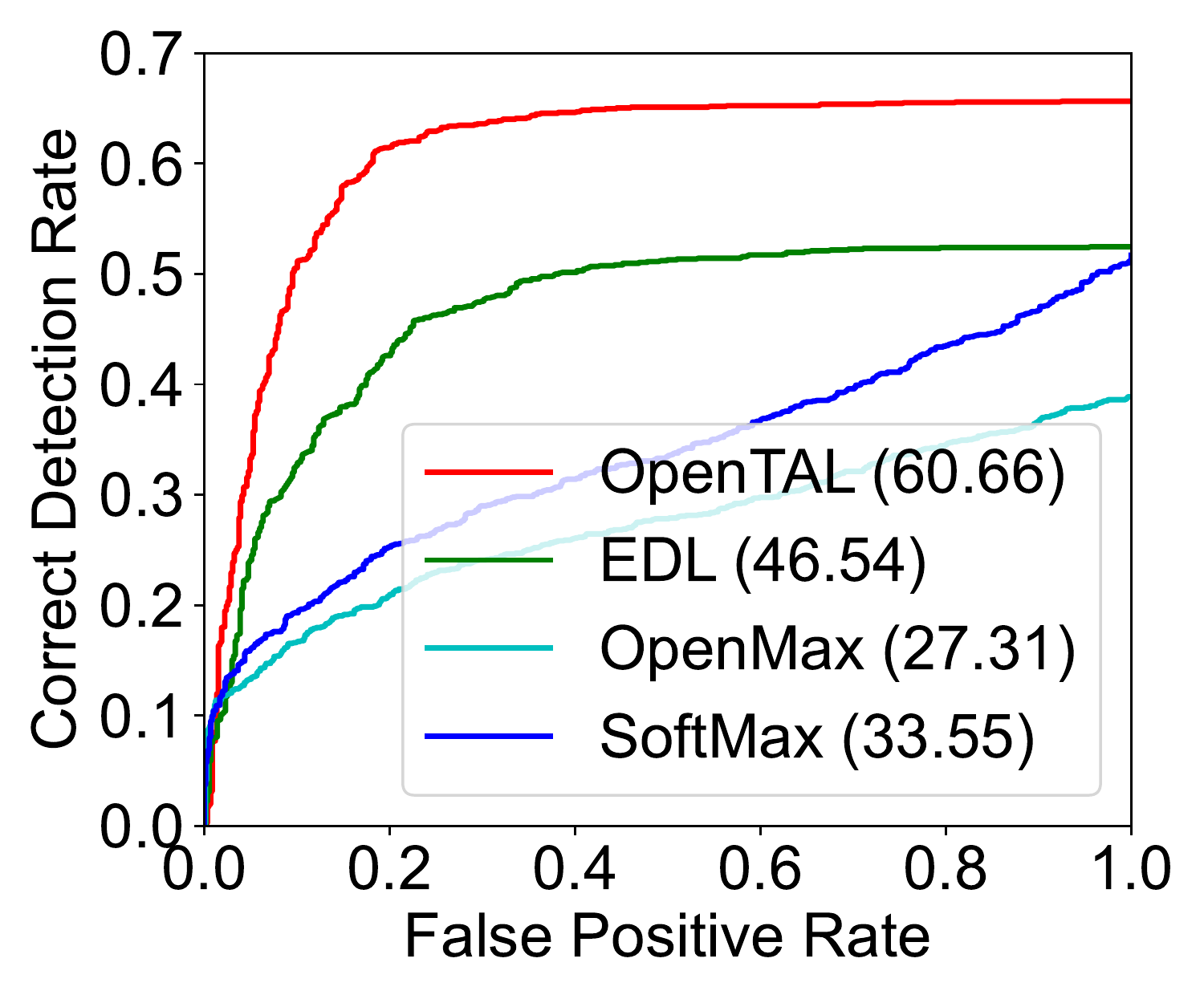}
\\
\parbox[c]{3mm}{\multirow{1}{*}[6.0em]{\rotatebox[origin=c]{90}{\textbf{split 3}}}} &
\includegraphics[width=\framewidth]{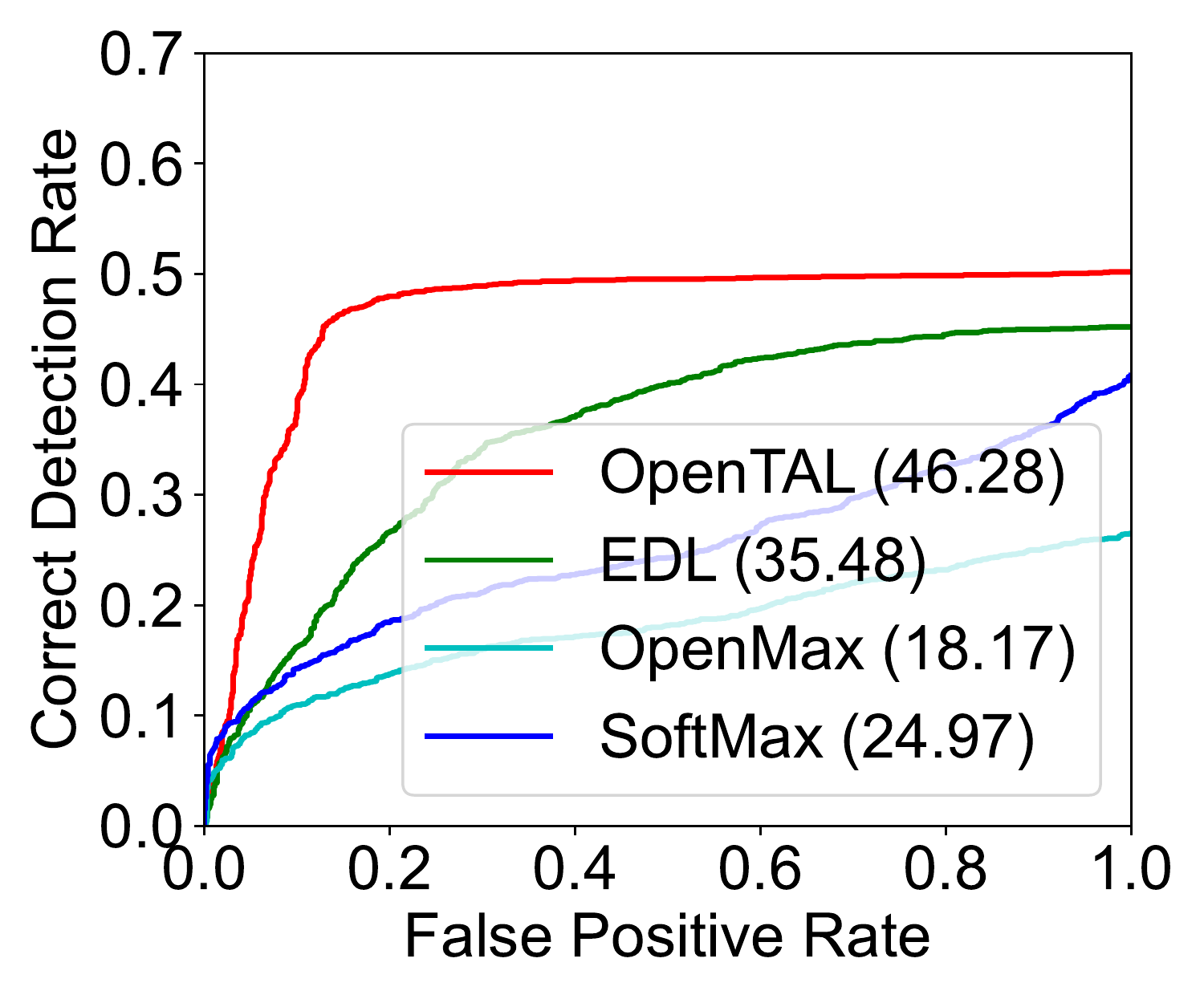} &
\includegraphics[width=\framewidth]{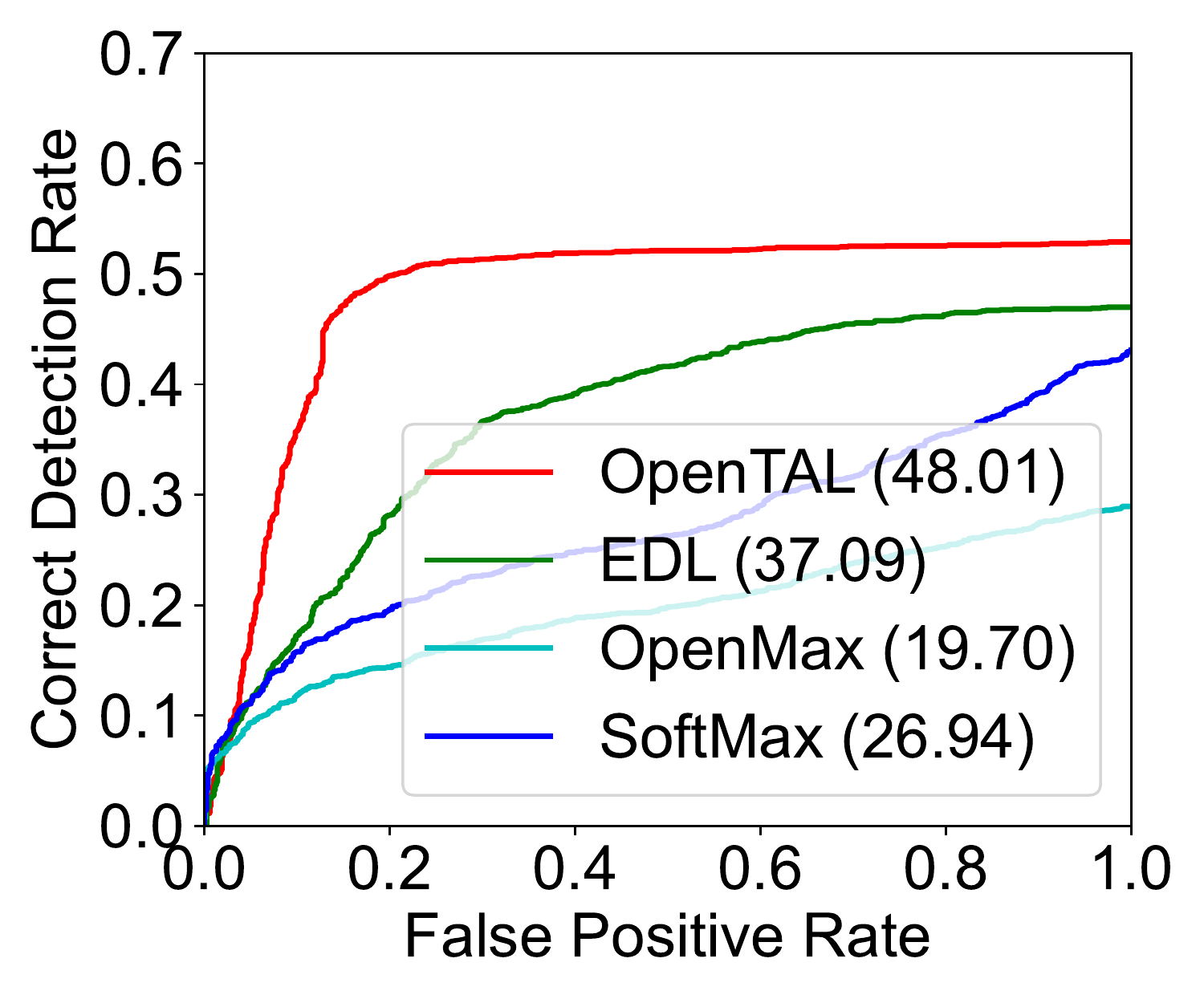} &
\includegraphics[width=\framewidth]{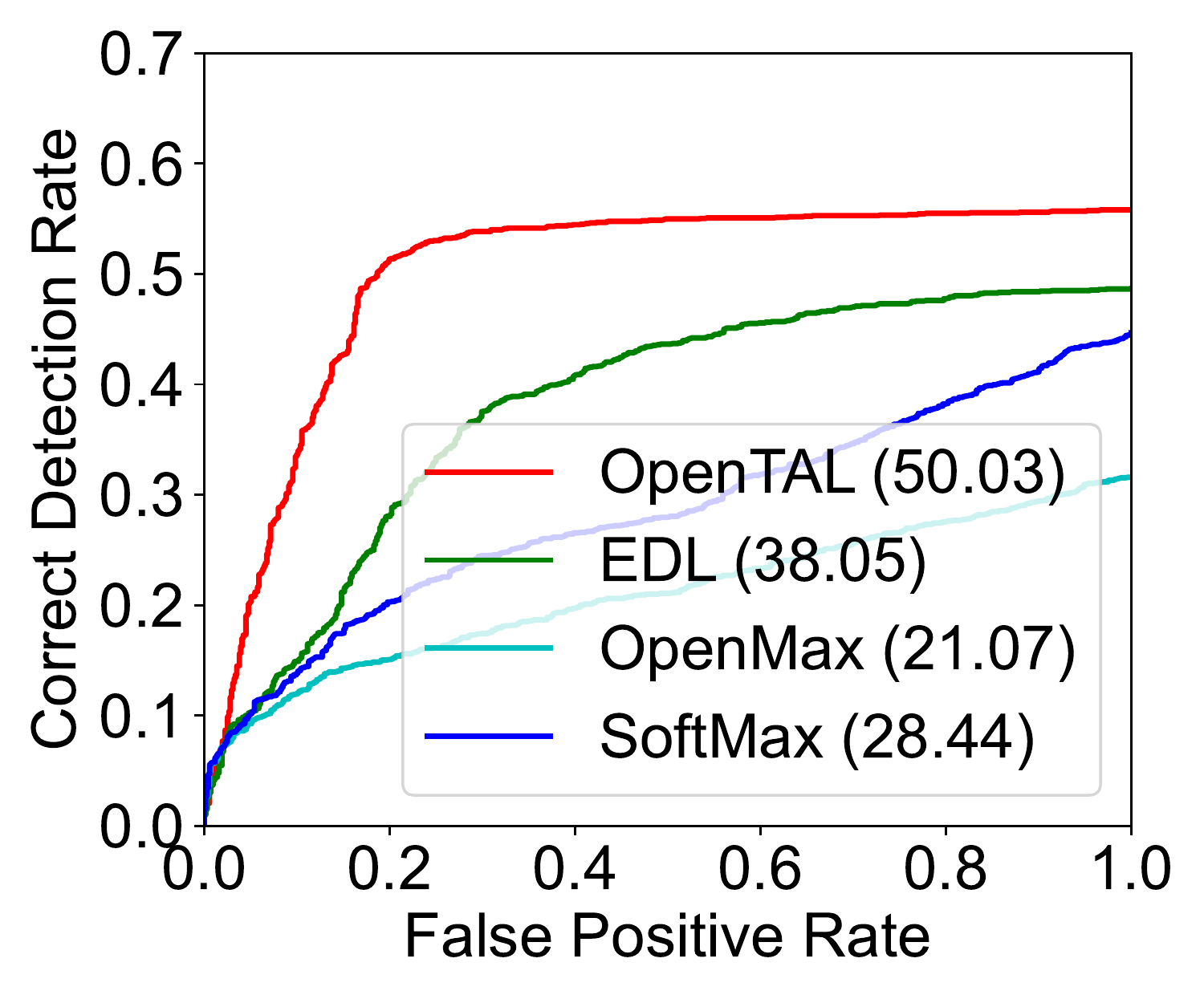} &
\includegraphics[width=\framewidth]{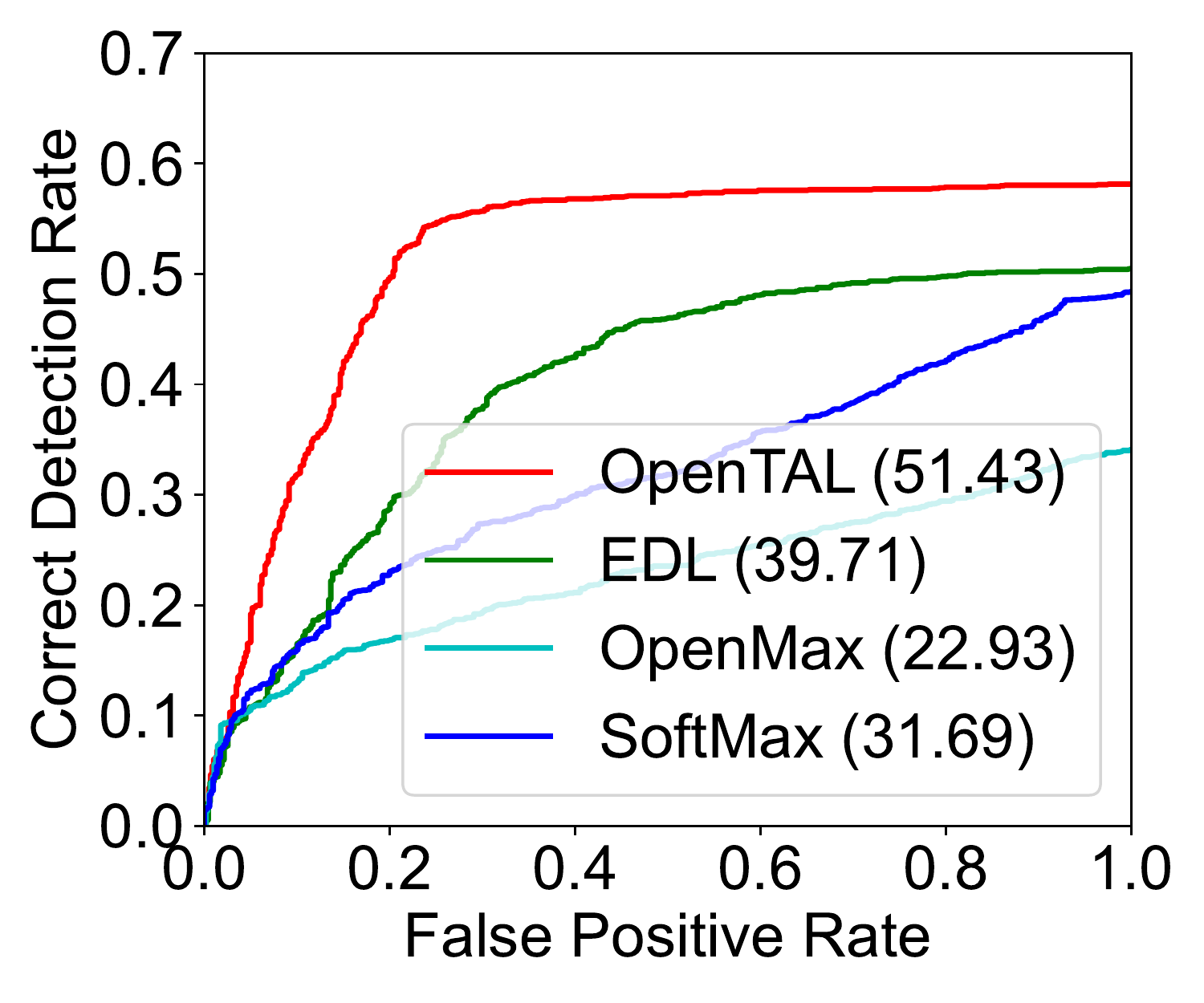} &
\includegraphics[width=\framewidth]{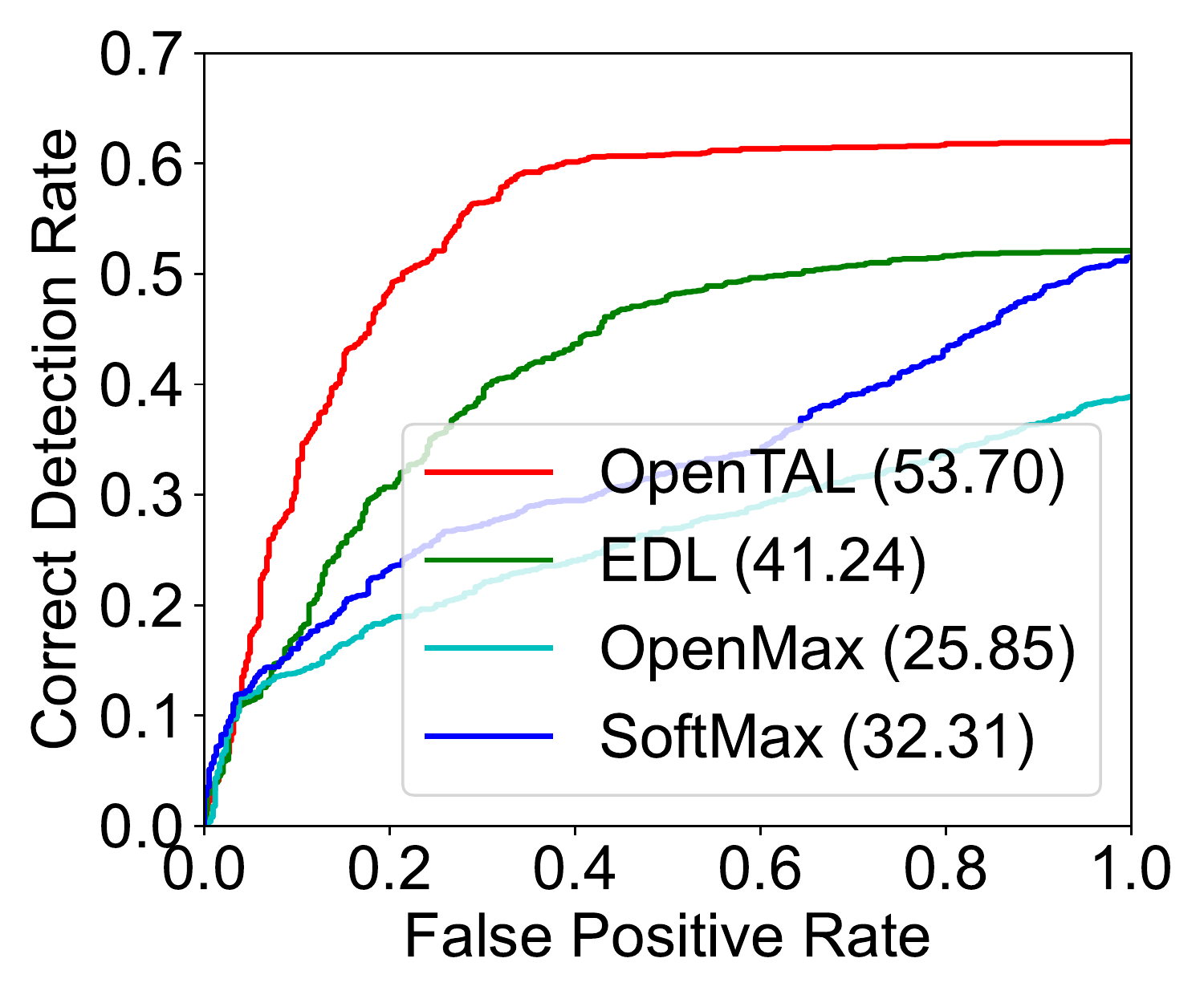}
\\
\end{tabular}
\captionsetup{font=small,aboveskip=3pt}
\caption{\textbf{OSDR Curves.} These figures show the method comparison by OSDR curves on THUMOS14 open set splits. Numbers in parentheses are OSDR values. They show that \ul{the OSDR performance varies significantly both across dataset splits and tIoU thresholds}, and our proposed OpenTAL could consistently outperform baselines on all the three splits and five thresholds.}
\label{fig:osdr_curves}
\vspace{-10pt}
\end{figure*}

%% file: supp_tex/demo.tex
\begin{figure*}[t]
    \centering
    \includegraphics[width=0.495\textwidth]{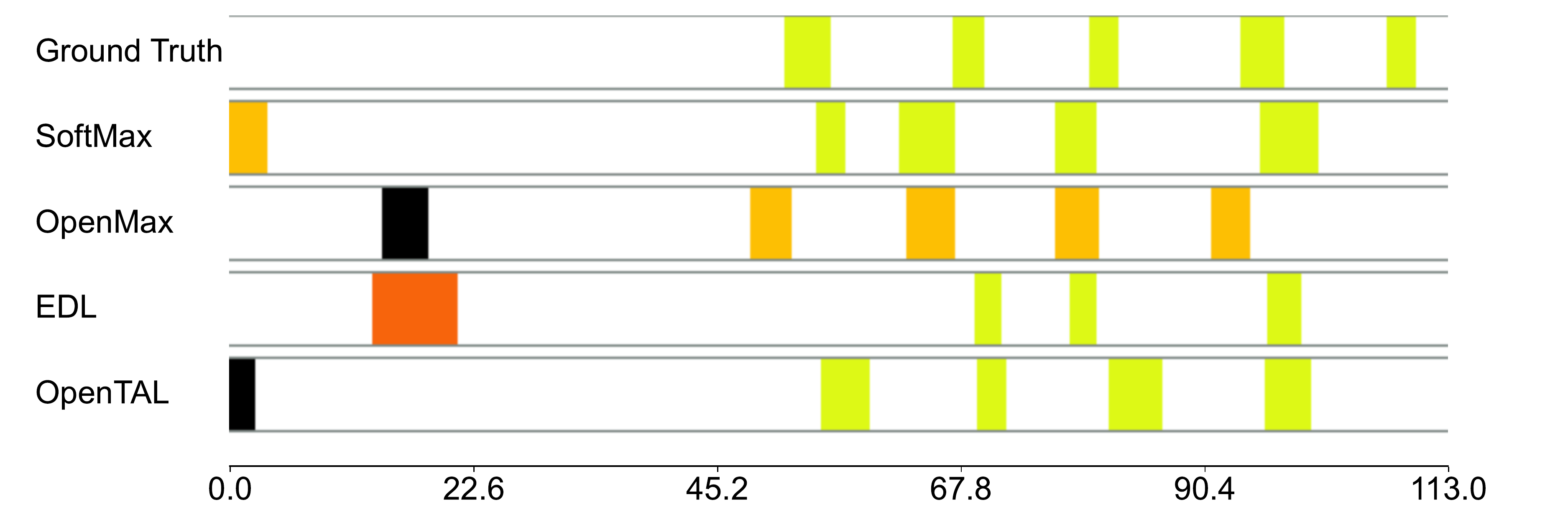}
    \includegraphics[width=0.495\textwidth]{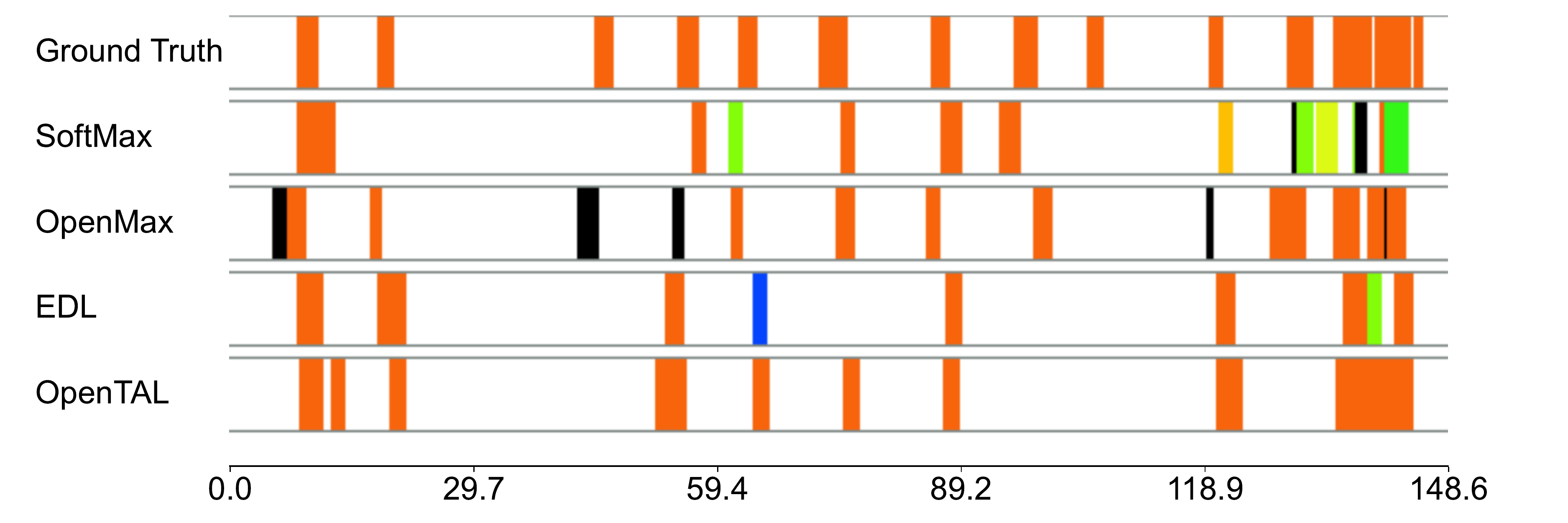}
    \includegraphics[width=0.495\textwidth]{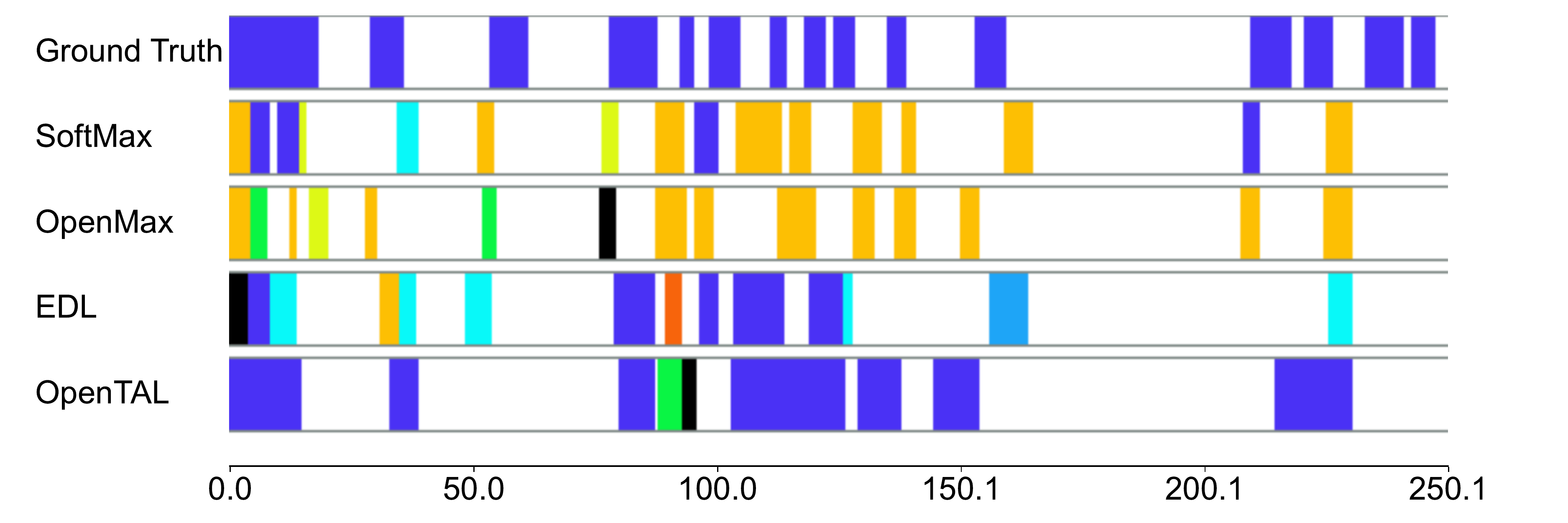}
    \includegraphics[width=0.495\textwidth]{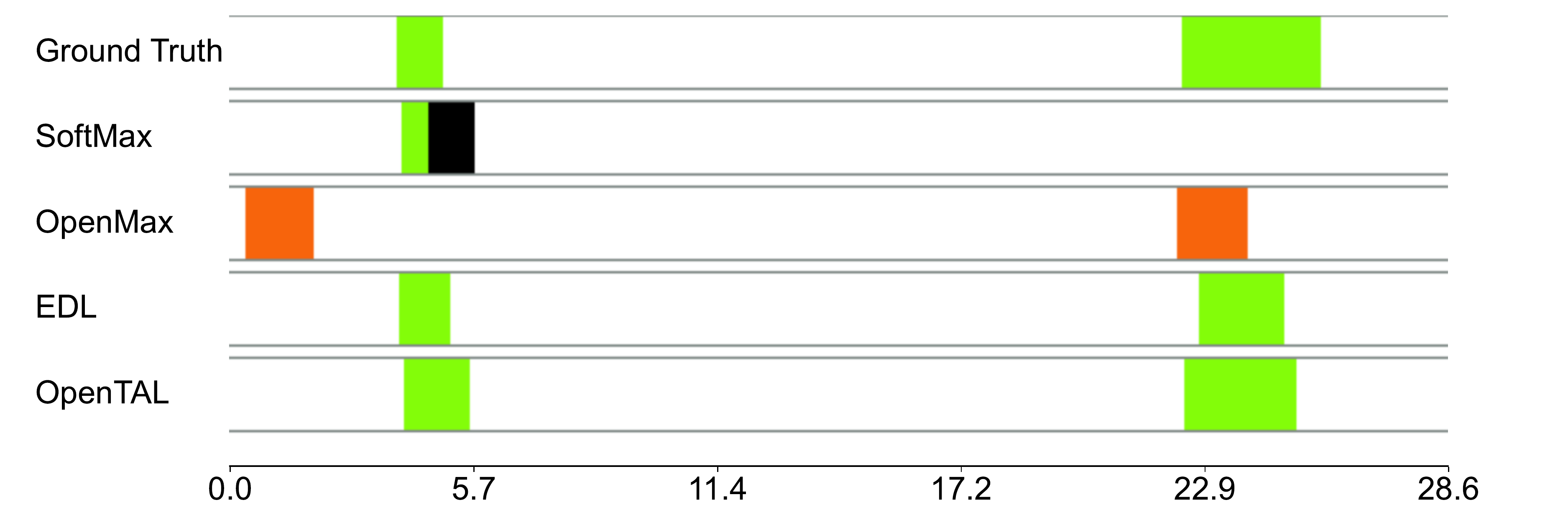}
    \includegraphics[width=0.495\textwidth]{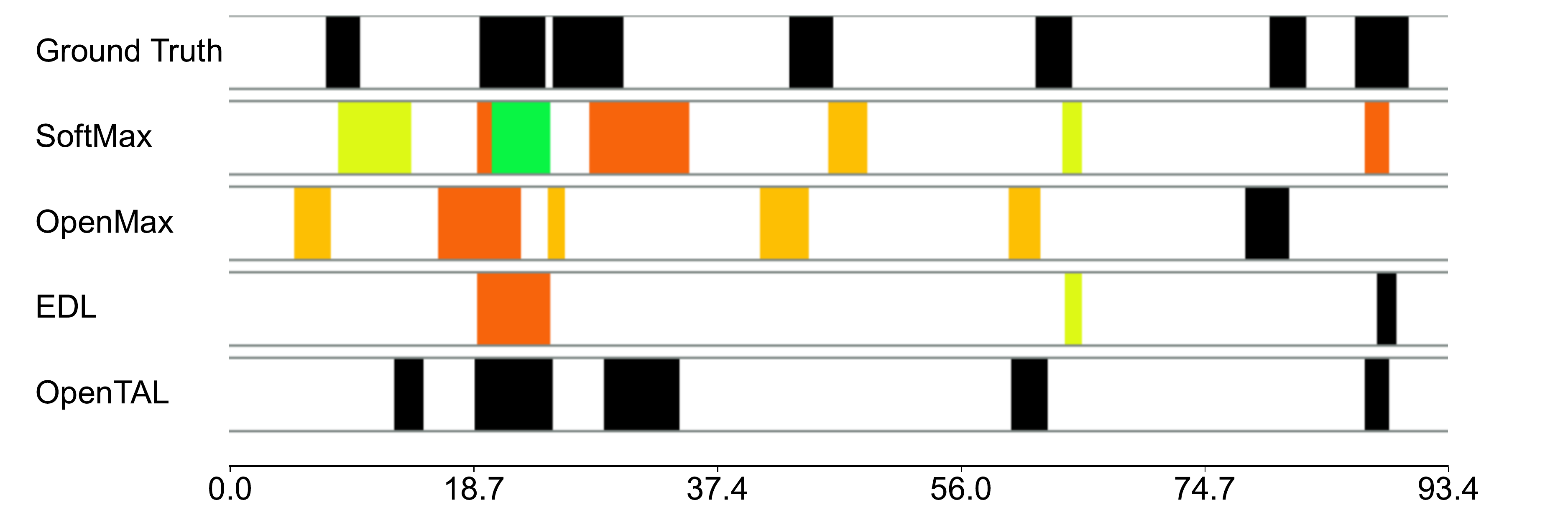}
    \includegraphics[width=0.495\textwidth]{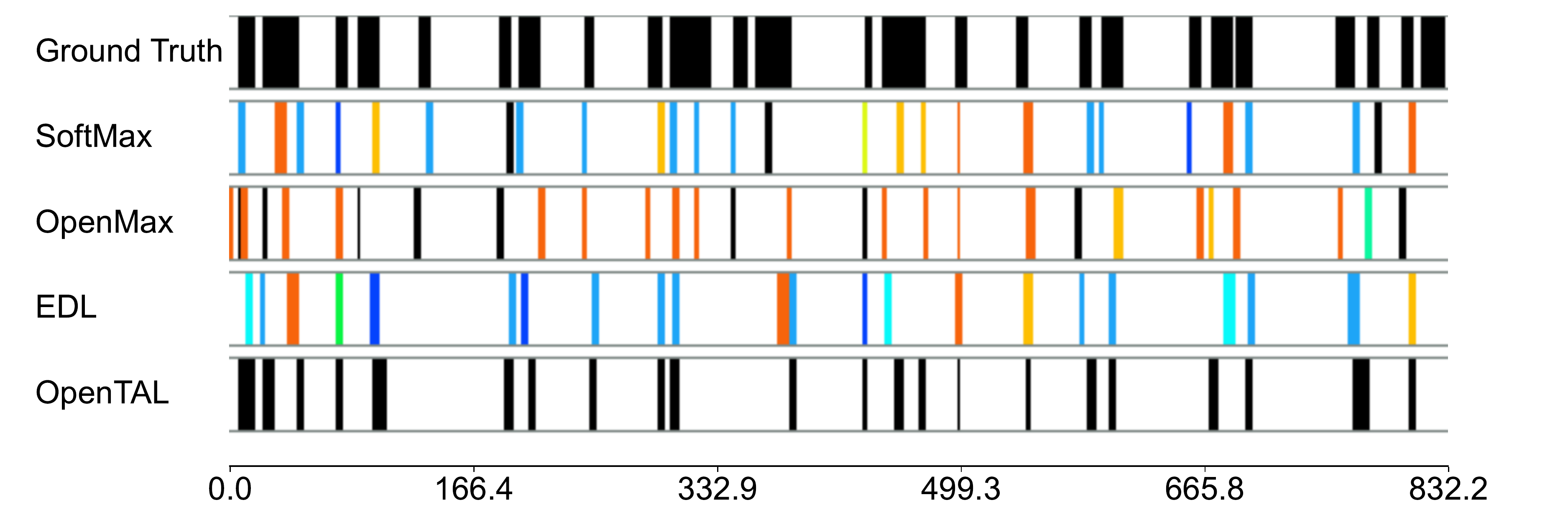}
    \includegraphics[width=0.495\textwidth]{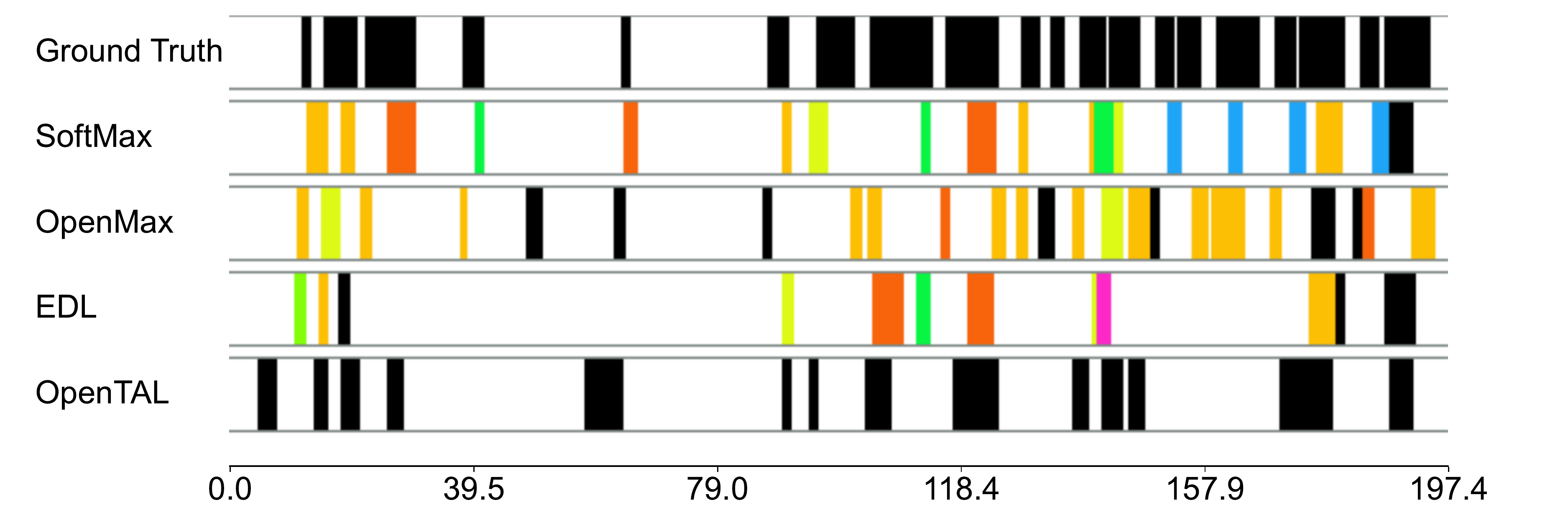}
    \includegraphics[width=0.495\textwidth]{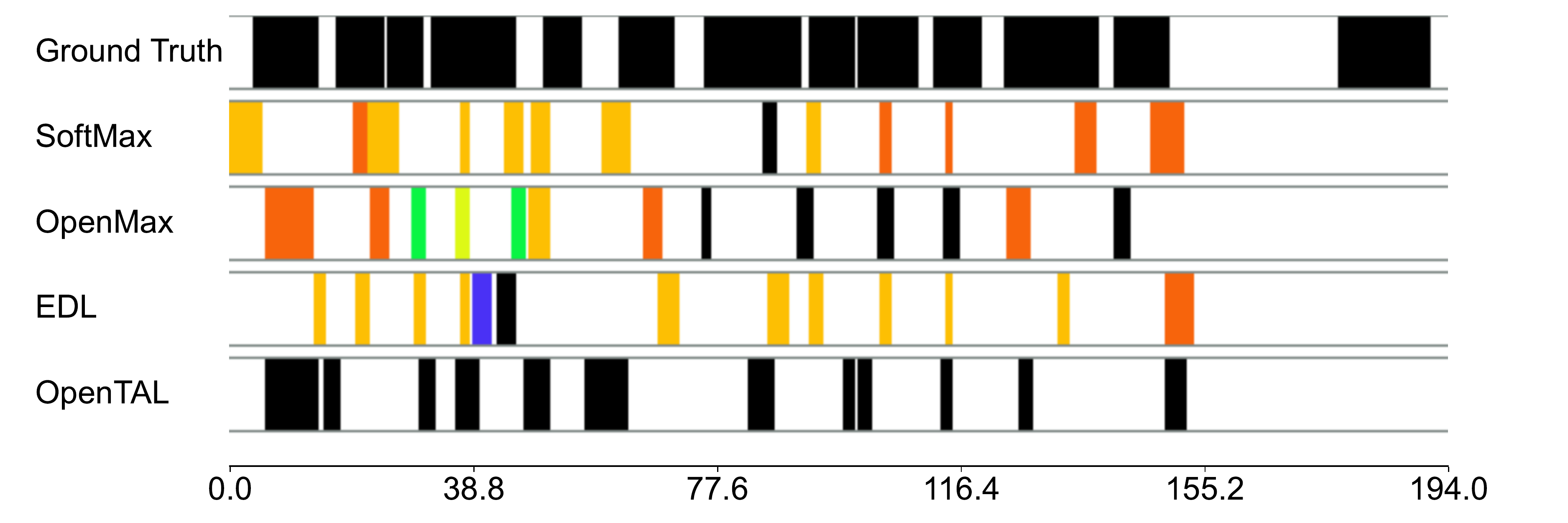}
    \includegraphics[width=0.495\textwidth]{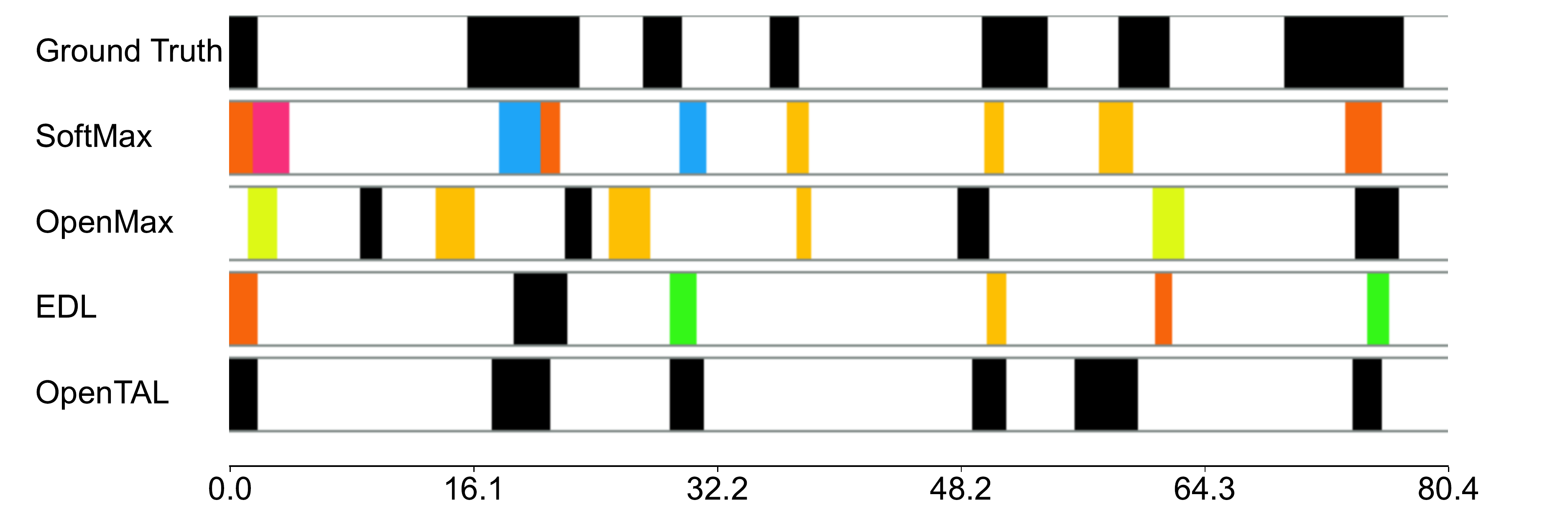}
    \includegraphics[width=0.495\textwidth]{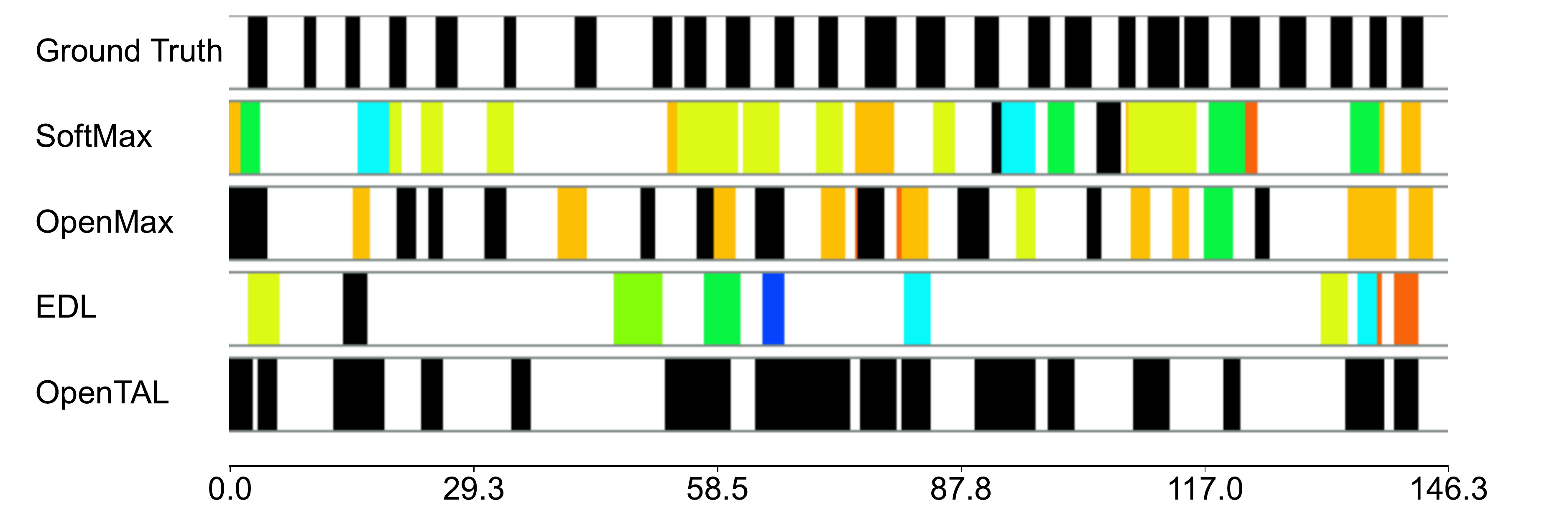}
    \includegraphics[width=0.495\textwidth]{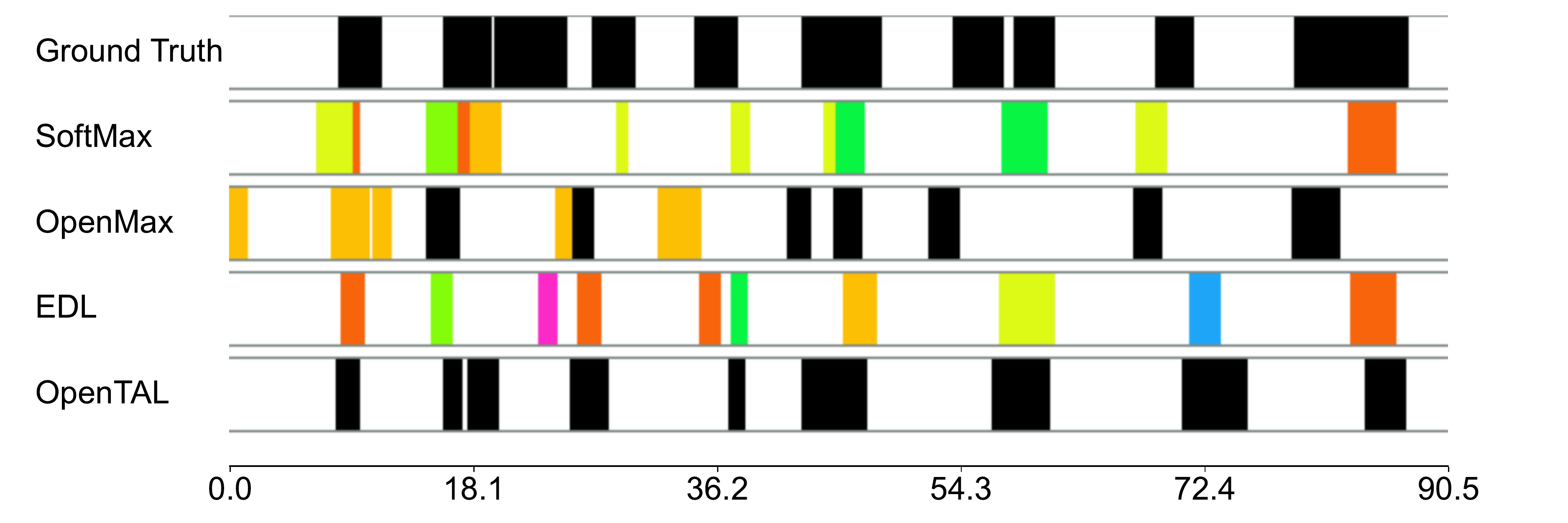}
    \includegraphics[width=0.495\textwidth]{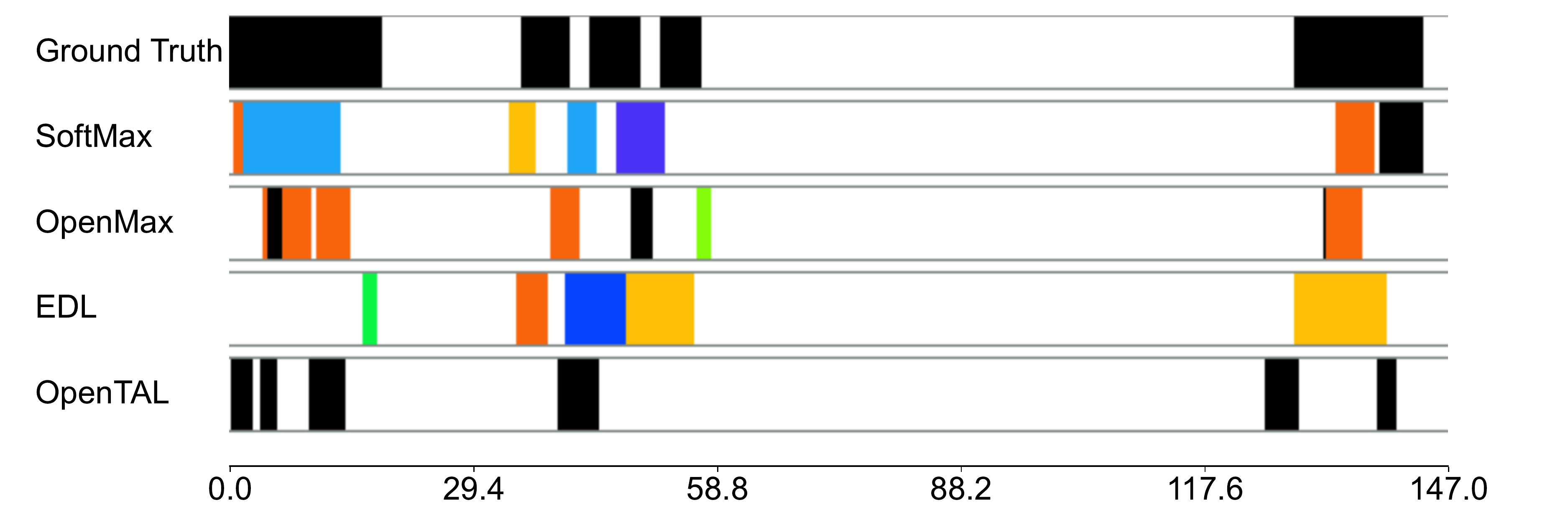}
    \includegraphics[width=0.495\textwidth]{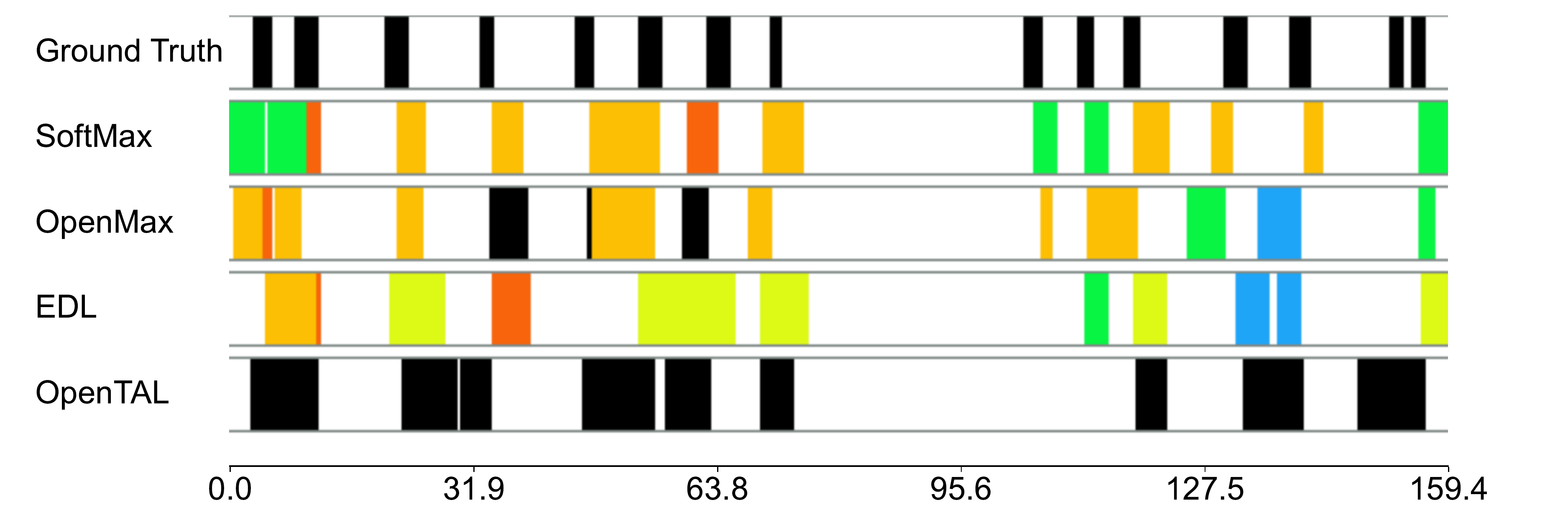}
    \includegraphics[width=0.495\textwidth]{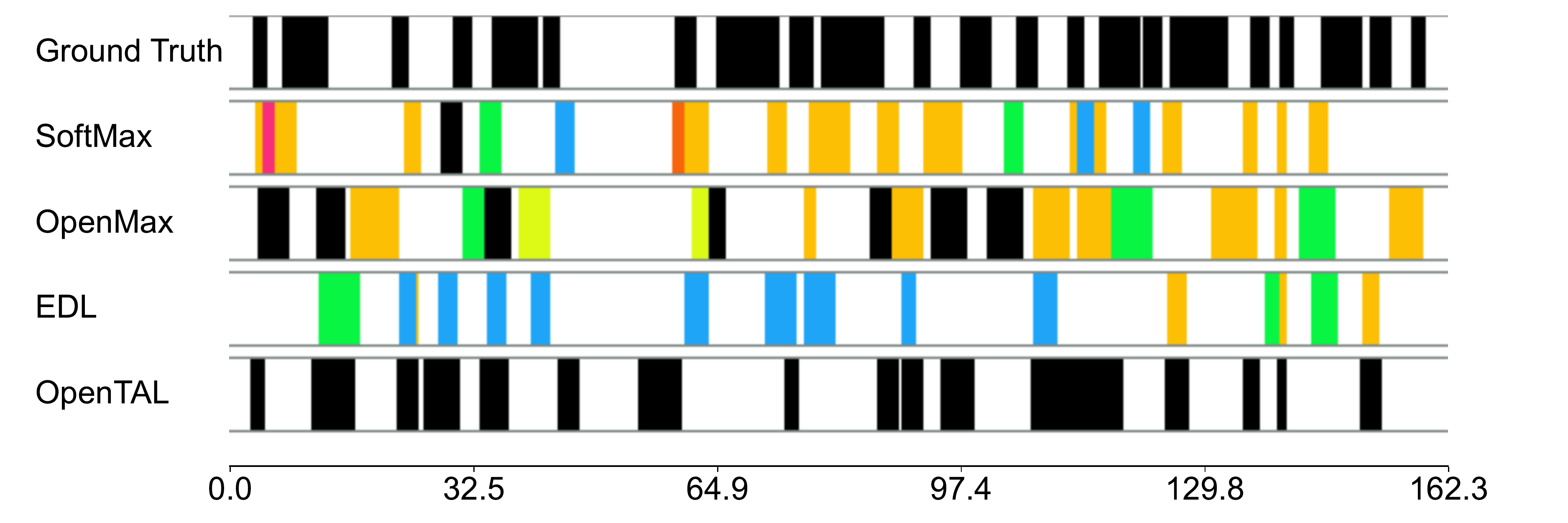}
    \includegraphics[width=0.495\textwidth]{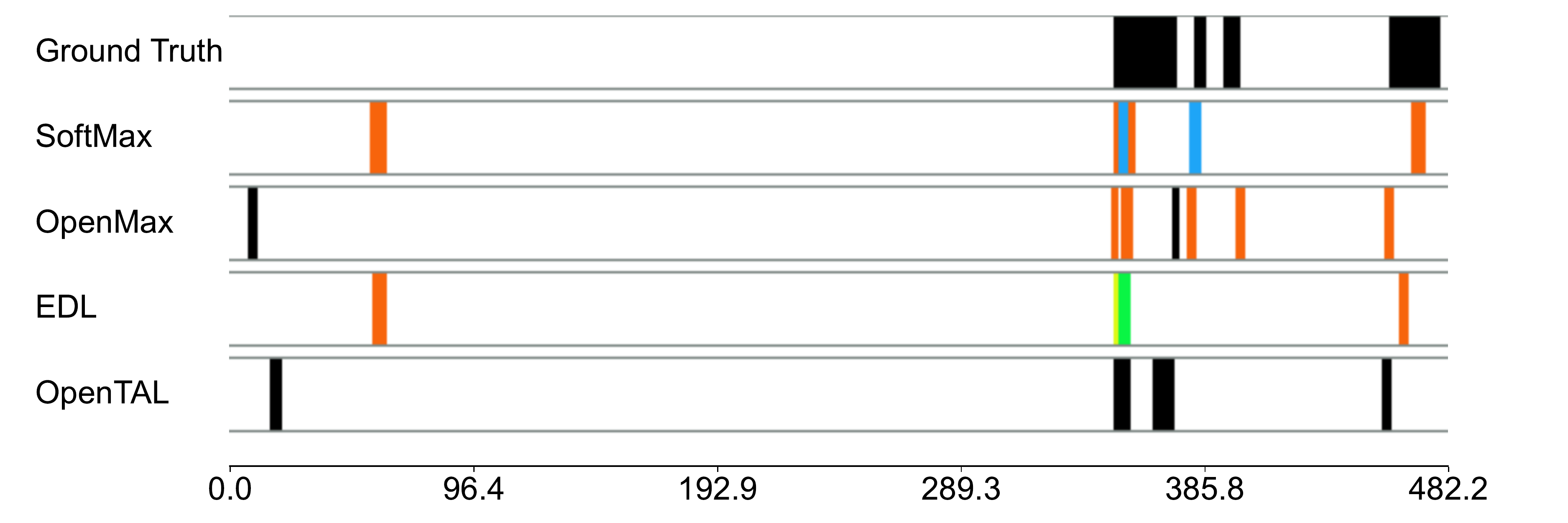}
    \includegraphics[width=0.495\textwidth]{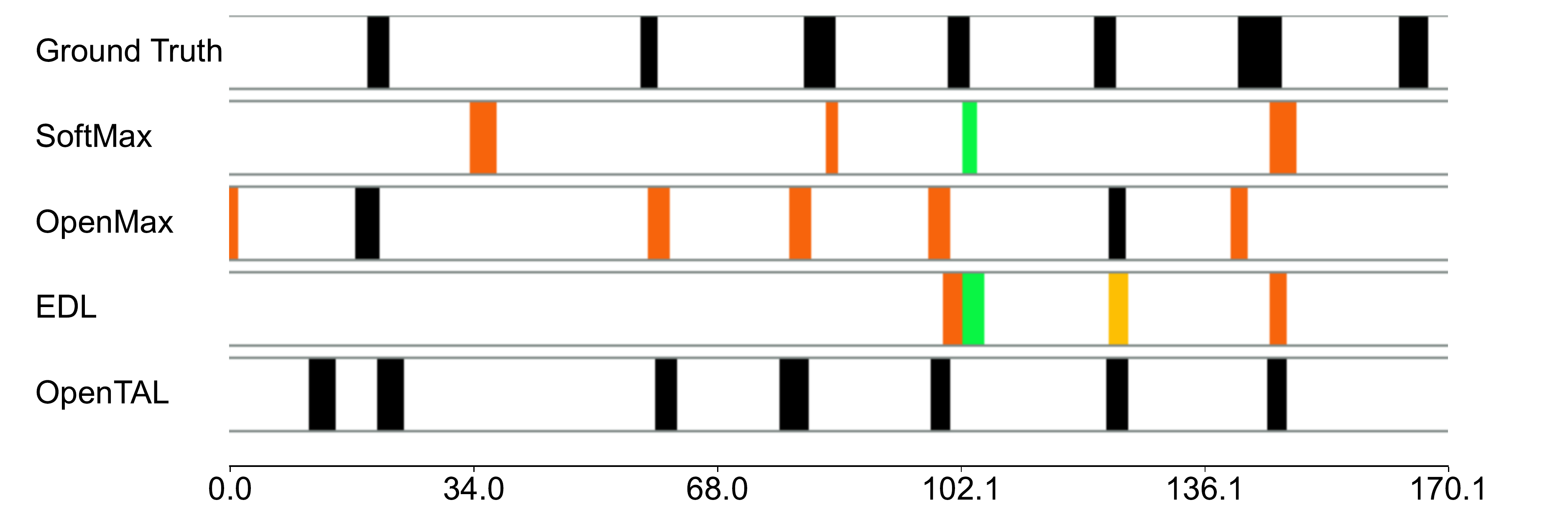}
    \caption{\textbf{Qualitative Results.} We show the actions of unknown classes with black color, while the rest colors are actions of known classes. The $x$-axis represents the timestamps (seconds).}
    \label{fig:supp_demo}
    \vspace{-53.43115pt}
\end{figure*}

%% file: main_v1.bbl
\begin{thebibliography}{10}\itemsep=-1pt

\bibitem{AminiNIPS2020}
Alexander Amini, Wilko Schwarting, Ava Soleimany, and Daniela Rus.
\newblock Deep evidential regression.
\newblock In {\em NeurIPS}, 2020.

\bibitem{bai2020boundary}
Yueran Bai, Yingying Wang, Yunhai Tong, Yang Yang, Qiyue Liu, and Junhui Liu.
\newblock Boundary content graph neural network for temporal action proposal
  generation.
\newblock In {\em ECCV}, 2020.

\bibitem{BaoIROS2020}
Wentao Bao, Qi Yu, and Yu Kong.
\newblock Object-aware centroid voting for monocular 3d object detection.
\newblock In {\em IROS}, 2020.

\bibitem{BaoICCV2021}
Wentao Bao, Qi Yu, and Yu Kong.
\newblock Evidential deep learning for open set action recognition.
\newblock In {\em ICCV}, 2021.

\bibitem{BekkerML2020}
Jessa Bekker and Jesse Davis.
\newblock Learning from positive and unlabeled data: A survey.
\newblock {\em Machine Learning}, 109(4):719--760, 2020.

\bibitem{BendaleCVPR2016}
Abhijit Bendale and Terrance~E. Boult.
\newblock Towards open set deep networks.
\newblock In {\em CVPR}, 2016.

\bibitem{buch2017end}
Shyamal Buch, Victor Escorcia, Bernard Ghanem, Li Fei-Fei, and Juan~Carlos
  Niebles.
\newblock End-to-end, single-stream temporal action detection in untrimmed
  videos.
\newblock In {\em BMVC}, 2017.

\bibitem{ANetCVPR2015}
Fabian Caba~Heilbron, Victor Escorcia, Bernard Ghanem, and Juan Carlos~Niebles.
\newblock Activitynet: A large-scale video benchmark for human activity
  understanding.
\newblock In {\em CVPR}, 2015.

\bibitem{detr_eccv20}
Nicolas Carion, Francisco Massa, Gabriel Synnaeve, Nicolas Usunier, Alexander
  Kirillov, and Sergey Zagoruyko.
\newblock End-to-end object detection with transformers.
\newblock In {\em ECCV}, 2020.

\bibitem{I3DCVPR2017}
J. Carreira and Andrew Zisserman.
\newblock Quo vadis, action recognition? {A} new model and the kinetics
  dataset.
\newblock In {\em CVPR}, 2017.

\bibitem{cenICCV2021}
Jun Cen, Peng Yun, Junhao Cai, Michael~Yu Wang, and Ming Liu.
\newblock Deep metric learning for open world semantic segmentation.
\newblock In {\em ICCV}, 2021.

\bibitem{TALNet_CVPR2018}
Yu-Wei Chao, Sudheendra Vijayanarasimhan, Bryan Seybold, David~A Ross, Jia
  Deng, and Rahul Sukthankar.
\newblock Rethinking the faster r-cnn architecture for temporal action
  localization.
\newblock In {\em CVPR}, 2018.

\bibitem{chenPAMI2021}
Guangyao Chen, Peixi Peng, Xiangqian Wang, and Yonghong Tian.
\newblock Adversarial reciprocal points learning for open set recognition.
\newblock {\em IEEE TPAMI}, 2021.

\bibitem{chenECCV2020}
Guangyao Chen, Limeng Qiao, Yemin Shi, Peixi Peng, Jia Li, Tiejun Huang,
  Shiliang Pu, and Yonghong Tian.
\newblock Learning open set network with discriminative reciprocal points.
\newblock In {\em ECCV}, 2020.

\bibitem{Chen2020ECCV}
Junwen Chen, Wentao Bao, and Yu Kong.
\newblock Group activity prediction with sequential relational anticipation
  model.
\newblock In {\em ECCV}, 2020.

\bibitem{DhamijaWACV2020}
Akshay Dhamija, Manuel Gunther, Jonathan Ventura, and Terrance Boult.
\newblock The overlooked elephant of object detection: Open set.
\newblock In {\em WACV}, 2020.

\bibitem{dhamija2018reducing}
Akshay~Raj Dhamija, Manuel G{\"u}nther, and Terrance~E Boult.
\newblock Reducing network agnostophobia.
\newblock In {\em NeurIPS}, 2018.

\bibitem{DitriaACCV2020}
Luke Ditria, Benjamin~J Meyer, and Tom Drummond.
\newblock {OpenGAN}: Open set generative adversarial networks.
\newblock In {\em ACCV}, 2020.

\bibitem{fangICML2021}
Zhen Fang, Jie Lu, Anjin Liu, Feng Liu, and Guangquan Zhang.
\newblock Learning bounds for open-set learning.
\newblock In {\em ICML}, 2021.

\bibitem{SlowFastICCV2019}
Christoph Feichtenhofer, Haoqi Fan, Jitendra Malik, and Kaiming He.
\newblock Slowfast networks for video recognition.
\newblock In {\em ICCV}, 2019.

\bibitem{GeBMVC2017}
Zongyuan Ge, Sergey Demyanov, Zetao Chen, and Rahil Garnavi.
\newblock Generative {OpenMax} for multi-class open set classification.
\newblock In {\em BMVC}, 2017.

\bibitem{GengTPAMI2020}
Chuanxing Geng, Sheng-jun Huang, and Songcan Chen.
\newblock Recent advances in open set recognition: A survey.
\newblock {\em IEEE TPAMI}, 2020.

\bibitem{GoodfellowNIPS2014}
Ian Goodfellow, Jean Pouget-Abadie, Mehdi Mirza, Bing Xu, David Warde-Farley,
  Sherjil Ozair, Aaron Courville, and Yoshua Bengio.
\newblock Generative adversarial nets.
\newblock In {\em NeurIPS}, 2014.

\bibitem{heilbron2016fast}
Fabian~Caba Heilbron, Juan~Carlos Niebles, and Bernard Ghanem.
\newblock Fast temporal activity proposals for efficient detection of human
  actions in untrimmed videos.
\newblock In {\em CVPR}, 2016.

\bibitem{hwang2021exemplar}
Jaedong Hwang, Seoung~Wug Oh, Joon-Young Lee, and Bohyung Han.
\newblock Exemplar-based open-set panoptic segmentation network.
\newblock In {\em CVPR}, 2021.

\bibitem{JainECCV2014}
Lalit~P Jain, Walter~J Scheirer, and Terrance~E Boult.
\newblock Multi-class open set recognition using probability of inclusion.
\newblock In {\em ECCV}, 2014.

\bibitem{THUMOS14}
Y.-G. Jiang, J. Liu, A. Roshan~Zamir, G. Toderici, I. Laptev, M. Shah, and R.
  Sukthankar.
\newblock {THUMOS} challenge: Action recognition with a large number of
  classes.
\newblock \url{http://crcv.ucf.edu/THUMOS14/}, 2014.

\bibitem{JosangBook2016}
Audun J{\o}sang.
\newblock {\em Subjective logic}.
\newblock Springer, 2016.

\bibitem{joseph2021towards}
KJ Joseph, Salman Khan, Fahad~Shahbaz Khan, and Vineeth~N Balasubramanian.
\newblock Towards open world object detection.
\newblock In {\em CVPR}, 2021.

\bibitem{junior2016specialized}
Pedro Ribeiro~Mendes J{\'u}nior, Terrance~E Boult, Jacques Wainer, and Anderson
  Rocha.
\newblock Specialized support vector machines for open-set recognition.
\newblock {\em arXiv preprint arXiv:1606.03802}, 2016.

\bibitem{kingmaICLR2014}
Diederik~P Kingma and Max Welling.
\newblock Auto-encoding variational bayes.
\newblock In {\em ICLR}, 2014.

\bibitem{koh2017understanding}
Pang~Wei Koh and Percy Liang.
\newblock Understanding black-box predictions via influence functions.
\newblock In {\em ICML}, 2017.

\bibitem{kong2021opengan}
Shu Kong and Deva Ramanan.
\newblock {OpenGAN}: Open-set recognition via open data generation.
\newblock In {\em ICCV}, 2021.

\bibitem{KongArXiv2018}
Yu Kong and Yun Fu.
\newblock Human action recognition and prediction: A survey.
\newblock {\em arXiv preprint arXiv:1806.11230}, 2018.

\bibitem{LiAAAI2019}
Buyu Li, Yu Liu, and Xiaogang Wang.
\newblock Gradient harmonized single-stage detector.
\newblock In {\em AAAI}, 2019.

\bibitem{AFSD_CVPR2021}
Chuming Lin, Chengming Xu, Donghao Luo, Yabiao Wang, Ying Tai, Chengjie Wang,
  Jilin Li, Feiyue Huang, and Yanwei Fu.
\newblock Learning salient boundary feature for anchor-free temporal action
  localization.
\newblock In {\em CVPR}, 2021.

\bibitem{lin2019bmn}
Tianwei Lin, Xiao Liu, Xin Li, Errui Ding, and Shilei Wen.
\newblock Bmn: Boundary-matching network for temporal action proposal
  generation.
\newblock In {\em ICCV}, 2019.

\bibitem{lin2018bsn}
Tianwei Lin, Xu Zhao, Haisheng Su, Chongjing Wang, and Ming Yang.
\newblock Bsn: Boundary sensitive network for temporal action proposal
  generation.
\newblock In {\em ECCV}, 2018.

\bibitem{LinICCV2017}
Tsung-Yi Lin, Priya Goyal, Ross Girshick, Kaiming He, and Piotr Doll{\'a}r.
\newblock Focal loss for dense object detection.
\newblock In {\em ICCV}, 2017.

\bibitem{long2019gaussian}
Fuchen Long, Ting Yao, Zhaofan Qiu, Xinmei Tian, Jiebo Luo, and Tao Mei.
\newblock Gaussian temporal awareness networks for action localization.
\newblock In {\em CVPR}, 2019.

\bibitem{miller2018dropout}
Dimity Miller, Lachlan Nicholson, Feras Dayoub, and Niko S{\"u}nderhauf.
\newblock Dropout sampling for robust object detection in open-set conditions.
\newblock In {\em ICRA}, 2018.

\bibitem{mundt2019open}
Martin Mundt, Iuliia Pliushch, Sagnik Majumder, and Visvanathan Ramesh.
\newblock Open set recognition through deep neural network uncertainty: Does
  out-of-distribution detection require generative classifiers?
\newblock In {\em ICCVW}, 2019.

\bibitem{neal2018open}
Lawrence Neal, Matthew Olson, Xiaoli Fern, Weng-Keen Wong, and Fuxin Li.
\newblock Open set learning with counterfactual images.
\newblock In {\em ECCV}, 2018.

\bibitem{oliveira2021fully}
Hugo Oliveira, Caio Silva, Gabriel~LS Machado, Keiller Nogueira, and
  Jefersson~A dos Santos.
\newblock Fully convolutional open set segmentation.
\newblock {\em Machine Learning}, pages 1--52, 2021.

\bibitem{OzaCVPR2019}
Poojan Oza and Vishal~M Patel.
\newblock {C2AE}: Class conditioned auto-encoder for open-set recognition.
\newblock In {\em CVPR}, 2019.

\bibitem{Park_2021_ICCV}
Seulki Park, Jongin Lim, Younghan Jeon, and Jin~Young Choi.
\newblock Influence-balanced loss for imbalanced visual classification.
\newblock In {\em ICCV}, 2021.

\bibitem{PereraCVPR2020}
Pramuditha Perera, Vlad~I Morariu, Rajiv Jain, Varun Manjunatha, Curtis
  Wigington, Vicente Ordonez, and Vishal~M Patel.
\newblock Generative-discriminative feature representations for open-set
  recognition.
\newblock In {\em CVPR}, 2020.

\bibitem{pham2018bayesian}
Trung Pham, Thanh-Toan Do, Gustavo Carneiro, Ian Reid, et~al.
\newblock Bayesian semantic instance segmentation in open set world.
\newblock In {\em ECCV}, 2018.

\bibitem{fasterrcnn_tpami}
Shaoqing Ren, Kaiming He, Ross Girshick, and Jian Sun.
\newblock Faster {R-CNN}: Towards real-time object detection with region
  proposal networks.
\newblock {\em IEEE TPAMI}, 39(6):1137--1149, 2017.

\bibitem{saito2021NIPS}
Kuniaki Saito, Donghyun Kim, and Kate Saenko.
\newblock Openmatch: Open-set semi-supervised learning with open-set
  consistency regularization.
\newblock In {\em NeurIPS}, 2021.

\bibitem{ScheirerTPAMI2012}
Walter~J Scheirer, Anderson de Rezende~Rocha, Archana Sapkota, and Terrance~E
  Boult.
\newblock Toward open set recognition.
\newblock {\em IEEE TPAMI}, 35(7):1757--1772, 2012.

\bibitem{ScheirerTPAMI2014}
Walter~J Scheirer, Lalit~P Jain, and Terrance~E Boult.
\newblock Probability models for open set recognition.
\newblock {\em IEEE TPAMI}, 36(11):2317--2324, 2014.

\bibitem{SensoyNIPS2018}
Murat Sensoy, Lance Kaplan, and Melih Kandemir.
\newblock Evidential deep learning to quantify classification uncertainty.
\newblock In {\em NeurIPS}, 2018.

\bibitem{SentzBook2002}
Kari Sentz, Scott Ferson, et~al.
\newblock {\em Combination of evidence in Dempster-Shafer theory}, volume 4015.
\newblock Sandia National Laboratories Albuquerque, 2002.

\bibitem{shou2017cdc}
Zheng Shou, Jonathan Chan, Alireza Zareian, Kazuyuki Miyazawa, and Shih-Fu
  Chang.
\newblock Cdc: Convolutional-de-convolutional networks for precise temporal
  action localization in untrimmed videos.
\newblock In {\em CVPR}, 2017.

\bibitem{shu2020p}
Yu Shu, Yemin Shi, Yaowei Wang, Tiejun Huang, and Yonghong Tian.
\newblock p-odn: prototype-based open deep network for open set recognition.
\newblock {\em Scientific reports}, 10(1):1--13, 2020.

\bibitem{sridhar2021class}
Deepak Sridhar, Niamul Quader, Srikanth Muralidharan, Yaoxin Li, Peng Dai, and
  Juwei Lu.
\newblock Class semantics-based attention for action detection.
\newblock In {\em ICCV}, 2021.

\bibitem{SunCVPR2020}
Xin Sun, Zhenning Yang, Chi Zhang, Keck-Voon Ling, and Guohao Peng.
\newblock Conditional {Gaussian} distribution learning for open set
  recognition.
\newblock In {\em CVPR}, 2020.

\bibitem{WangICCV2021Uniden}
Weiyao Wang, Matt Feiszli, Heng Wang, and Du Tran.
\newblock Unidentified video objects: A benchmark for dense, open-world
  segmentation.
\newblock In {\em ICCV}, 2021.

\bibitem{wangICCV2021}
Yezhen Wang, Bo Li, Tong Che, Kaiyang Zhou, Ziwei Liu, and Dongsheng Li.
\newblock Energy-based open-world uncertainty modeling for confidence
  calibration.
\newblock In {\em ICCV}, 2021.

\bibitem{xu2017r}
Huijuan Xu, Abir Das, and Kate Saenko.
\newblock R-c3d: Region convolutional 3d network for temporal activity
  detection.
\newblock In {\em ICCV}, 2017.

\bibitem{GTAD_CVPR2020}
Mengmeng Xu, Chen Zhao, David~S Rojas, Ali Thabet, and Bernard Ghanem.
\newblock G-tad: Sub-graph localization for temporal action detection.
\newblock In {\em CVPR}, 2020.

\bibitem{YangCVPR2018}
Hong-Ming Yang, Xu-Yao Zhang, Fei Yin, and Cheng-Lin Liu.
\newblock Robust classification with convolutional prototype learning.
\newblock In {\em CVPR}, 2018.

\bibitem{YangTPAMI2020}
Hong-Ming Yang, Xu-Yao Zhang, Fei Yin, Qing Yang, and Cheng-Lin Liu.
\newblock Convolutional prototype network for open set recognition.
\newblock {\em IEEE TPAMI}, 2020.

\bibitem{yeung2016end}
Serena Yeung, Olga Russakovsky, Greg Mori, and Li Fei-Fei.
\newblock End-to-end learning of action detection from frame glimpses in
  videos.
\newblock In {\em CVPR}, 2016.

\bibitem{yoshihashiCVPR2019}
Ryota Yoshihashi, Wen Shao, Rei Kawakami, Shaodi You, Makoto Iida, and Takeshi
  Naemura.
\newblock Classification-reconstruction learning for open-set recognition.
\newblock In {\em CVPR}, 2019.

\bibitem{yu2020multi}
Qing Yu, Daiki Ikami, Go Irie, and Kiyoharu Aizawa.
\newblock Multi-task curriculum framework for open-set semi-supervised
  learning.
\newblock In {\em ECCV}, 2020.

\bibitem{Yue_2021_CVPR}
Zhongqi Yue, Tan Wang, Qianru Sun, Xian-Sheng Hua, and Hanwang Zhang.
\newblock Counterfactual zero-shot and open-set visual recognition.
\newblock In {\em CVPR}, 2021.

\bibitem{PGCN_ICCV2019}
Runhao Zeng, Wenbing Huang, Mingkui Tan, Yu Rong, Peilin Zhao, Junzhou Huang,
  and Chuang Gan.
\newblock Graph convolutional networks for temporal action localization.
\newblock In {\em ICCV}, 2019.

\bibitem{zhangICCV2021}
Hui Zhang and Henghui Ding.
\newblock Prototypical matching and open set rejection for zero-shot semantic
  segmentation.
\newblock In {\em ICCV}, 2021.

\bibitem{zhao2020bottom}
Peisen Zhao, Lingxi Xie, Chen Ju, Ya Zhang, Yanfeng Wang, and Qi Tian.
\newblock Bottom-up temporal action localization with mutual regularization.
\newblock In {\em ECCV}, 2020.

\bibitem{ZhaoAAAI2019}
Xujiang Zhao, Yuzhe Ou, Lance Kaplan, Feng Chen, and Jin-Hee Cho.
\newblock Quantifying classification uncertainty using regularized evidential
  neural networks.
\newblock In {\em AAAI}, 2019.

\bibitem{zhao2017temporal}
Yue Zhao, Yuanjun Xiong, Limin Wang, Zhirong Wu, Xiaoou Tang, and Dahua Lin.
\newblock Temporal action detection with structured segment networks.
\newblock In {\em ICCV}, 2017.

\bibitem{zhou2021learning}
Da-Wei Zhou, Han-Jia Ye, and De-Chuan Zhan.
\newblock Learning placeholders for open-set recognition.
\newblock In {\em CVPR}, 2021.

\bibitem{zhu2021enriching}
Zixin Zhu, Wei Tang, Le Wang, Nanning Zheng, and Gang Hua.
\newblock Enriching local and global contexts for temporal action localization.
\newblock In {\em ICCV}, 2021.

\bibitem{Zou2021MM}
Yixiong Zou, Shanghang Zhang, Guangyao Chen, Yonghong Tian, Kurt Keutzer, and
  Jos\'{e} M.~F. Moura.
\newblock Annotation-efficient untrimmed video action recognition.
\newblock In {\em ACM MM}, 2021.

\end{thebibliography}
